\documentclass[10pt,journal,compsoc]{IEEEtran}
\ifCLASSOPTIONcompsoc
\else
\fi
\usepackage[utf8]{inputenc} % allow utf-8 input
\usepackage[T1]{fontenc}    % use 8-bit T1 fonts
\usepackage{hyperref}       % hyperlinks
\usepackage{url}            % simple URL typesetting
\usepackage{booktabs}       % professional-quality tables
\usepackage{amsfonts}       % blackboard math symbols
\usepackage{nicefrac}       % compact symbols for 1/2, etc.
\usepackage{microtype}      % microtypography
\usepackage{bm}
\usepackage{enumitem}
\usepackage{floatrow}
\usepackage{algorithm}
\usepackage{algorithmic}
\usepackage{graphicx}
\usepackage{epstopdf}
\usepackage{multirow}
\usepackage{amsmath}
\usepackage{amssymb}
\usepackage[american]{babel}
\usepackage{subcaption}
\usepackage{color}

\newcommand{\argmax}{\mathop{\arg\max}}

\newcommand{\x}{{\bf x}}

\newcommand{\s}{{\bf s}}

\newcommand{\p}{{\bf p}}

\newcommand{\w}{{\bf w}}

\newcommand{\y}{{\bf y}}

\newcommand{\ie}{{\it i.e.}}
\newcommand{\eg}{{\it e.g.}}
\newcommand{\name}{\textsc{ReFilled}}

\ifCLASSINFOpdf
\else
\fi

\begin{document}
\title{Generalized Knowledge Distillation\\ via Relationship Matching}

\author{Han-Jia Ye,
	    Su Lu,
        De-Chuan Zhan
\IEEEcompsocitemizethanks{
\IEEEcompsocthanksitem H.-J. Ye, S. Lu, and D.-C. Zhan are with State Key Laboratory for Novel Software Technology,
       Nanjing University, Nanjing, 210023, China.
       \protect\\
E-mail: \{yehj,lus,zhandc\}@lamda.nju.edu.cn}% <-this % stops an unwanted space
}


\markboth{Journal of \LaTeX\ Class Files,~Vol.~Xx, No.~X, Xxxx~20Xx}%
{Ye \MakeLowercase{\textit{et al.}}: Generalized Knowledge Distillation via Relationship Matching}



\IEEEtitleabstractindextext{%
\begin{abstract}
The knowledge of a well-trained deep neural network (a.k.a. the ``teacher'') is valuable for learning similar tasks.
Knowledge distillation extracts knowledge from the teacher and integrates it with the target model (a.k.a. the ``student''), which expands the student's knowledge and improves its learning efficacy.
Instead of enforcing the teacher to work on the same task as the student, we borrow the knowledge from a {\em  teacher trained from a general label space} --- in this ``Generalized Knowledge Distillation~(GKD)'', the classes of the teacher and the student {\em may be the same, completely different, or partially overlapped}.
We claim that the comparison ability between instances acts as an essential factor threading knowledge across tasks, and propose the \textbf{RE}lationship \textbf{F}ac\textbf{I}litated \textbf{L}ocal c\textbf{L}assifi\textbf{E}r \textbf{D}istillation~({\name}) approach, which decouples the GKD flow of the embedding and the top-layer classifier. 
In particular, different from reconciling the instance-label confidence between models, {\name} requires the teacher to {\em reweight} the hard tuples pushed forward by the student and then matches the similarity comparison levels between instances.
An embedding-induced classifier based on the teacher model supervises the student's classification confidence and adaptively emphasizes the most related supervision from the teacher.
{\name} demonstrates strong discriminative ability when the classes of the teacher vary from the same to a fully non-overlapped set w.r.t. the student. It also achieves state-of-the-art performance on standard knowledge distillation, one-step incremental learning, and few-shot learning tasks.
\end{abstract}

}

\maketitle

\IEEEpeerreviewmaketitle

\IEEEraisesectionheading{\section{Introduction}}\label{sec:intro}
\IEEEPARstart{S}{upervised} deep learning has demonstrated success in a variety of fields~\cite{Krizhevsky2017ImageNet}. 
Given the instances and corresponding annotations from the target task, we train a deep neural network to minimize the discrepancy between the model predictions and the ground-truth labels.
Knowledge distillation~(KD)~\cite{BucilaCN06,hinton2015kd,YimJBK17} facilitates the learning efficiency of a deep neural network via taking advantage of the ``dark knowledge'' from another well-trained model.
In detail, a strong classifier, \eg, a neural network trained with deeper architectures~\cite{RomeroBKCGB14}, high-quality images~\cite{Yu2019Learning}, or precise optimization strategies~\cite{FurlanelloLTIA18,Yang2018Snapshot}, acts as a ``teacher'' and guides the training of a ``student'' model by richer supervision, so that the learning experience from a related task is reused in the current task.
KD improves the discriminative ability of the target student model~\cite{Liu2020More,Passalis2020Heterogeneous}, relieves the burden of model storage~\cite{hinton2015kd,RomeroBKCGB14,YimJBK17,FurlanelloLTIA18,Hou2020DynaBERT,Wang2020MiniLM} and enables the training of a deep neural network in low-resource environments~\cite{Guo2020Online,Li2020Few}.
Applications of KD have been witnessed in a wide range of domains such as model/dataset compression~\cite{Wang2018Dataset,Ahn2019Variational,Mirzadeh2019Improved,Nayak2019Zero,Cho2019On,Zhao2020Dataset}, multi-task learning~\cite{Zhang2018Deep,Kundu2019UM}, and incremental image classification~\cite{Zhou2019M2KD,Javed2018Revisiting}. 

The teacher's class posterior probability over an instance is the most common dark knowledge, as it indicates the teacher's estimation of how similar an instance is to candidate categories.
Besides the extreme ``black or white'' supervision, the student is asked to align its posterior with the teacher during its training progress.
Although prediction matching allows knowledge to be transferred across different architectures~\cite{hinton2015kd,Mirzadeh2019Improved}, its dependence on instance-label relationship restricts both teacher and student to the same label space. 

\begin{figure}[!t]
	\centering
	\includegraphics[width=0.95\textwidth]{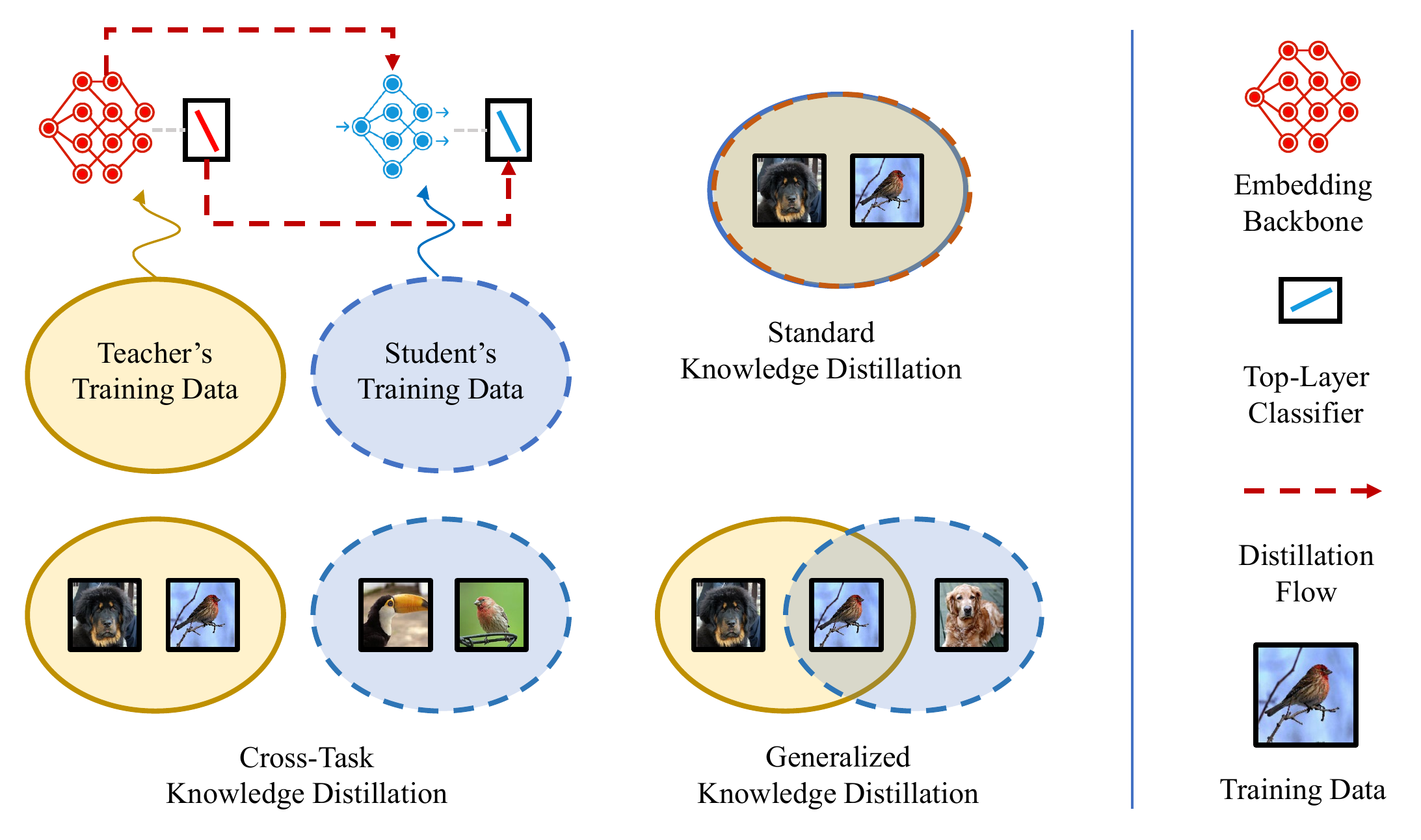}\\
	\caption{\small An illustration of strengthening a student model on the target task via distilling the knowledge from a teacher model. 
	In standard Knowledge Distillation (KD), teacher and student share the same set of classes. In cross-task KD, a teacher is learned from images with fully non-overlapping classes, while its learning experience is distilled to facilitate the training of the student model. Generalized KD is a more general case of the previous two, where the student could have {\em the same, different, or partially overlapped} classes w.r.t. the teacher. 
	The red dotted line indicates the distillation flow for the embedding backbone and the top-layer classifier, respectively, in our proposed model.}
	\label{fig:teaser}
\end{figure}

We emphasize the necessity to use a general teacher and extend KD to more practical applications. In other words, the related teacher should not be limited to having classes that are completely same as the target task. For example, it is intuitive to reuse a well-trained model classifying animals to help the training of a student model on fine-grained birds. 
In {\em cross-task KD}, a student distills the knowledge from a teacher trained on related but non-overlapping label spaces, where the label discrepancy between teacher and student impedes the learning experience transition~\cite{Hsu2018Learning}.
{\em Generalized Knowledge Distillation}~(GKD) is a general case of standard KD and cross-task KD, where a teacher could have {\em the same, fully different, or partially overlapped} classes w.r.t. the student.
Figure~\ref{fig:teaser} illustrates the notion of KD variants.

The {\em comparison ability} --- measuring similarity level between two instances based on their embeddings --- captures a kind of invariant nature of the model~\cite{Achille2018Emergence} and is {\em free from the label constraint}~\cite{SongXJS16,ManmathaWSK17,Hsu2018Learning}.
For a teacher and a student discerning `Husky vs. Birman'' and ``Poodle vs. Persian'' respectively, the teacher's discriminative embedding encoding the ``dog-cat'' related characteristics can compare Poodle/Persian in the student's task and should be helpful for student's training. 
We expect the student to benefit from the teacher's knowledge if they are related, \ie, the teacher's comparison ability fits the student's task. Otherwise, the student will perform as well as the one trained without a teacher.
Thus, we bridge the knowledge transfer in GKD with the {\em instance-wise relationship} and thread the knowledge reuse for both embedding and top-layer classifier by taking advantage of the teacher's comparison ability. 

To this end, we propose a 2-stage approach \textbf{RE}lationship \textbf{F}ac\textbf{I}litated \textbf{L}ocal c\textbf{L}assifi\textbf{E}r \textbf{D}istillation~({\name}) based on current task's data and a well-trained teacher.
First, the discriminative ability of embedding is stressed. For those hard similarity tuples determined by the student's embedding, how the teacher compares them acts as additional supervision. 
In other words, the teacher promotes the discriminative ability of the student's embedding by specifying {\em how much} dissimilar an object's impostors should be far away from its target nearest neighbor. The instance-wise knowledge from the teacher is distilled via matching comparisons.
The teacher then constructs soft supervision for classifying each instance based on the similarity between the instance and an embedding center, eliminating the restrictions between label spaces. Specifically, the classification confidences of the student are aligned with the embedding-induced ``instance-label'' predictions from the teacher. The strength of the supervision is weighted by the relatedness between teacher and student automatically.

Empirical results verify that {\name} effectively transfers the classification ability from various configurations of teachers to a student, including teachers with the same, different, and partially overlapped label spaces.
{\name} also outperforms recent methods in one-step incremental learning, few-shot learning, and middle-shot learning problems.
In summary, our contributions are
\begin{itemize}
	\item We investigate GKD to enhance the training efficiency of a deep neural network by reusing the knowledge from a well-trained teacher {\em without label restriction}.
	\item We propose {\name} which aligns the high-order comparison between models locally and weights the most helpful supervision from the teacher adaptively.
	\item {\name} works well in generalized KD, incremental learning, and few-shot learning benchmarks.
\end{itemize}

After the related literature and the preliminary in Section~\ref{sec:related} and Section~\ref{sec:problem}, we formalize our {\name} approach in Section~\ref{sec:method}. Finally are experiments and conclusion.

\section{Related Work}
\label{sec:related}
\noindent{\bf Knowledge Distillation~(KD).} Rich supervision plays a crucial role in building a machine learning or visual recognition system, where taking advantage of the learning experience from related pre-trained models becomes a shortcut to facilitate the model training in the current task~\cite{Zhou2016Learnware}. 
Different from fine-tuning~\cite{HeZRS15} or weights  matching~\cite{Kuzborskij2017Fast,DuKSP17Hypothesis,LiGD18Explicit,Srinivas2018Knowledge,Ye2018Rectify,Ye2021Heterogeneous} that regularize the model from the ``parameters'' perspective, we can reuse the dark knowledge/privileged information~\cite{Vapnik2009Learning,Vapnik2015Learning,Vapnik2016Learning} to explain or assist the training process of the model from the ``data'' aspect~\cite{ZhouJ04,hinton2015kd}.
Denote a fixed well-trained model from a related task and the model in the current task as the ``teacher'' and the ``student'', respectively. KD matches the behaviors of two models on the current task's data~\cite{Phuong2019Towards,Gotmare2019Closer,Cho2019On}.
The teacher could be a high-capacity deep neural network trained on the same task~\cite{Guo2020Online,Liu2020More,Tao2020Topology} or a previous generation of the model along the training progress~\cite{Bagherinezhad2018Label,FurlanelloLTIA18,Yang2018Snapshot}.
The dark knowledge in KD can be implemented as the soft label, \ie, the posterior probability of an instance~\cite{hinton2015kd,SauB16,Mirzadeh2019Improved}, hidden layer activation~\cite{RomeroBKCGB14,ZagoruykoK16a,Czarnecki2017Sobolev,Koratana2019LIT}, parameter flows~\cite{YimJBK17}, transformations~\cite{LeeKS18Self}, and path-wise statistics~\cite{Li2020Local}.
Distilling the knowledge from one model to another has been investigated for model compression~\cite{BucilaCN06,Hou2020DynaBERT,Wang2020MiniLM,Li2020Residual} and incremental learning~\cite{Li2018Learning,Liu2020More,Tao2020Topology}.

\noindent{\bf Cross-task knowledge transfer.} 
In practical applications, the knowledge from teachers of other tasks, \ie, teachers trained on non-overlapping sets of labels, also assists the training of a target task student. 
Heterogeneous transfer learning updates both student and teacher on the current and related domains (resp. tasks) to close distribution (resp. label) divergence~\cite{Kundu2019UM,MaoMCSC21}. Heterogeneous model reuse takes advantage of the teacher from a related task, which relieves the burden of data storage so as to decrease the risk of privacy leaking~\cite{Ye2018Rectify,Wu2019Heterogeneous}.
Meta-learning has also been utilized to transfer knowledge across different label spaces, \eg, few-shot learning~\cite{Vinyals2016Matching,Snell2017Proto,Finn2017MAML,Qi2018Lowshot,Ye2018Learning,PangZWXG21}. These approaches usually require special training strategies of the teacher.
A {\em fixed} well-trained teacher is provided in KD, but since KD usually relies on the correspondence between classifier and categories, it is challenging to reuse the classification knowledge from a cross-task teacher. In {\name}, we bridge the label divergence via comparison matching. 

Class incremental learning also takes advantage of KD in a cross-task scenario, where non-overlapped sets of classes arrive sequentially. The classifier on previously seen classes is the teacher, which is incorporated in training the current stage's student without storing historical data~\cite{Javed2018Revisiting,Liu2020More,Tao2020Topology,Zhou2021Co}. 
KD helps avoid catastrophic forgetting by matching the student's predictions over previous classes with the teacher~\cite{Rebuffi17iCARL,Li2018Learning}.
The goal of {\name} is not to avoid forgetting but transferring the knowledge of the teacher to improve the target classifier. Furthermore, {\name} utilizes teachers with general label spaces and transfers the classification ability {\em across different architectures}.

\noindent{\bf Embedding learning for KD.}
Embedding learning improves the feature representation by pulling similar instances together and pushing dissimilar ones away~\cite{WeinbergerS2009Distance,SchroffKP15,ManmathaWSK17,Ye2019What,Chen2020Simple,Ye2020Learning}. 
Benefited from kinds of side-information~\cite{DavisKJSD07ITML,WeinbergerS2009Distance}, embeddings are learned to explain the given instance-wise relationships~\cite{maaten2008visualizing,Maaten2012stochastic,Amid2015Multiview}. 
Instead of matching the instance-label predictions between models, matching the embedding~\cite{Chen2018Learning,Ahn2019Variational,Budnik2021Asymmetric,chen2021Wasserstein,Heo2019Comprehensive}, pairwise distance~\cite{Peng2019Correlation,Tung2019Similarity}, and similarity graph~\cite{Liu2019Knowledge,Hu2020Creating,Li2020Local,Oki2020Triplet,Zhu2021Complementary} have been investigated. Then ``downstream'' cross-task clustering and representation learning tasks could be improved~\cite{Hsu2018Learning,Qi2018Lowshot,Yu2019Learning}. 
For example, RKD~\cite{Park2019Relational} constructs angels over triplets and matches the angels by regression.
We emphasize the relationship matching in distilling from a {\em general teacher} trained from possible {\em in-task and cross-task} classes w.r.t. the target student.
In our embedding distillation stage, we organize instances in a {\em tuple}, which captures high-order local comparisons efficiently and provides richer supervision from the teacher. The superiority of {\name} is validated in experiments.

Some concurrent Self-Supervised Learning~(SSL) methods distill the relationship in mini-batches~\cite{Koohpayegani2020Compress,Tejankar2020ISD,Xu20Knowledge,Fang2021Seed}. A stronger model or previous generations during the training progress becomes the teacher to compress the model or improve the embedding quality. Different from constructing comparisons with class semantics, in SSL, augmented views of an instance are treated as similar ones while different instances are dissimilar. 
{\name} utilizes the characteristic of embeddings to bridge the class gap in GKD, and further emphasize the transfer of classification ability. We investigate our differences with SSL w.r.t. both the distillation strategy and the similarity measure in experiments. 

\section{Knowledge Reuse via Distillation}
\label{sec:problem}
We briefly introduce the standard Knowledge Distillation~(KD) via matching the soft labels at first. Then we describe the concrete settings of cross-task knowledge distillation and Generalized Knowledge Distillation (GKD).

\subsection{Background and Notations}
For a $C$-class classification task, we denote the training data with $N$ examples as $\mathcal{D} = \{(\x_i, \y_i)\}_{i=1}^N$, where $\x_i\in\mathbb{R}^D$ and $\y_i\in\{0,1\}^C$ are the instance and the corresponding one-hot label, respectively. 
Index of 1 in $\y_i$ indicates the class of $\x_i$.
Denote the class set as $\mathcal{C}$, where $|\mathcal{C}|=C$.
A classifier $f(\x)\colon \mathbb{R}^D \mapsto \{0,1\}^C$ (\eg, a deep neural network) predicts the label for an instance $\x$, which could be represented as $f(\x)=W^\top\phi(\x)$.\footnote{We omit the bias term for discussion simplicity.} There are two components in $f$, the feature extractor $\phi\colon \mathbb{R}^D \mapsto \mathbb{R}^d$ mapping the raw input to a $d$ dimensional latent space, and a linear classifier $W=[\w_1,\ldots,\w_C]\in\mathbb{R}^{d\times C}$ based on the extracted features.
The objective minimizes the discrepancy between the prediction and the true label over all instances in $\mathcal{D}$:
\begin{equation}
	\min_{f} \sum_{i=1}^N \ell\left(f(\x_i),\;\y_i\right)\;.\label{eq:vanilla_train}
\end{equation}
$\ell$ is the loss such as the cross-entropy. We denote optimizing Eq.~\ref{eq:vanilla_train} from scratch as the vanilla supervised deep learning.

\subsection{Standard Knowledge Distillation}
Given a well-trained $C$-class classifier $f_T$ for the class set $\mathcal{C}'$ with $\mathcal{C}'=\mathcal{C}$, it is an effective manner to distill the ``dark knowledge'' from the {\em fixed} $f_T$ to help the training progress of the target model $f$. Subscript ``T'' denotes the model/parameters of the teacher.
To improve the training efficacy of $f$, Hinton, Vinyals, and Dean~\cite{hinton2015kd} suggest to align the soft targets of two models besides the vanilla objective:
\begin{equation}   \label{eq:kd}
	\min_{f}\sum_{i=1}^N\ell(f(\x_i),\y_i) + \lambda\mathcal{R}(\s_{\tau}(f_T(\x_i)),\s_{\tau}(f(\x_i)))\;.
\end{equation}
$\lambda > 0$ is a trade-off parameter. $\s_{\tau}$ transforms the confidence for all classes into a $C$-class posterior probability.
% \begin{equation}\label{eq:logit}
$\s_{\tau}(f(\x_i)) = \mathbf{softmax}(\frac{f(\x_i)}{\tau})$.
% \end{equation}
$\tau$ is a non-negative temperature, the larger the value of $\tau$, the smoother the output. 
$\mathcal{R}(\cdot)$ measures the difference between two distributions, \eg, the Kullback-Leibler (KL) divergence. In Eq.~\ref{eq:kd}, the student not only minimizes the mapping $f$ from an instance to its label over $\mathcal{D}$, but also aligns its predictions with the teacher on the same set of instances. Note that the student and the teacher could use different temperatures. 

In standard KD, both teacher $f_T$ and student $f$ target the same $C$ classes. Given the training data and $f_T$, richer supervision like the soft labels is incorporated when training $f$, which encodes the relationship between an instance and $C$ candidate classes. $f_T$ could be a deep neural network with a larger capacity~\cite{hinton2015kd,Chen2018Learning,Mirzadeh2019Improved}, which makes $f$ compact and discriminative.
$f_T$ could also be a certain generation of $f$ along with the whole training progress. Such self-distillation reduces the training cost and simultaneously enables sufficient training w.r.t. the vanilla strategy~\cite{FurlanelloLTIA18,Yang2018Snapshot}.
%Other forms of dark knowledge along the thread of instance-label mapping between the teacher and the student could be aligned in a similar manner, such as the hidden activation~\cite{RomeroBKCGB14} and the parameter flows~\cite{YimJBK17}.

\subsection{Cross-Task Knowledge Distillation}
The standard KD in Eq.~\ref{eq:kd} requires the student to be trained for the same labels $\mathcal{C}$ as the teacher, so that their classification results on the same instance could be matched. In a general scenario, it is necessary to borrow the learning experience from a {\em cross-task} teacher, \ie, $f_T$ has {\em a non-overlapping class set} $\mathcal{C}'$ and $\mathcal{C}'\cap\mathcal{C}=\emptyset$. 
Relaxing the requirement of the teacher enables KD in more applications.

\subsection{Generalized Knowledge Distillation}
GKD takes a further step given a teacher trained on a {\em general} class set $\mathcal{C}'$ --- we do not restrict $\mathcal{C}'$ to be non-intersected with the target classes $\mathcal{C}$. In other words, it could be either $\mathcal{C}'=\mathcal{C}$ (as in standard KD), $\mathcal{C}'\cap\mathcal{C}=\emptyset$ (as in cross-task KD), or even $\mathcal{C}'\cap\mathcal{C}\neq\emptyset$. In the third case, only parts of the target task's classes are related to classes $f_T$ trained for. 
The teacher provides additional supervision with $f_T$ in the training progress of $f$. However, due to the fact that the student is {\em agnostic} of which part of the teacher's supervision is related, \eg, the classes indexes of teacher's predictions that are overlapped with the target classes, it should identify and extract the helpful supervision from the teacher as much as possible instead of treating teacher's supervision uniformly.
For example, if $f_T$ is trained on generic animals, it could provide helpful supervision on animal-like instances in the current task but not man-made objects. If all classes in $\mathcal{C}'$ are distant from those in $\mathcal{C}$, we expect $f$ to perform as well as the one trained in the vanilla case without a teacher.
GKD helps build a visual recognition system efficiently. Specifically, we can leverage well-trained models from any tasks with different architectures to improve the discerning ability of the system without accessing their training data. 

\section{{\name} for Generalized KD}
\label{sec:method}
\begin{figure*}[t!]
	\centering
	\includegraphics[width=\linewidth]{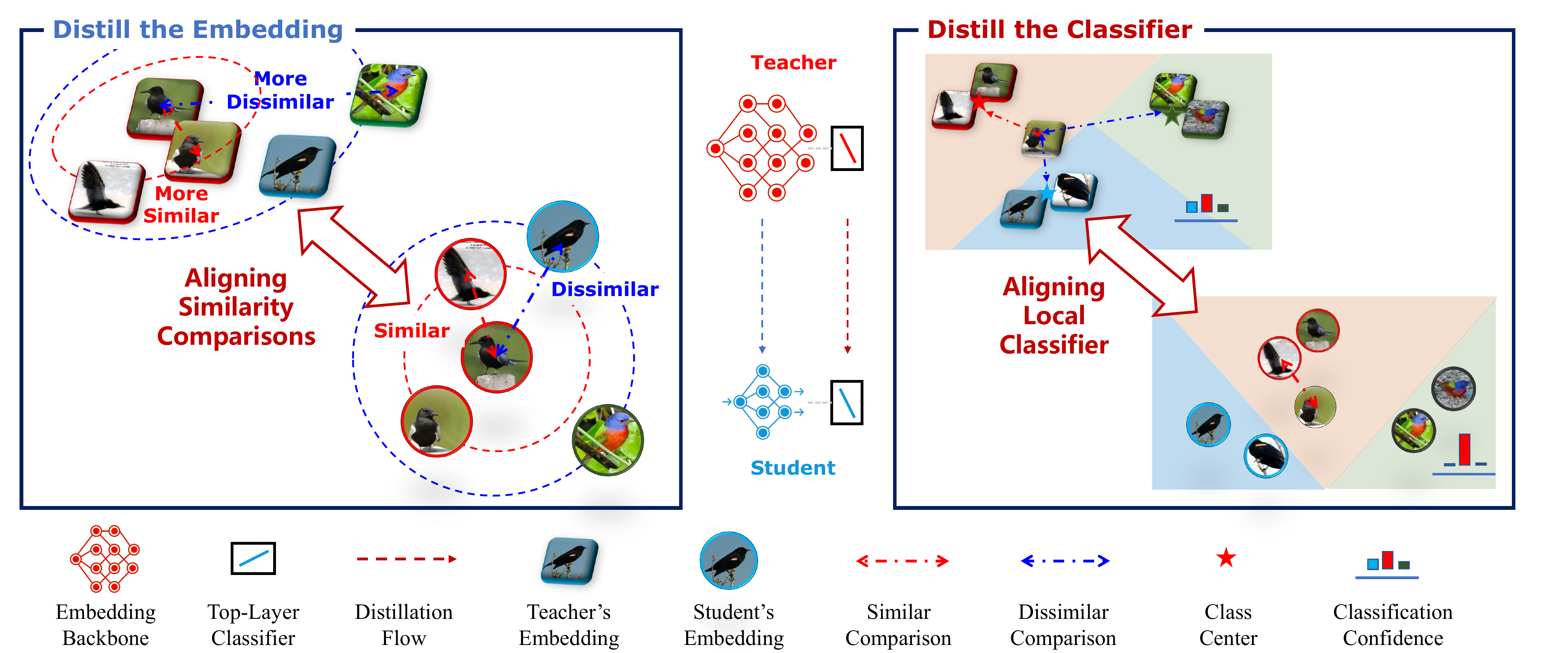}
	\caption{\small An illustration of the proposed \textbf{RE}lationship \textbf{F}ac\textbf{I}litated \textbf{L}ocal c\textbf{L}assifi\textbf{E}r \textbf{D}istillation~({\name}) approach. 
	The embedding and the top-layer classifier of the student distill the dark knowledge from the teacher's corresponding components, respectively. 
    First, the student organizes instances into tuples and aligns their similarity comparisons with the teacher (left plot), introducing richer supervision such as the similar and dissimilar levels of relationships. 
    {\name} then matches classification confidences between two models over instances in the target task. 
    We construct an embedding-based classifier with teacher's class prototypes (denoted by stars), which provides the posterior classification probability over both in-task and cross-task categories. 
	}
	\label{fig:overview}
\end{figure*}

We introduce the main idea of \textbf{RE}lationship \textbf{F}ac\textbf{I}litated \textbf{L}ocal c\textbf{L}assifi\textbf{E}r \textbf{D}istillation~({\name}) approach, followed by
analyses and discussions of its two stages.

\subsection{Decoupling the Distillation via \textbf{{\scshape ReFilled}}}
The two components in the target model $f$, \ie, the embedding $\phi$ and the top-layer classifier $W$, capture the correlations between instances and classes, respectively. 
In GKD, {\name} makes use of the characteristic of each part in $f$ and transfers the rich supervision from teacher to student by distilling knowledge for corresponding components.
The teacher's {\em comparison ability} does {\em not depend on concrete labels} and bridges the possible label gap between two models. Matching the relationship makes the distillation in {\name} be agnostic of classes. 
As in Figure~\ref{fig:overview}, we align instance-wise comparisons measured by $\phi$ with the teacher. The embedding distillation specifies how similar two objects are when training $\phi$, and makes $\phi$ as discriminative as the teacher.
After that, we construct a similarity-based classifier for target classes based on the teacher's embedding. We also derive a confidence-based criterion to identify helpful instance-wise supervision from the teacher adaptively, which extracts helpful supervision when training $f$ in GKD.

\subsection{Distill the Embedding}
Instance embedding $\phi(\x)$, the penultimate layer output of a deep neural network, encodes discriminative property of objects~\cite{Wen2016Center,He2018Triplet,Achille2018Emergence} without a direct dependence on labels~\cite{WeinbergerS2009Distance,AmidU15,SchroffKP15,SongXJS16,ManmathaWSK17,Hsu2018Learning}.
The instance-wise similarity computed based on their embeddings reveals whether two objects are similar or not, and how much they are similar. Therefore, similar instances are close to each other (with smaller distances) and dissimilar ones are far away. 
To distill the knowledge from a teacher with possible cross-task classes, we first focus on the transferable embedding, making the student's embedding as discriminative as the teacher. 

\subsubsection{Comparison Matching} 
Based on class semantics, denote two instances are similar if they come from the same class, and they are dissimilar if they have different labels. The distance between a pair of instances $(\x_i, \x_j)$ based on the embedding function $\phi$ is $\mathbf{D}_{\phi}(\x_i, \x_j)=||\phi(\x_i)-\phi(\x_j)||_2$. A good embedding makes embedding-based distances small for similar pairs and large for dissimilar ones. We formulate the similarity relationship between $\x_i$ and others into a {\em tuple}, \ie, 
\begin{equation}
(\x_i,\; \x^\mathcal{P}_i,\; \x^\mathcal{N}_{i1},\; \ldots,\; \x^\mathcal{N}_{iK})\;,\label{eq:tuple}
\end{equation}
which contains one similar positive neighbor $\x^\mathcal{P}_i$ w.r.t. the anchor $\x_i$ and $K$ dissimilar negative impostors $\{\x^\mathcal{N}_{i1}, \ldots, \x^\mathcal{N}_{iK}\}$.\footnote{We assume the same $K$ for different instances for simplicity, and $K$ could vary across tuples automatically. See Section~\ref{sec:construction} for details.}
We transform the similarity in the tuple into a probability with softmax operator, which reveals how much the anchor is close to its target neighbor than those impostors:
\begin{align}
p_{i}(\phi) = \s_{\tau}&\left(\left[-\mathbf{D}_{\phi}\left(\x_i,\; \x^\mathcal{P}_i\right),\;\right.\right.\label{eq:prob_neighbor}\\
&\quad\left.\left.-\mathbf{D}_{\phi}\left(\x_i,\; \x^\mathcal{N}_{i1}\right),\;\ldots,\; -\mathbf{D}_{\phi}\left(\x_i,\; \x^\mathcal{N}_{iK}\right)\right]\right)\;.\notag
\end{align}
Eq.~\ref{eq:prob_neighbor} measures the relative instance-wise similarities. The closer a neighbor or an impostor with $\x_i$, the larger the corresponding element in $p_{i}(\phi)$. For example, if the target neighbor $\x^\mathcal{P}_i$ has a very small distance with $\x_i$, then $p_{i}(\phi)$ becomes a one-hot label with only the first element equals 1.
In the vanilla scenario, we minimize the distance between the anchor with the target neighbor and push all impostors away based on the ``similar or not'' binary supervision~\cite{WeinbergerS2009Distance,SchroffKP15,SongXJS16,ManmathaWSK17}, which is the same as the case to minimize the discrepancy between $p_{i}(\phi)$ with the $K+1$ dimensional ground-truth probability $[1,0,0,\ldots,0]$. 

In {\name}, we take advantage of the ``dark knowledge'' in $p_{i}(\phi)$ with richer similarity comparison information --- the similar degree of an instance $\x_i$ with a positive and multiple negative candidates are characterized in detail. 
We improve the discriminative ability of the student's embedding $\phi$ by distilling the tuple comparison knowledge from the teacher, \ie, we minimize the KL-divergence over all tuples:\footnote{One instance $\x_i$ may have multiple comparison tuples with different target neighbors. We use one $p_{i}(\phi)$ to denote them for notation simplicity in the summation. Details are in Section~\ref{sec:construction}.}
\begin{equation}
\min_{\phi} \sum_{i} \mathbf{KL} \left(p_{i}(\phi_T) \;\big\|\; p_{i}(\phi)\right)\;.\;\label{eq:obj1}
\end{equation}
Eq.~\ref{eq:prob_neighbor} describes the fine-grained differences measured by the teacher inside tuples. By aligning comparisons in Eq.~\ref{eq:obj1}, the student is expected to be able to compare instances as well as the teacher.
For example, the student treats a flying ``black tern'' and a ``red-winged blackbird'' as dissimilar in the vanilla scenario and pushes their embeddings apart. The teacher's tuple similarity vector $p_i(\phi_T)$ may indicate that a flying ``red-winged blackbird'' is more similar to a flying ``black tern'' than a ``black tern'' sipping the water, then the student can benefit from the richer supervision through Eq.~\ref{eq:prob_neighbor}. 
In experiments, we show that if the comparison ability of the teacher matches the student's task, \eg, both on fine-grained animals, comparison matching makes $\phi$ discriminative even with limited examples. 
Additionally, the heterogeneity between teacher and student, \eg, different scales or dimensions between $\phi_T$ and $\phi$, will not influence the matching in Eq.~\ref{eq:obj1}. Therefore, it facilitates knowledge transfer across different architectures.

We can rethink Eq.~\ref{eq:prob_neighbor} from a retrieval perspective. Given the query instance $\x_i$, we'd like to find its most similar neighbor in $(\x^\mathcal{P}_i, \x^\mathcal{N}_{i1}, \ldots, \x^\mathcal{N}_{iK})$ with the smallest distance. Eq.~\ref{eq:prob_neighbor} encodes the probability to query each candidate instance. Eq.~\ref{eq:obj1} matches a {\em soft} version of the anchor query probability between teacher and student, which promotes the retrieval ability of the student as effective as the teacher.

\subsubsection{Construction of the comparisons}\label{sec:construction}
Tuples in the form of Eq.~\ref{eq:tuple} contain instances from the target task, and we construct ``semi-hard'' tuples for comparison matching.
In particular, for each instance $\x_i$ in the mini-batch, we first look for every $\x^\mathcal{P}_i$ from the same class as its neighbors. Then for each pair of $(\x_i,\x^\mathcal{P}_i)$, we enumerate those impostors $\x^\mathcal{N}_{i}$ with different labels to $\x_i$ but have larger distances than $\mathbf{D}_{\phi}\left(\x_i, \x^\mathcal{P}_i\right)$~\cite{SchroffKP15}. 
The impostors increase the difficulty of comparing instances and avoid embedding collapse. The distance is measured based on the $\ell_2$-normalized embeddings with the current $\phi$ during the optimization progress of Eq.~\ref{eq:obj1}. 
The impostor number $K$ in Eq.~\ref{eq:tuple} is determined by the ``semi-hard'' impostors in a mini-batch, which varies for $(\x_i,\x^\mathcal{P}_i)$.
We abbreviate multiple tuples for $\x_i$ with one $p_{i}(\phi)$ in Eq.~\ref{eq:obj1}.
In summary, if the student finds tuples are hard to evaluate, it will ask the teacher for help about the concrete measures of the similarity levels.

\subsection{Distill the Classifier Adaptively}
Benefited from the distilled embedding for instance-instance comparisons, {\name} further distills the knowledge from the teacher for instance-label classification. A similarity-based classifier is constructed with the help of the teacher's embedding to guide the update of the student's classifier. An adaptive weight is derived from the teacher's confidence to filter out its unhelpful supervision for GKD.

\subsubsection{A General Approach for Classifier Distillation}
We propose a general approach to distill the discerning ability of the top-layer classifier. A Nearest Class Mean~(NCM) classifier~\cite{Mensink2013Distance,Snell2017Proto,Ye2018Learning} is constructed based on the teacher's embeddings $\phi_T$, which captures the instance-label relationship for categories in both previous and target tasks without requiring the teacher to share the same label space.

\noindent{\bf Embedding-based classifier for GKD.}
With the teacher's embeddings $\phi_T(X)\in\mathbb{R}^{N\times d}$ on $X$, we compute the embedding center of all $C$ classes in the target task by
\begin{equation}\label{eq:prototype}
P = {\rm diag}(\bm{1}\oslash (Y^\top\bm{1}))Y^\top \phi_T(X) \in \mathbb{R}^{C\times d}\;.
\end{equation}
$\oslash$ denotes the element-wise division. Each row $\p_c\in\mathbb{R}^d$ of $P$ corresponds to the center of the $c$-th class. For any instance $\x$ in the target task, we can determine its label based on its similarity with the $C$ centers:
\begin{equation}
p_T(c \mid \x) = \frac{\left(\left(\phi_T(\x)^\top \p_c\right)\;/\;\|\p_c\|_2\right)}{\sum_{c'=1}^C \left(\left(\phi_T(\x)^\top \p_{c'}\right)\;/\;\|\p_{c'}\|_2\right)}\;.\label{eq:NCM}
\end{equation}
The larger the cosine similarity between an instance embedding $\phi_T(\x_i)$ to the $c$-th class center $\p_c$ in the teacher's embedding space, the larger the posterior probability $p_{T}(c \mid \x)$ of class $c$. 
The similarity-based classifier takes advantage of the discriminative embedding of the teacher and could be applied for the target task even $\phi_T$ is trained from non-overlapped label spaces. Thus, we use Eq.~\ref{eq:NCM} to bridge the possible cross-task label gap in GKD, which reveals the relationship between an instance and multiple classes. 

\noindent{\bf Local Knowledge Distillation~(LKD).}
We incorporate the instance-label relationship indicated by the teacher's embedding in Eq.~\ref{eq:NCM} into the training progress of the student's top-layer classifier $W$. 
In particular, we match the student's prediction with the teacher through a {\em local} KD term. If the set of classes in the sampled mini-batch is $\mathcal{S}$ and $\mathcal{S}\subseteq\mathcal{C}$, then only the posterior over $\mathcal{S}$ are considered.
\begin{align}\label{eq:obj2}
\min_{f}\sum_{i=1}^N\ell\left(f(\x_i),\;\y_i\right)& + \lambda\mathbf{KL}\big(\hat{p}_{T}(\mathcal{S}\mid\x_i)\;\|\; \hat{\s}_{\tau}(f(\x_i))\big)\\
\hat{p}_{T}(\mathcal{S}\mid\x_i)&=\mathbf{softmax}(\{p_T(s|\x)\}_{s\in\mathcal{S}})\notag\\
\hat{\s}_{\tau}(f(\x_i))&=\mathbf{softmax}(\{\w_s^\top\phi(\x)/ \tau\}_{s\in\mathcal{S}})\notag\;.
\end{align}
In the second term of Eq.~\ref{eq:obj2}, rather than aligning two model's confidences of all target classes $\mathcal{C}$, only {\em partial posteriors} of classes in $\mathcal{S}$ are matched. 
LKD is not only efficient but also effective in emphasizing the difference between classes when the class number $C$ is large (analyses are in section~\ref{lkd_discuss}). 

\subsubsection{Adaptively Weighted LKD} 
In GKD, a teacher may be trained from labels only partially overlapped with the target classes, so uniformly matching the student's predictions with the teacher's predictions via Eq.~\ref{eq:obj2} is not optimal. 
If a target class $c\in\mathcal{C}$ is related to a particular class in $\mathcal{C}'$ that the teacher trained on, the teacher may provide more precise estimations on $p(c\mid\x)$ since it is more familiar with instances in class $c$.
If all classes in $\mathcal{C}'$ are not related to $c$, the teacher may hesitate to provide a confident and helpful instance-label relationship estimation using Eq.~\ref{eq:NCM}. 
Thus, we provide an {\em adaptive} weight for LKD to emphasize the teacher's guidance over more related instances, which makes the student extract helpful supervision from the teacher.
In other words, the student relies on the teacher's supervision over instances from classes that overlapped with the teacher's label space while weakening the teacher's guide for those ``novel'' classes w.r.t. the teacher.

Particularly, we measure the helpfulness of a teacher based on its prediction confidence. Define the ``pseudo'' label of an instance as the class index of the maximum confidence, \ie, $\hat{s} = \argmax_{s}\;(\{p_T(s|\x)\}_{s\in\mathcal{S}})$, the discrepancy between the teacher's prediction $\mathbf{softmax}(\{p_T(s|\x)\}_{s\in\mathcal{S}})$ with the ``pseudo'' label measures how much the teacher is confident of its supervision~\cite{chen2021Learning}. We set
\begin{equation}
\lambda_i = 2\lambda \times  \sigma\left(-\ell\left(\mathbf{softmax}(\{p_T(s|\x)\}_{s\in\mathcal{S}}),\; \hat{s}\right)\right),
\label{eq:adaptive_weight}
\end{equation}
which is an instance-specific weight for the distillation term.
We use the logistic function to transform the range of $\lambda_i$ to $[0, \lambda]$. 
Finally, we have the following objective:
\begin{equation}\label{eq:obj3}
\min_{f}\sum_{i=1}^N\ell\left(f(\x_i),\;\y_i\right) + \lambda_i\mathbf{KL}\big(\hat{p}_{T}(\mathcal{S}\mid\x_i)\;\|\; \hat{\s}_{\tau}(f(\x_i))\big).
\end{equation}
We detach the gradient of $\lambda_i$ during the optimization. $\lambda_i$ becomes larger when it is applied to those instances the teacher is familiar with and confident (with smaller loss values in Eq.~\ref{eq:adaptive_weight}), and has a smaller value when the teacher cannot provide strong supervision. In section~\ref{sec:GKD}, we verify such adaptive weight is able to differentiate the teacher's classes overlapped or different from the target ones. Another re-weight strategy based on the discrepancy between student's and teacher's predictions is also investigated in experiments. 

Eq.~\ref{eq:obj3} helps standard KD as well, where $f_T$ and $f$ have the same class set. 
Usually, $f_T$ is confident over most instances and almost all $\lambda_i$ values will be close to 1. Then Eq.~\ref{eq:obj3} degenerates to Eq.~\ref{eq:obj2}.
Otherwise, if there exist instances $f_T$ is unconfident, those instances may be noisy labeled or corrupted ones. In this case, a smaller $\lambda_i$ helps the student avoid being negatively affected by the teacher.

\subsection{Summary and Discussions of {\name}}
We decouple the distillation flow into two stages for embedding and the top-layer classifier, respectively. First, {\name} improves the discerning ability of the student's embedding by comparison matching in Eq.~\ref{eq:obj1}.
Then the classification confidence between teacher and student is aligned in a local manner in Eq.~\ref{eq:obj1}, where helpful supervision from the teacher is emphasized with adaptively weighted distillation for GKD. The main flow of {\name} is summarized in Alg.~\ref{alg:flow}. 
% We also consider a combined objective optimizing the two stages jointly, and detailed comparisons could be found in the experimental part.

\begin{algorithm}[t]
\begin{flushleft}
\caption{The Flow of {\name}.}
\label{alg:flow}
\hspace*{\algorithmicindent} \textbf{Input} Pre-trained Teacher's Embedding $\phi_T$. \\
\hspace*{\algorithmicindent} \textbf{Output} The target model $f = W\circ \phi$.
\end{flushleft}
\begin{algorithmic}{
\STATE \textbf{\textit{Distill the Embedding}}:
\FORALL{Iter = 1,...,MaxIter} 
\STATE {Sample a mini-batch $\{(\x_i,\y_i)\}$.}
\STATE {Generate tuples $\{(\x_i, \x^\mathcal{P}_i, \x^\mathcal{N}_{i1}, \ldots, \x^\mathcal{N}_{iK})\}$ with $\{\phi(\x)\}$.}
\STATE {Compute probability of tuples $p_{i}(\phi_T)$ as Eq.~\ref{eq:prob_neighbor}.}
\STATE {Optimizing $\phi$ by aligning comparisons in Eq.~\ref{eq:obj1}.}
\ENDFOR
			
\STATE \textbf{\textit{Distill the Classifier}}:
\STATE {Initialize $f$ with $\phi$.}
\STATE {Compute the instance-specific weight $\lambda_i$.}
\STATE {Optimizing $f$ with Eq.~\ref{eq:obj3}.}
			
% \STATE \textbf{\textit{Distill the Classifier (GKD Version)}}:
% \STATE {Initialize $f$ with $\phi$.}
% \STATE {Optimizing $f$ and $\hat{W}$ with Eq.~\ref{eq:obj3}.}
}
\end{algorithmic}
\end{algorithm}

\subsubsection{Discussions on Embedding Distillation} 
\noindent{\bf Why not a direct embedding matching?} 
One intuitive way to match the instance-wise relationship between teacher and student is to align their embeddings directly, \eg, minimizing the loss $\|\phi(\x) - \phi_T(\x)\|_2^2$ over all instances in the current task~\cite{Chen2018Learning,Gao2018Feature,Koratana2019LIT}. This constraint requires both models to have the same size of embeddings, which is too strong to satisfy, especially there exists an architecture gap between the two models.
An asymmetric map is learned to match embeddings with different dimensions in~\cite{Budnik2021Asymmetric}, whose distillation quality is influenced by the dimension difference between teacher and student. 
\cite{Liu2019Knowledge,Park2019Relational,Peng2019Correlation,Tung2019Similarity,Li2020Local} keep the embedding-based pairwise relationship (\eg, distances) between teacher and student have similar values. 
If two models measure similarity at different scales, regularizing teacher and student have same distances between pairs still has drawbacks. Even though the student already obtains the right similarity relationship, the teacher could wrongly adjust it due to their scale differences. 
Therefore in {\name}, we ask the teacher to provide its estimation about {\em relative comparisons} among instances in the form of {\em tuples} and require the student to align such relative similarity measure for discriminative embeddings. The high order comparisons between instances are utilized.

\noindent{\bf Another perspective on comparison matching.}
We illustrate the effect of comparison matching in Eq.~\ref{eq:obj1} in a special case with one negative impostor ($K=1$). More analyses for $K>1$ are in the supplementary.
With only one impostor in the tuple, we can simplify the term $p_i(\phi)$ in Eq.~\ref{eq:prob_neighbor} as $\sigma(\mathbf{D}_{\phi}\left(\x_i, \x^\mathcal{N}_{i}\right)-\mathbf{D}_{\phi}\left(\x_i, \x^\mathcal{P}_i\right))$, where $\sigma(x)=1/(1+e^{-x})$ is the logistic function squashing the input into $[0,1]$.

Define $\rho_{i}=1-p_{i}(\phi_T)$ and the logistic loss $\iota(x) = \ln(1+\exp(-x))$, we can reformulate Eq.~\ref{eq:obj1} as
\begin{align}
&\mathbf{KL}\left(p_{i}(\phi_T) \;\|\; p_i(\phi)\right)\;\cong\;\;\iota\left(\mathbf{D}_{\phi}(\x_i,\x^\mathcal{N}_i) - \mathbf{D}_{\phi}(\x_i,\x^\mathcal{P}_i)\right)\notag\\
&\;\qquad\qquad\qquad+\;\rho_{i}\left(\mathbf{D}_{\phi}(\x_i,\x^\mathcal{N}_i) - \mathbf{D}_{\phi}(\x_i,\x^\mathcal{P}_i)\right).\label{eq:expansion}
\end{align} 
$\cong$ neglects constants in the derivation. 
The first term in Eq.~\ref{eq:expansion} with loss $\iota$ forces the distance $\mathbf{D}_{\phi}(\x_i,\x^\mathcal{N}_i)$ between dissimilar instances larger than the distance $\mathbf{D}_{\phi}(\x_i,\x^\mathcal{P}_i)$ between similar ones, so that the distance with $\phi$ is matched with the similarity relationship indicated by the tuple. Minimizing the first term equals vanilla embedding learning, making same-class instances close and different-class ones apart.

An additional term in Eq.~\ref{eq:expansion} is introduced in comparison matching, which makes the anchor's distance between neighbor and impostor close. The strength $\rho_i$ on the second term indicates the teacher's confidence over the given triplet. 
If $\mathbf{D}_{\phi_T}(\x_i,\x^\mathcal{P}_i)$ is smaller than $\mathbf{D}_{\phi_T}(\x_i,\x^\mathcal{N}_i)$, \ie, the teacher measures the target neighbor much similar w.r.t. the anchor than the impostor, then $p_{i}(\phi_T)$ is large and $\rho_{i}$ is small. In this case, minimizing Eq.~\ref{eq:expansion} emphasizes the first term, which makes the comparison matching the same as the vanilla embedding learning. 
In contrast, if the similarity relationship provided by the tuple is not very consistent with the one measured by the teacher, then $\rho_{i}$ becomes large. For example, given (a) cat, (b) tiger, (c) bear, we will treat both (a, b) and (a, c) as dissimilar ones since they come from different classes. A well-trained $\phi_T$ may discover the relatedness between (a) and (b), so it would use less force to push (a, b) apart than that for (a, c).
Therefore, large $\rho_{i}$ weakens $\iota$ with an opposite objective in the additional term. 

In summary, different from the binary label (``similar'' or ``dissimilar'') indicated by the tuple, comparison matching {\em rectifies the strength} when minimizing (resp. maximizing) the distance between similar (resp. dissimilar) pairs based on the teacher's estimation with $\phi_T$.

\begin{figure}[t]
	\centering
	\includegraphics[width=0.9\linewidth]{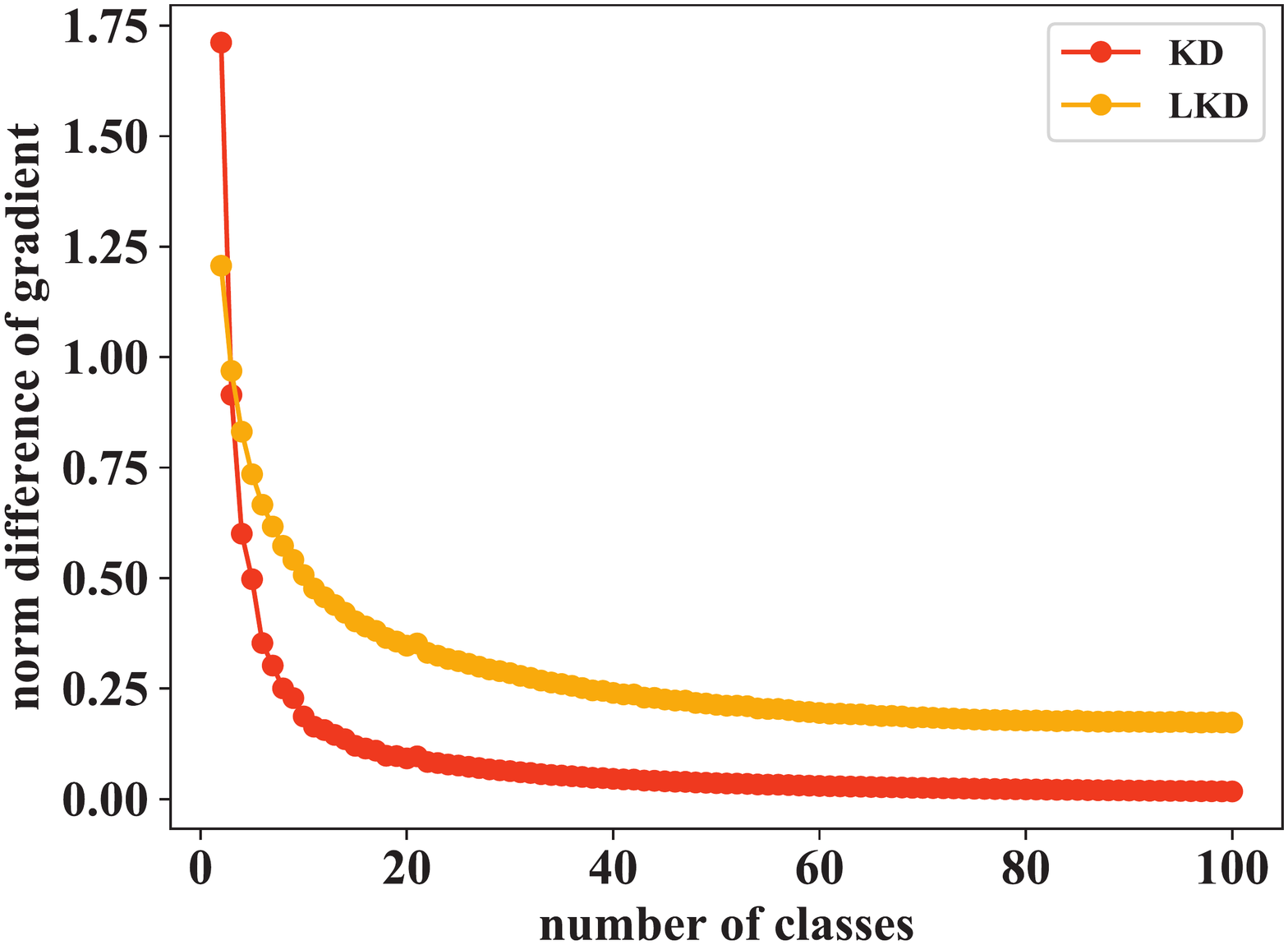}\\
	\caption{\small The averaged norm differences between the vanilla cross-entropy loss in Eq.~\ref{eq:vanilla_train} and the KD objective in Eq.~\ref{eq:kd} over the gradient of all top-layer classifiers, \ie, $\mathbf{mean}_c\|\frac{\partial O_\mathbf{ce}}{\partial \w_c}-\frac{\partial O_\mathbf{kd}}{\partial \w_c}\|_2$ is shown in red. Its LKD counterpart of the average gradient between Eq.~\ref{eq:vanilla_train} and Eq.~\ref{eq:obj3} is shown in yellow. When the number of classes in the target task grows, the norm difference based on the KD objective decreases fast, which indicates weaker additional supervision introduced by the distillation term. However, such a decrease is mitigated with LKD loss.}
	\label{fig:norm}
\end{figure}

\subsubsection{Discussions on LKD}\label{lkd_discuss}
We analyze the effectiveness of KD by its gradient over the top-layer classifier $W\in\mathbb{R}^{d\times C}$. Without loss of generality, we take the gradient of $\w_c$ over {\em one single} instance $\x$ as an example, whose target label is $c$. Denote $p_c$ and $q_c$ as the $c$-th element in the student's and teacher's normalized prediction $\mathbf{softmax}(f(\x))$ and $\mathbf{softmax}(f_T(\x))$ (the posterior probability of the $c$-th class given the student's and the teacher's embedding of the instance), respectively.
In vanilla learning scenario in Eq.~\ref{eq:vanilla_train} with objective $O_\mathbf{ce}$, the gradient w.r.t. $\w_c$ is 
\begin{equation}\label{eq:xent_gradient}
\frac{\partial O_\mathbf{ce}}{\partial \w_c} = \left[-p_c(1-p_c)\right]\phi(\x)\;.
\end{equation}

With KD in Eq.~\ref{eq:kd} (denote its objective as $O_\mathbf{kd}$), the gradient over the classifier $\w_c$ of the $c$-th class is:
\begin{equation}\label{eq:gradient}
\frac{\partial O_\mathbf{kd}}{\partial \w_c} = \sum_\x \left[-p_c + \sum_{c'=1}^C p_{c'}q_c\right]\phi(\x)\;.
\end{equation}
When considering the soft supervision from the teacher in KD, not only the instance from the target class but also those from helpful related classes (the ones with large $p_{c'}$) will be incorporated to guide the update of the classifier. Since the summation in Eq.~\ref{eq:gradient} is computed over all $C$ classes, the normalized class posterior $q_c$ becomes small if $C$ is large, so that the helpful class instance will not be stressed obviously. Therefore, we consider a {\em local} version of the knowledge distillation term LKD in Eq.~\ref{eq:obj2}, where only the class set $\mathcal{S}$ in the current mini-batch are considered, \ie, the influence of a helpful related class selected by the teacher will be better emphasized in the update of $\w_c$.

We empirically verify this claim on CIFAR-100. We compute the gradient difference between the vanilla Cross-Entropy~(CE) loss and the KD variants w.r.t. all top-layer classifiers, and further use the averaged norm over randomly sampled instances to measure the additional supervision introduced with KD variants. 
The smaller the norm difference, the weaker the additional supervision signal provided by the teacher.
Figure~\ref{fig:norm} plots the change of the norm difference when we increase the number of randomly sampled classes in a task from 2 to 100, and all the gradients are measured during the initial optimization of the model.
We find when the number of classes becomes larger, the norm of gradient difference between vanilla KD loss and CE loss decreases faster than that between LKD loss and CE loss. Thus the supervision made by the vanilla KD teacher is weakened more than the supervision made by the LKD counterpart.

\section{Experiments}
\label{sec:exp}
We verify {\name} on a variety of classification tasks, namely GKD, standard KD, one-step incremental learning, and few/middle-shot learning.
We analyze the results first, followed by ablation studies and visualizations in each part. The code is available at \url{https://github.com/njulus/GKD}.
\begin{figure}[t]
	\centering
	\includegraphics[width=\linewidth]{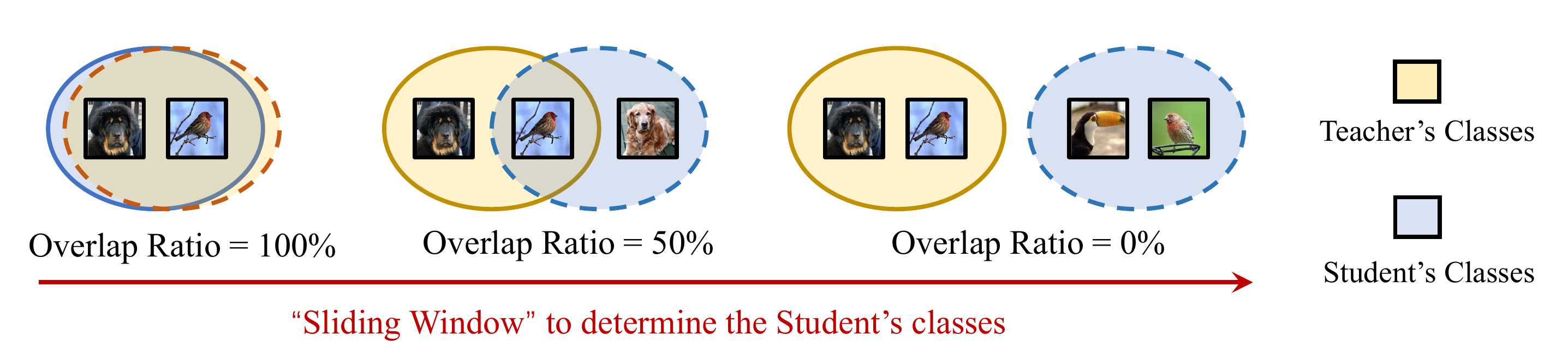}\\
	\caption{\small An illustration of the ``sliding windows'' to generate different configurations of teacher's and student's classes.}
	\label{fig:setup}
\end{figure}

\begin{figure*}[t]
	\begin{subfigure}[b]{0.24\linewidth}
		\centering
		\includegraphics[width=\linewidth]{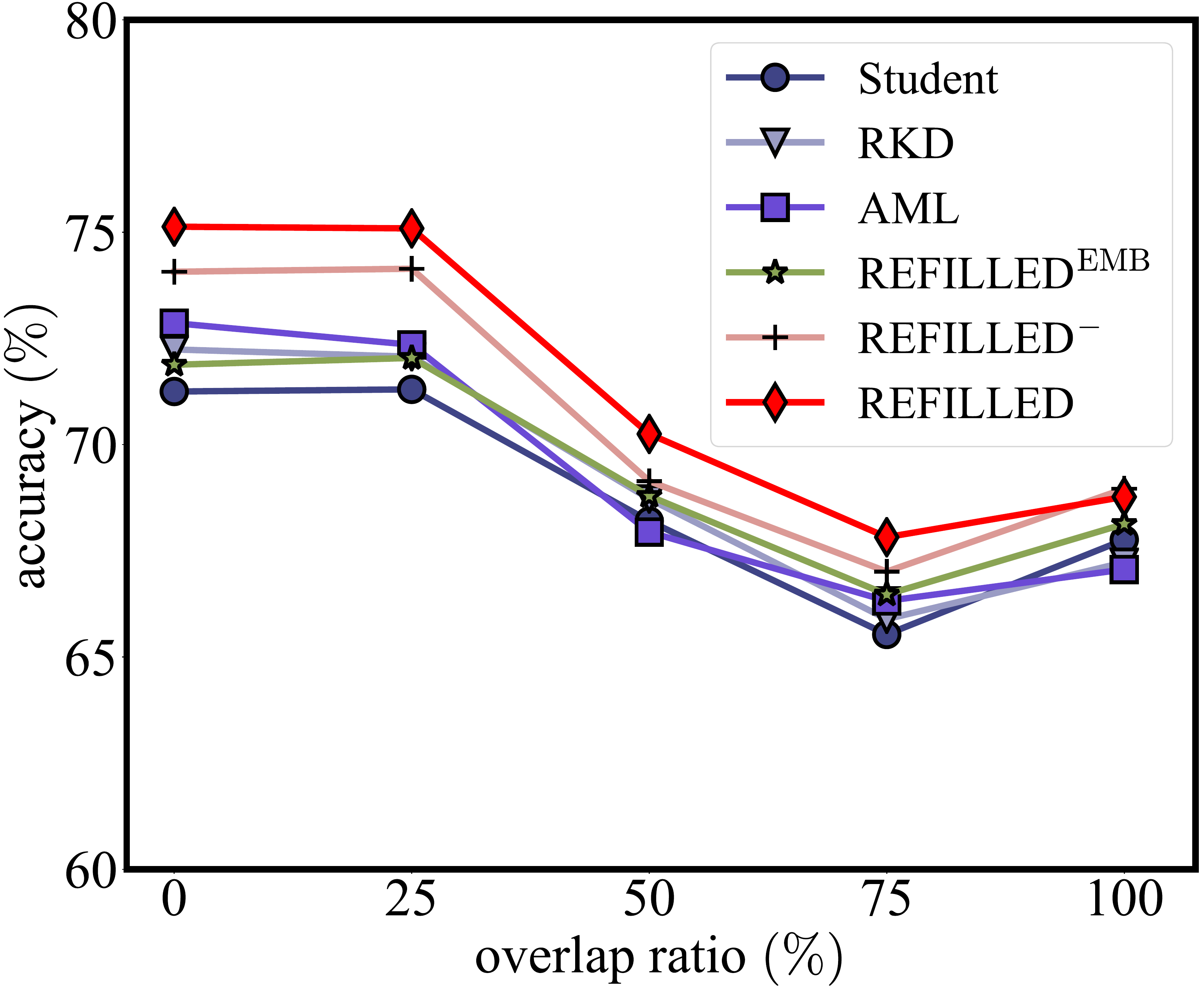}
		\caption{\small CUB: Student=1}
	\end{subfigure}
	\begin{subfigure}[b]{0.24\linewidth}
		\centering
		\includegraphics[width=\linewidth]{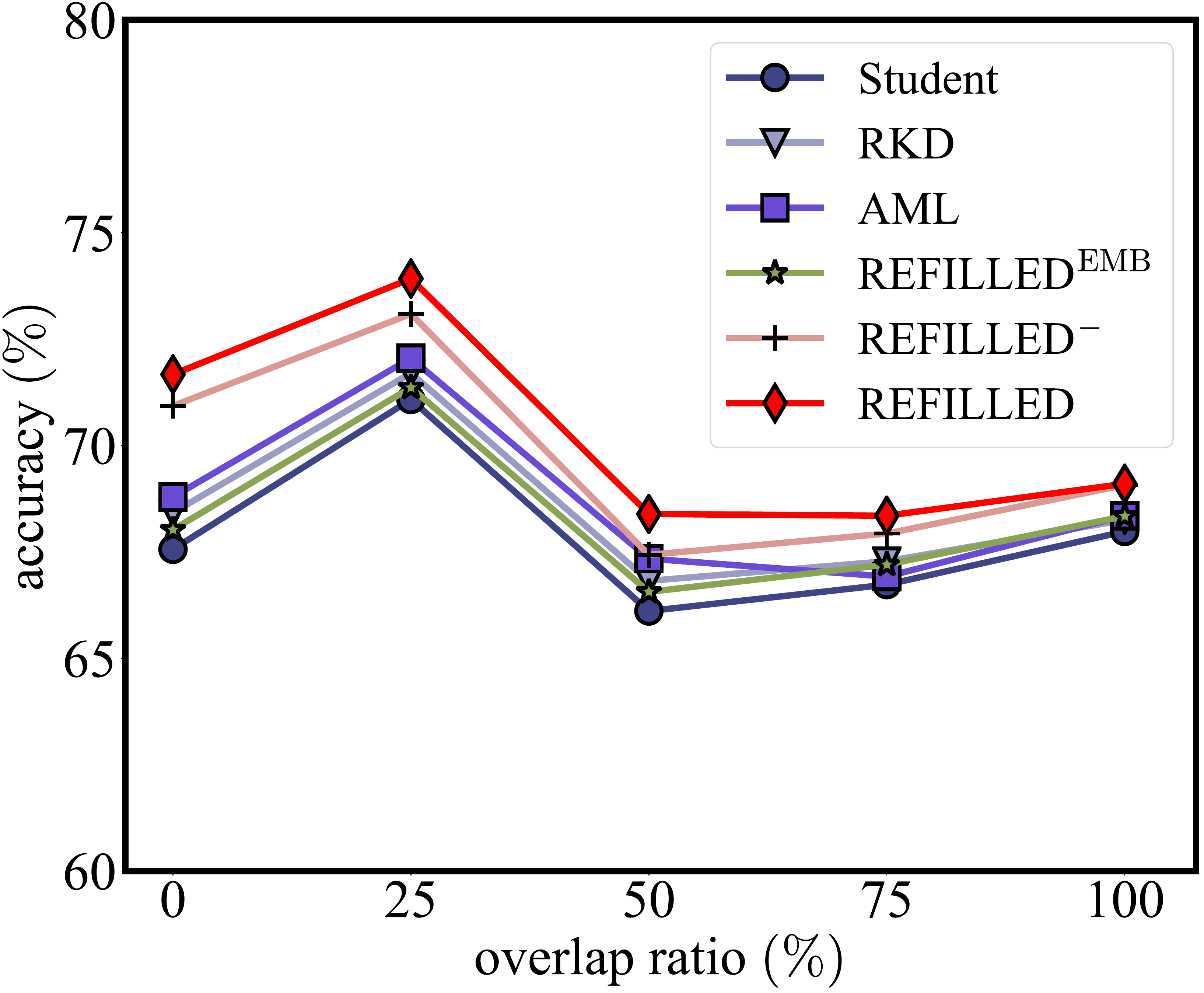}
		\caption{\small CUB: Student=0.75}
	\end{subfigure}
	\begin{subfigure}[b]{0.24\linewidth}
		\centering
		\includegraphics[width=\linewidth]{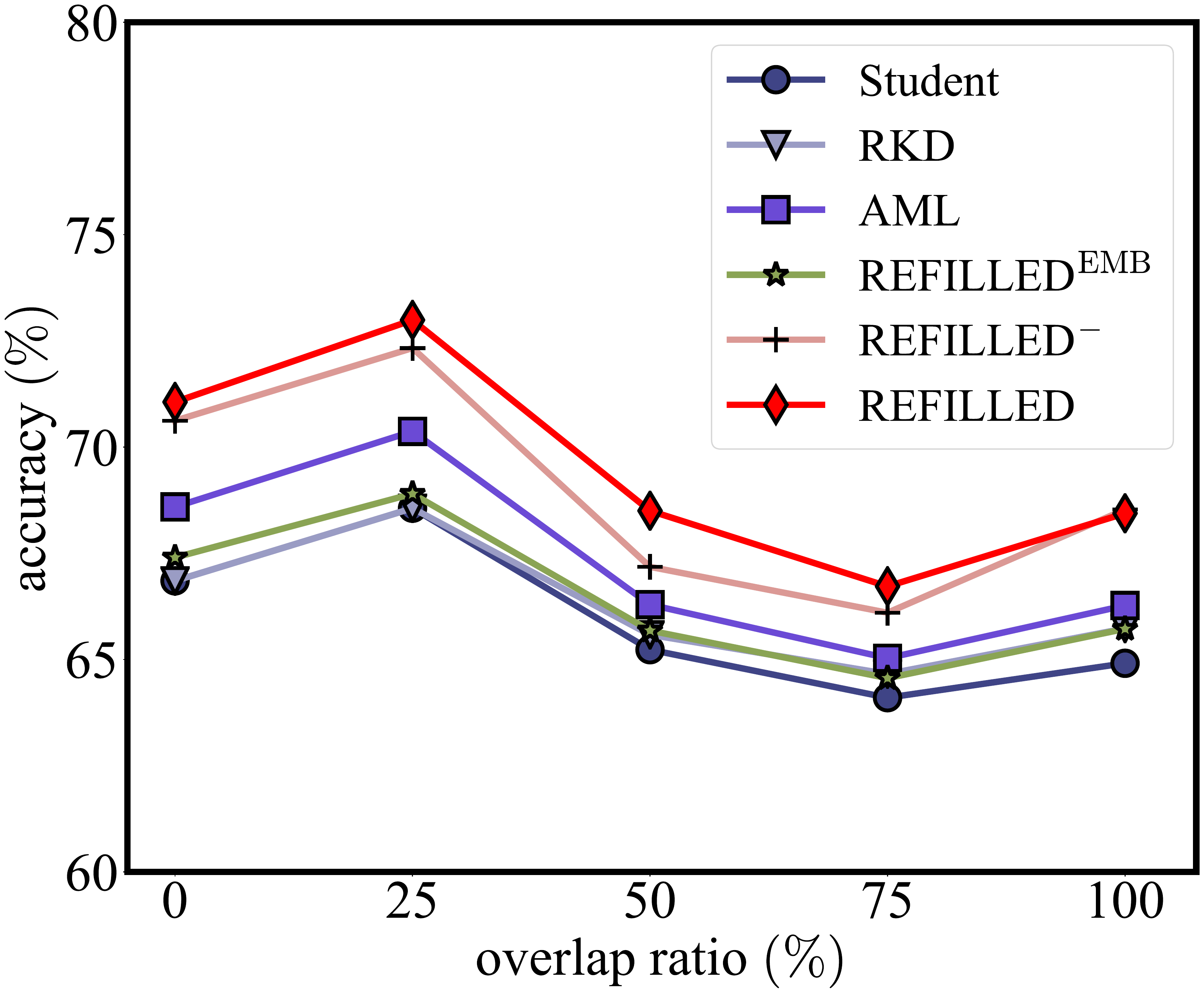}
		\caption{\small CUB: Student=0.5}
	\end{subfigure}
	\begin{subfigure}[b]{0.24\linewidth}
		\centering
		\includegraphics[width=\linewidth]{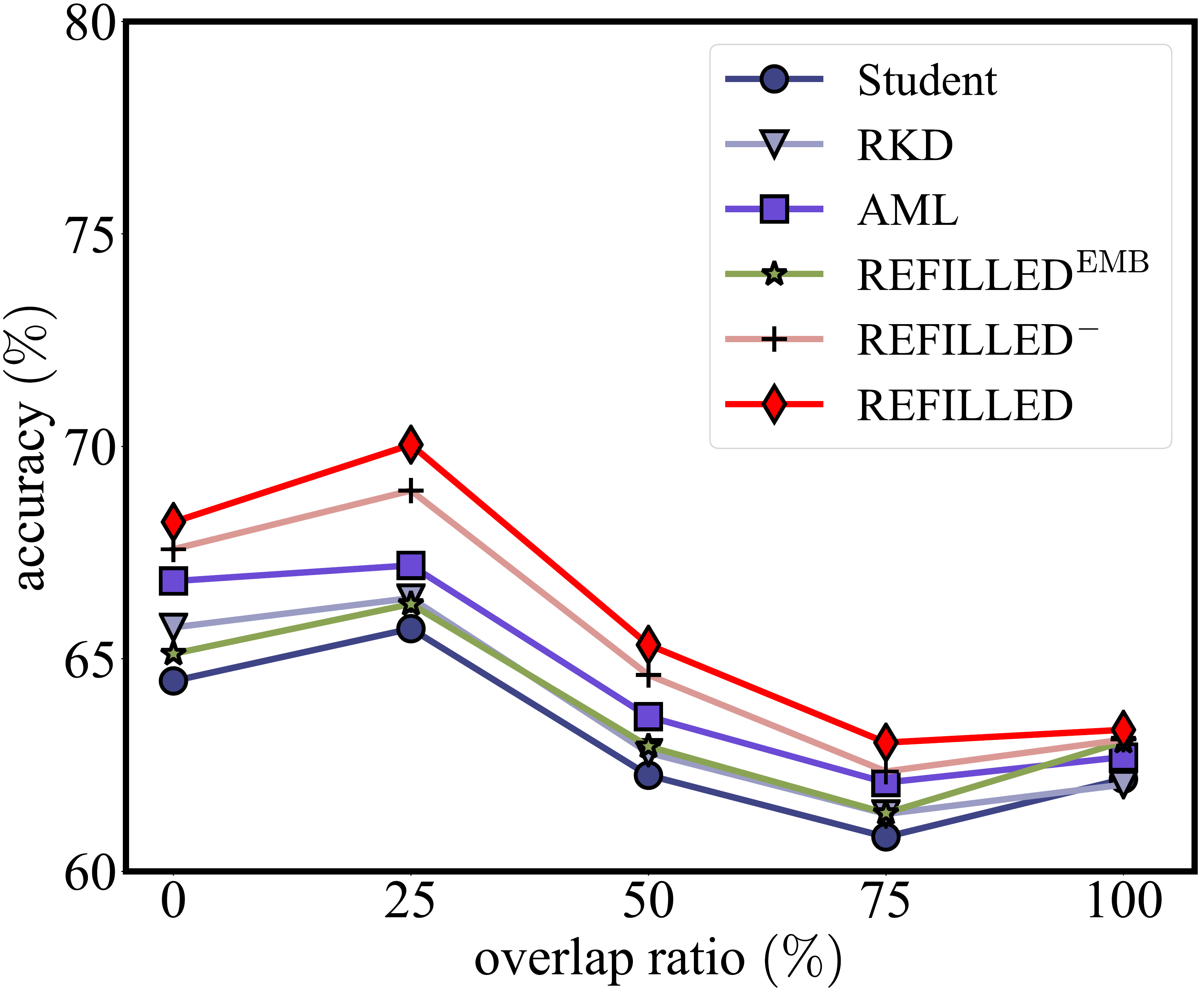}
		\caption{\small CUB: Student=0.25}
	\end{subfigure}
	
	\begin{subfigure}[b]{0.24\linewidth}
		\centering
		\includegraphics[width=\linewidth]{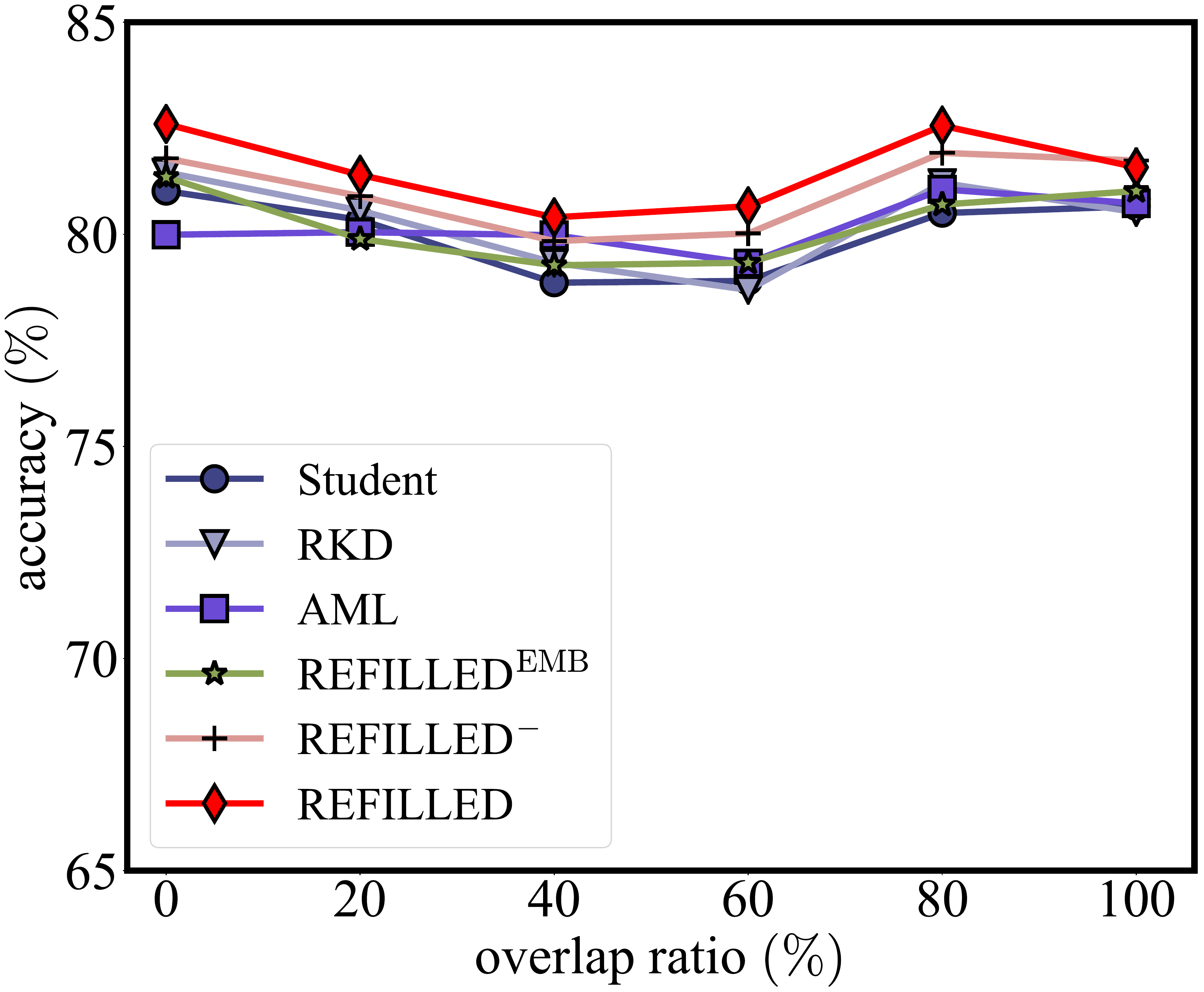}
		\caption{\small CIFAR: Student=(40,2)}
	\end{subfigure}
	\begin{subfigure}[b]{0.24\linewidth}
		\centering
		\includegraphics[width=\linewidth]{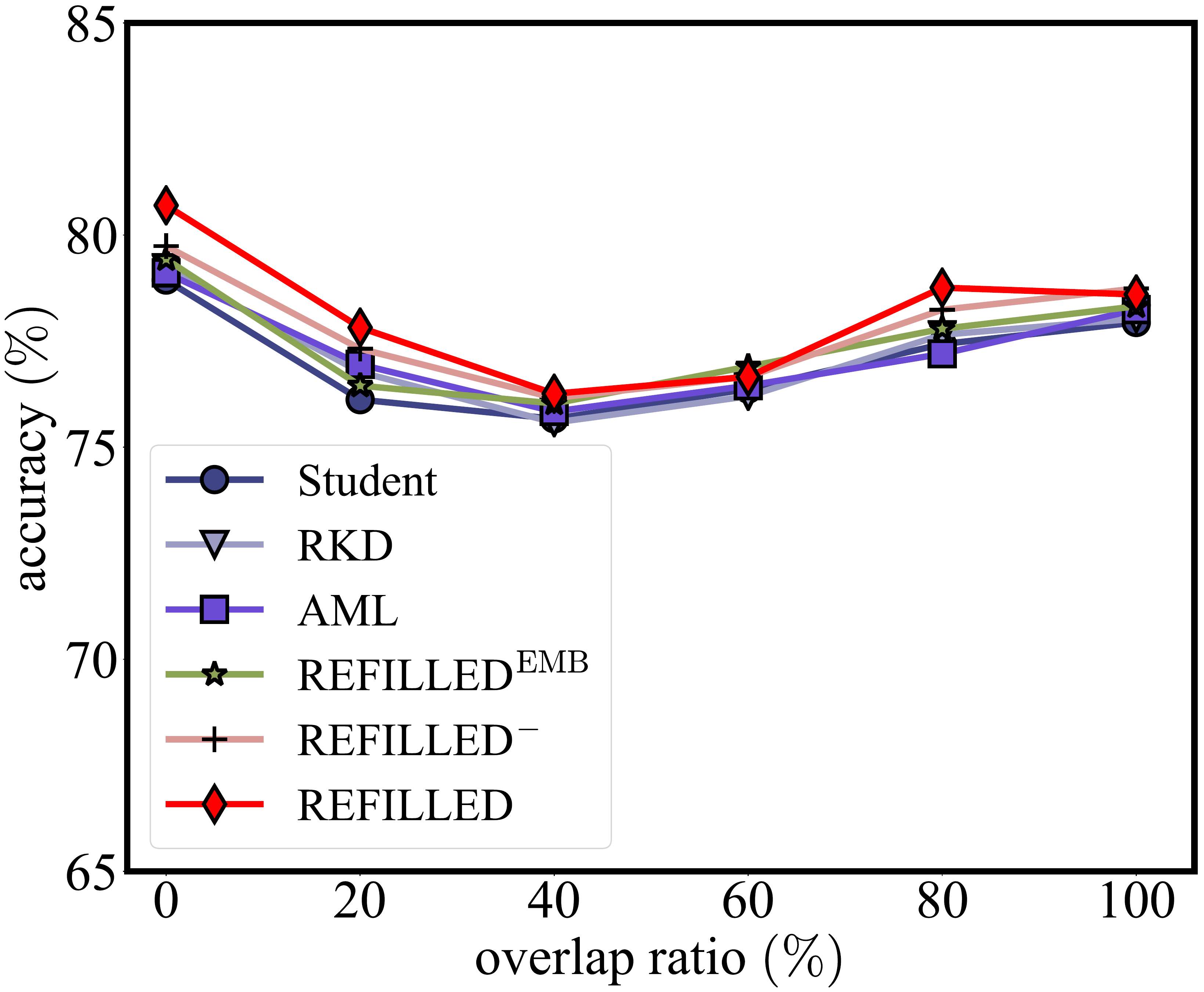}
		\caption{\small CIFAR: Student=(16,2)}
	\end{subfigure}
	\begin{subfigure}[b]{0.24\linewidth}
		\centering
		\includegraphics[width=\linewidth]{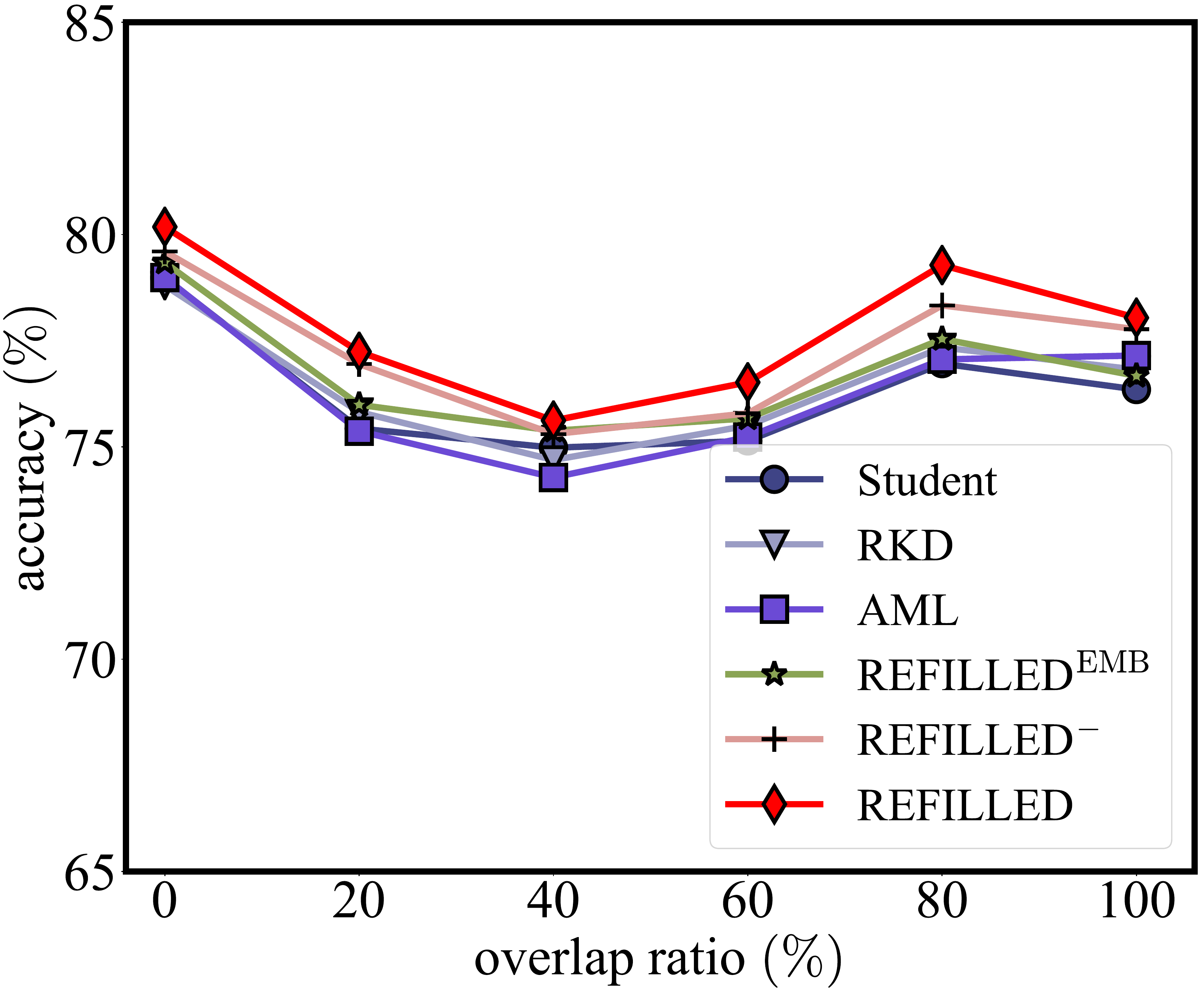}
		\caption{\small CIFAR: Student=(40,1)}
	\end{subfigure}
	\begin{subfigure}[b]{0.24\linewidth}
		\centering
		\includegraphics[width=\linewidth]{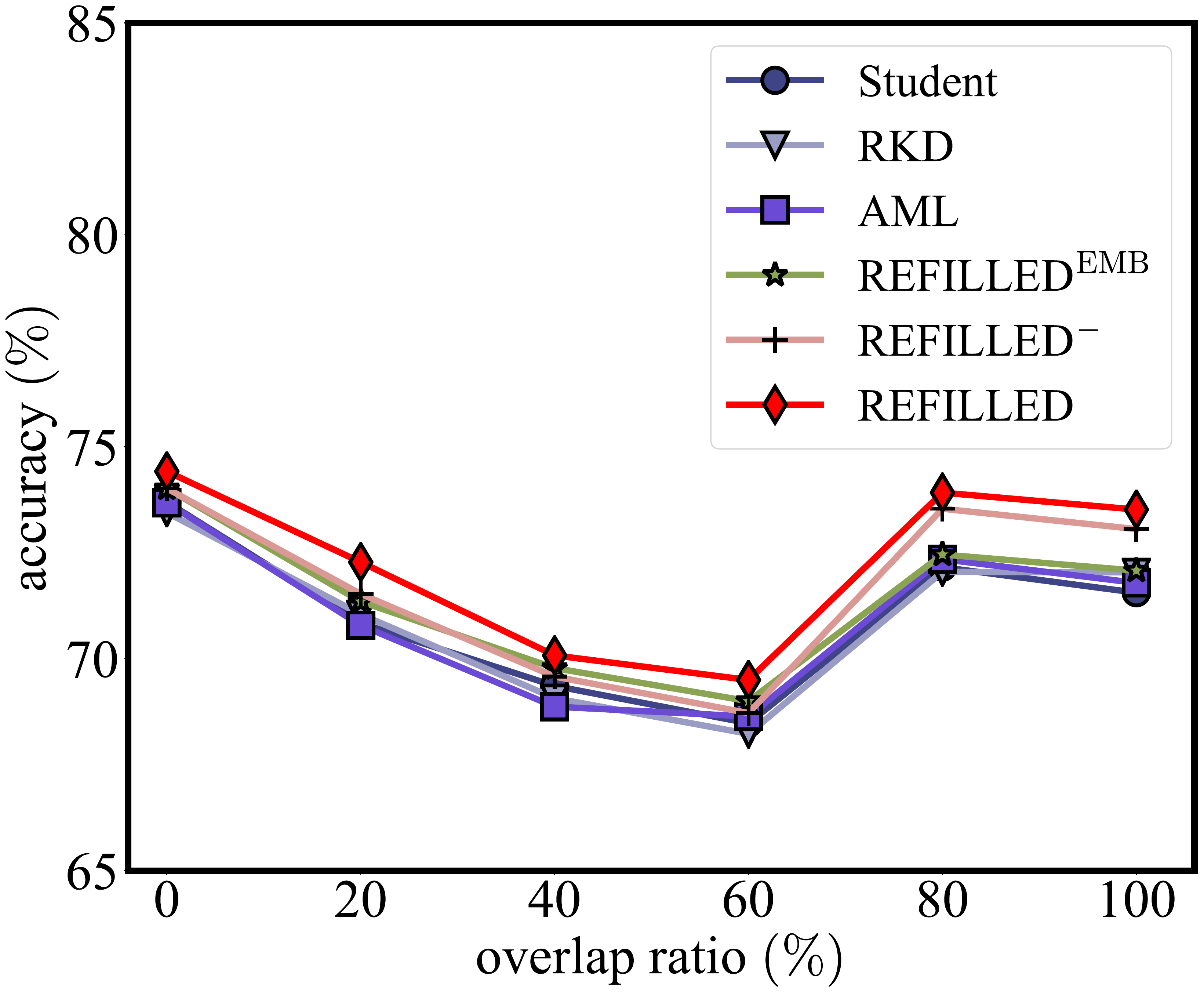}
		\caption{\small CIFAR: Student=(16,1)}
	\end{subfigure}
	\caption{\small The mean accuracy on GKD tasks upon CUB (upper) and CIFAR-100 (lower). The overlap ratio of student's label space w.r.t. teacher's changes from $0\%$ (cross-task KD) to $100\%$ (standard KD). The architecture of teacher is MobileNet-1.0 and WRN-(40,2) for CUB and CIFAR, respectively. We vary the architecture of the student. {\name}$^{\rm EMB}$ only distills the embedding from the teacher, and {\name}$^-$ does not utilizes the adaptive weights to select the helpful knowledge from the teacher.}
	\label{fig:cross_task_distillation}
\end{figure*}

\subsection{Generalized Knowledge Distillation}\label{sec:GKD}
{\name} can distill knowledge from a general teacher no matter how its label space overlapped or not w.r.t. the target classes. We evaluate {\name} for both cases.

\subsubsection{Setups}
\noindent{\bf Datasets.} \textbf{Caltech-UCSD Birds-200-2011~(CUB)}~\cite{wah2011caltech} is a fine-grained dataset with 200 different species of birds. 
We split two sets of the 100 classes as candidate class sets based on the given class indexes. Since classes in CUB are sorted in alphabetic order and classes with numerically close indexes are more similar, there is a relatively large semantic gap between these two sets.
We use the first 100 classes to train the teacher. For the student, we change its target classes over 100 classes with a ``sliding window'' from one candidate set to another (illustrated in Figure~\ref{fig:setup}). In other words, the student shares the same 100 classes with the teacher at first, and targets 100 non-overlapped classes at last. The student and the teacher have overlapped but not the same set of classes in the intermediate cases --- we investigate the cases when the student has a class overlap ratio $\{0\%, 25\%, 50\%, 75\%, 100\%\}$ with the teacher. In each 100-way classification task, both teacher and student use randomly sampled 70\% data in each class for training and the remaining instances for test. We do not use the attribute information of CUB. As a pre-processing, we crop all images based on the provided bounding boxes.
We also consider \textbf{CIFAR-100}~\cite{krizhevsky2009learning}, which contains 100 classes with 600 $32\times 32$ images per class. In each class, there are 500 images for training and 100 images for test.
We use a similar strategy as~\cite{Oreshkin2018TADAM} to split two 50-class sets, and the overlap ratio of the student's label space w.r.t. teacher's changes from $\{0\%, 20\%, 40\%, 60\%, 80\%, 100\%\}$.

\noindent{\bf Evaluations.}
We use the classification accuracy over the student's 100 classes on CUB (50 classes on CIFAR-100) as the criterion and evaluate whether the learned teacher can successfully help the student {\em no matter how many classes they share}. The average value over three random trials is reported.

\noindent{\bf Implementation details.}
Through minimizing the cross-entropy objective as Eq.~\ref{eq:vanilla_train}, we train a teacher model based on its corresponding training set with MobileNets~\cite{Howard2017Mobile} (width multiplier is $1.0$) and Wide ResNets~\cite{Zagoruyko2016Wide} (WRN, width is $2$ and depth is $40$) for CUB and CIFAR-100, respectively. We use the Stochastic Gradient Descent~(SGD) as the default optimizer, where the momentum is $0.9$, batch-size is $256$, maximum epoch number is $200$, initial learning rate is $0.1$, and we time the learning rate by $0.2$ after 50 epochs. We hold out a part of examples from the training set for validation, from which the best set of hyper-parameters are selected. With the best-selected hyper-parameters, we re-train the teacher model on the whole training set. 
During the training, we use the random crop together with the horizontal flip as the data augmentation. This is the same when training the student.
For CUB, we use different configurations of the MobileNets and adjust the model complexity with different width multipliers (complicated models have larger multipliers). While for CIFAR-100, we change the (depth, width) pair of the WRN.
There are two stages for the student. In both stages, the temperature $\tau$ of the teacher's model is set to 2, and we do not smooth the logits of the student. When distilling the embedding, we set momentum to $0.9$, batch-size to $256$, maximum epoch to $200$, initial learning rate to $0.1$, and we time the learning rate by $0.2$ after 50 epochs. While in the second stage, we use the same hyper-parameters. 
We tune $\lambda$ from the hold out validation set. We find the best $\lambda$ is close to $2$ and the performance of {\name} is not very sensitive to $\lambda$. Note that we construct an instance-specific weight ($\lambda_i$ in Eq.~\ref{eq:obj3}) based on this $\lambda$ value.

\noindent{\bf GKD baselines.} There are three types of baselines.
\begin{itemize}[leftmargin=*,noitemsep,topsep=0pt]
	\item \textit{Classification on the teacher's embedding}. 
	We construct classifiers for the target classes based on teacher's embeddings $\phi_T$ {\em only when teacher and student share the same architecture}.
	Based on this, we apply the nearest neighbor (1NN) classifier and the linear logistic regression (LR). 
	Besides, we fine-tune the teacher's model over instances in the student's split (FT). Since using a large learning rate will make the student obtain the same weights as training from scratch, we use a small initial learning rate (0.0001) and a fixed number of epochs (50) in our experiments.
	\item \textit{Variants of cross-task KD}. We compare our method with recent representative embedding-based KD approaches, \ie, the Relational Knowledge Distillation (RKD)~\cite{Park2019Relational} and Asymmetric Metric Learning (AML)~\cite{Budnik2021Asymmetric}. We then fine-tune the whole student model with its distilled embedding. Hyper-parameters are tuned in the same way as {\name}. 
	\item \textit{Variants of {\name}}. We investigate the importance of different components in {\name}. We consider fine-tuning the model with cross-entropy based on the embedding distilled by comparison matching in Eq.~\ref{eq:obj1}, which is denoted as ``{\name}$^{\rm EMB}$''. ``{\name}$^{\rm LKD}$'' means we train the student with cross-entropy and LKD from scratch without using the distilled embedding. ``{\name}$^-$'' denotes the {\name} variant without instance-adaptive weights $\lambda_i$.
\end{itemize}
\begin{table}[tbp]
	\centering
	\caption{\small Mean accuracy of student models on CUB. The overlap ratio of student's label space w.r.t. teacher's is fixed to $50\%$. The ``Teacher's EMB'' baseline only applies when it has the same architecture (same embedding dimension) with the student. 
	We set teacher to MobileNet-1.0 and vary the width multiplier of the student in $\{1,0.75,0.5,0.25\}$.
	}
	\begin{tabular}{@{}c|cccc@{}}
	\toprule
	Width Multiplier & 1     & 0.75  & 0.5   & 0.25  \\ \midrule
	Teacher's EMB          & \multicolumn{4}{c}{1NN: 43.29, LR: 51.66, FT: 61.78} \\
	Student          & 68.20  & 66.11 & 65.23 & 62.26 \\
	{\name}$^{\rm EMB}$    & 68.79 & 66.56 & 65.68 & 62.94 \\
	{\name}$^{\rm LKD}$  & 68.64 & 66.91 & 66.03 & 63.55 \\
	{\name}$^-$      & 69.14 & 67.43 & 67.18 & 64.62 \\
	\midrule
	{\name}          & \bf 70.25 & \bf 68.39 & \bf 68.50  & \bf 65.53 \\ \bottomrule
	\end{tabular}
	\label{tab:cross_task_ablation_cub}
\end{table}

\subsubsection{Results and Analyses on GKD}
The results of GKD in Figure~\ref{fig:cross_task_distillation} include standard KD (overlap ratio=$100\%$), cross-task KD (overlap ratio=$0\%$), and other general cases. The test accuracy of the student becomes higher when learning the task with more complicated models.
Points in Figure~\ref{fig:cross_task_distillation} denote fully different target datasets since the student possesses diverse subsets of 100 and 50 classes in CUB and CIFAR-100, respectively. 
The accuracy of the student indicates the difficulty of the target task, where the student gets low accuracy when the overlap ratio nears 75\% on CUB and 50\% on CIFAR. 

In both CUB and CIFAR, we find that the embedding-based distillation methods such as RKD~\cite{Park2019Relational} and AML~\cite{Budnik2021Asymmetric} improve over the vanilla training denoted as ``student''. 
Our {\name} greatly improves student's performance, outperforming RKD and AML {\em no matter how the student's label space and architecture change w.r.t. the teacher's}. 
Specifically, our re-weighted version {\name} helps more w.r.t. {\name}$^-$ without adaptive weights when the overlap ratio is low (especially in cross-task KD). When student and teacher share the same classes in standard KD, {\name} achieves little improvements than {\name}$^-$. Detailed results of various methods in Figure~\ref{fig:cross_task_distillation} are reported in the supplementary.

\begin{table}[tbp]
	\centering
	\caption{\small Mean accuracy of student on CIFAR-100. The class overlap ratio between student and teacher is $60\%$. 
	We set teacher to WRN-(40,2) and vary the (depth, width) of the student.
	}
	\begin{tabular}{@{}c|cccc@{}}
	\toprule
	(depth, width) & (40, 2)     & (16, 2)  & (40, 1)   & (16, 1)  \\ \midrule
	Teacher's EMB          & \multicolumn{4}{c}{1NN: 64.27, LR: 67.94, FT: 71.13} \\	
	Student          & 78.90  & 76.37 & 75.14 & 68.48 \\
	{\name}$^{\rm EMB}$    & 79.33 & 76.90 & 75.67 & 69.00 \\
	{\name}$^{\rm LKD}$  & 79.47 & 76.29 & 75.30 & 68.14 \\
	{\name}$^-$      & 80.02 & 76.66 & 75.79 & 68.72 \\
	\midrule
	{\name}          & \bf 80.66 & \bf 76.66 & \bf 76.52 & \bf 69.50 \\ \bottomrule
	\end{tabular}
	\label{tab:cross_task_ablation_cifar}
\end{table}

Next, we investigate the GKD configurations when nearly half of the student's labels are overlapped with the teacher (\textit{overlap ratio equals $50\%$ for CUB and $60\%$ for CIFAR}).

\noindent{\bf Will all components in {\name} help?}
Given the well-trained teacher, we investigate three variants in Table~\ref{tab:cross_task_ablation_cub} and Table~\ref{tab:cross_task_ablation_cifar} besides training the student model directly (denoted as ``student''). 
We find {\name} and its variants improve a lot over the ``student'', which verifies the effectiveness of each component. In particular, by comparing {\name}$^{\rm EMB}$ and ``Student'', we find {\name} learns more discriminative embeddings after the first stage, fine-tuning upon which leads to better ``downstream'' classification results. 
{\name}$^{\rm LKD}$ applies our LKD together with vanilla objective, whose results indicate LKD itself acts as a useful distillation strategy.
The improvements between {\name}$^-$ over {\name}$^{\rm EMB}$/{\name}$^{\rm LKD}$ show that our comparison matching and LKD facilitate knowledge transfer in an orthogonal manner. The best performance of {\name} validates that if we adjust the influence of the teacher for different instances during the GKD, the helpful supervision from the teacher will direct the student to generalize better.

\begin{figure}[t]
	\begin{subfigure}[b]{0.48\linewidth}
	\centering
	\includegraphics[width=\textwidth]{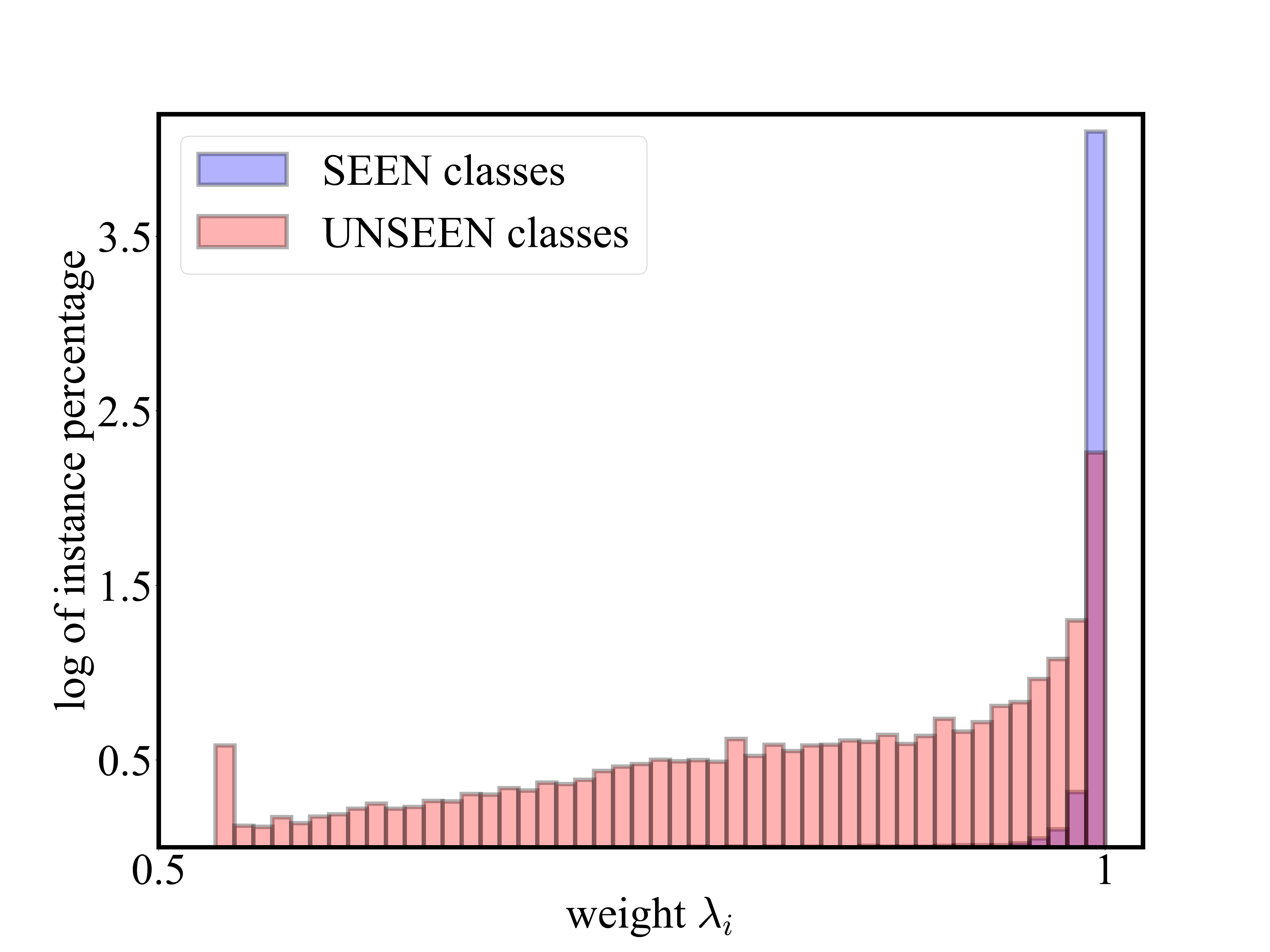}
	\caption{\small Weight distribution.}
	\label{fig:weight_dis}
	\end{subfigure}
	\begin{subfigure}[b]{0.48\linewidth}
	\centering
	\includegraphics[width=\textwidth]{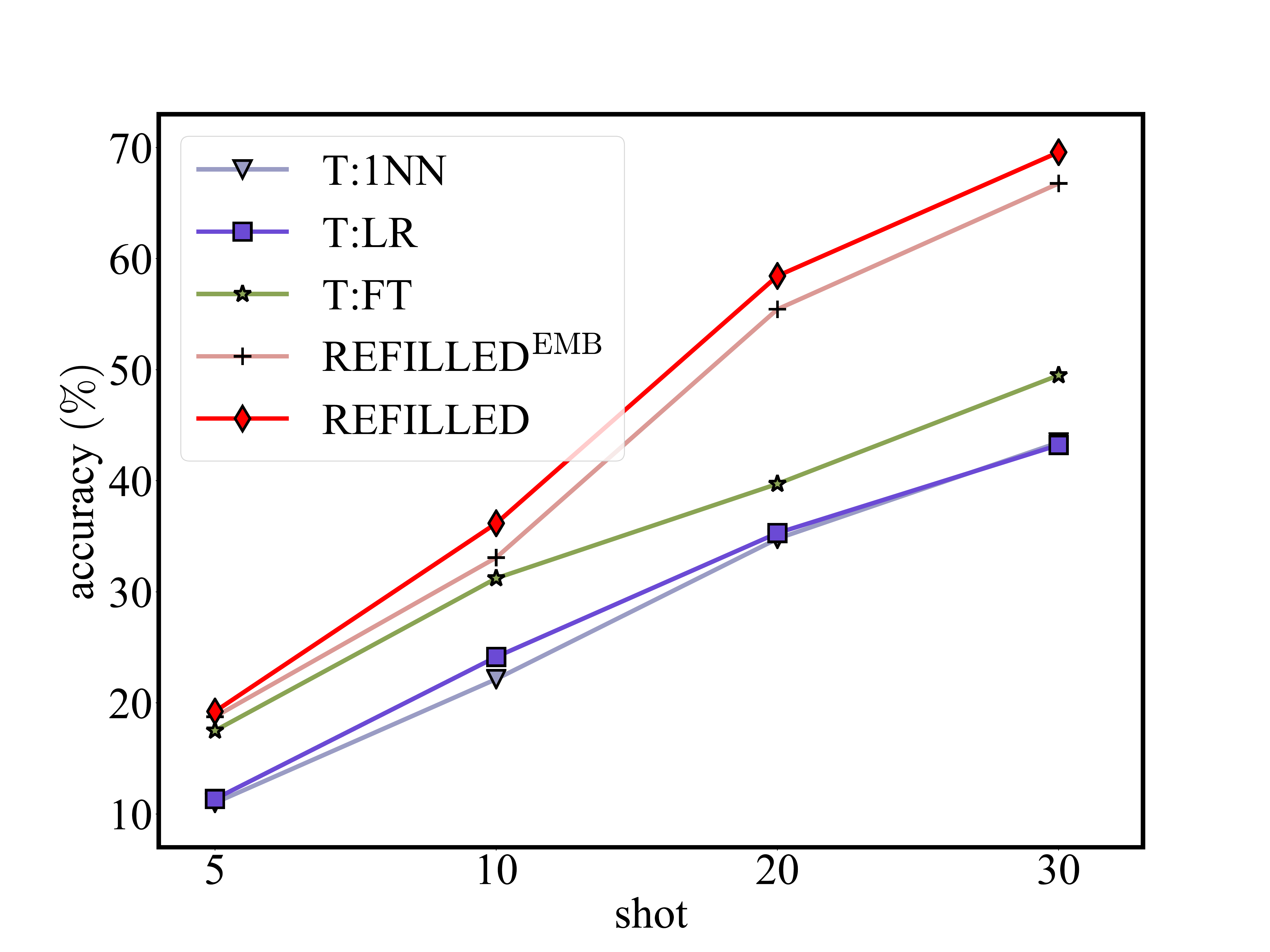}
	\caption{\small Change of accuracy.}
	\label{fig:n_training_instances}
	\end{subfigure}
	\caption{\small (a) Teacher's weight distribution of instances belonging to both seen classes and unseen classes. For simplicity, $\lambda$ is set to 1 and $\lambda_i\in(0, 1]$. The weights of instances from seen classes are observably higher than those of unseen classes, and it is easy to recognize unseen classes. (b) The change of accuracy when the number of instances per class (shot) varies. The overlap ratio of student's labels space w.r.t. teacher's is set to $0\%$. The width multiplier of student is set to 1. T:1NN means the nearest neighbour classifier based on teacher's embedding network. T:LR means the logistic regression classifier trained on teacher's embedding network. T:FT means fine-tuning teacher's embedding network together with a linear classifier.}
\end{figure}

\noindent{\bf Will adaptive weights $\mathbf{\lambda_i}$ differentiate seen and unseen classes?}
As demonstrated, the instance-specific weights $\lambda_i$ in Eq.~\ref{eq:adaptive_weight} is a key component for GKD. Since there simultaneously exist cross-task instances (denoted as unseen) and same-task instances (denoted as seen) in the student's training set, the teacher may predict with different confidences to them, \eg, higher confidence over those seen instances where the teacher is trained from. We check whether the adaptive $\lambda_i$ can differentiate seen and unseen classes. Figure~\ref{fig:weight_dis} illustrates the distribution of $\lambda_{i}$ on CIFAR-100. 
where we set $\lambda$ to 1 and $\lambda_i\in(0, 1]$. 
We find weights for instances belonging to the seen classes are observably larger than those for instances belonging to unseen classes. This means our re-weight strategy can successfully recognize seen/unseen classes and put larger weights on those familiar instances automatically. In addition, we also compute the AUC when we use $\lambda_i$ to differentiate seen and unseen classes in the GKD scenario, whose value is 0.959.

\noindent{\bf Other re-weight strategies of $\mathbf{\lambda_i}$.}
In Eq.~\ref{eq:adaptive_weight}, {\name} re-weights the distillation term with higher weights on a target task instance if the teacher's prediction is close to its Pseudo Label (denoted as ``PL-Weight''). There are some other re-weight strategies~\cite{Li2021Probabilisitic}. For example, ~\cite{Zhang2020Prime} constructs a probabilistic model and utilizes another branch to estimate the variance in distillation, which requires more parameters.
We also investigate another re-weight implementation inspired from~\cite{Zhang2020Prime}, which incorporates both student's prediction $p_S(s|\x)$ and teacher's prediction $p_T(s|\x)$ into account without additional branches. In detail, we set $\lambda_i$ based on the divergence between two predictions, \ie,
\begin{equation*}
\lambda_i = 2\lambda \times  \sigma\left(-\mathbf{KL}\left(\{p_T(s|\x)\}_{s\in\mathcal{S}},\; \{p_S(s|\x)\}_{s\in\mathcal{S}}\right)\right)\;,
\end{equation*}
A larger gap indicates that an instance is more dissimilar to the target task, and this strategy will output smaller weight $\lambda_i$. We denote the manner setting $\lambda_i$ above as ``Gap-Weight'', and compare with our previous ``PL-Weight'' in Table~\ref{tab:table1_aug} and Table~\ref{tab:table2_aug} on CUB and CIFAR, respectively. We find both strategies improve w.r.t. the vanilla version {\name}$^-$ without additional $\lambda_i$, and PL-weight gets slightly better results in most cases.

\begin{table}[t]
  \centering
  \caption{\small Mean accuracy of student models on CUB. The overlap ratio of student’s label space w.r.t. teacher’s is fixed to 50\%.}
    \begin{tabular}{c|cccc}
    \addlinespace
    \toprule
    width multiplier & 1     & 0.75  & 0.5   & 0.25 \\
    \midrule
    {\name}$^-$ & 69.14 & 67.43 & 67.18 & 64.62 \\
    Gap Weight & 70.13 & 68.05 & 68.13 & 65.47 \\
    PL Weight & \bf 70.25 & \bf 68.39 & \bf 68.50  & \bf 65.53 \\
    \bottomrule
    \end{tabular}
  \label{tab:table1_aug}
\end{table}

\begin{table}[t]
  \centering
  \caption{\small Mean accuracy of student models on CIFAR-100. The overlap ratio of student’s label space w.r.t. teacher’s is 60\%.}
    \begin{tabular}{c|cccc}
    \addlinespace
    \toprule
    (depth, width) & (40, 2) & (16, 2) & (40, 1) & (16, 1) \\
    \midrule
    {\name}$^-$ & 80.02 & 76.66 & 75.79 & 68.72 \\
    Gap Weight & 80.53 & 76.71 & 76.38 & 69.61 \\
    PL Weight & 80.66 & 76.66 & 76.52 & 69.50 \\
    \bottomrule
    \end{tabular}
  \label{tab:table2_aug}
\end{table}

\begin{table}[t]
	\centering
	\caption{\small Mean accuracy on CUB using one-stage learning with additional balance hyper-parameter $\gamma$ (denoted as {\name}$^\gamma$) and the proposed two-stage learning. The overlap ratio of student's label space w.r.t. teacher's is set to $50\%$ (upper) and $0\%$ (lower) for GKD and cross-task KD, respectively.}
	\begin{tabular}{@{}c|cccc@{}}
	\toprule
	Overlap Ratio = 50\% & 1     & 0.75  & 0.5   & 0.25  \\ \midrule
	{\name}          & \bf 75.13 & \bf 71.67 & \bf 71.06  & \bf 68.22 \\
	{\name}$^{\gamma}$  & 74.25 & 70.38 & 69.94 & 67.52 \\
	\midrule\midrule
	Overlap Ratio = 0\% & 1     & 0.75  & 0.5   & 0.25  \\ \midrule
	{\name}          & \bf 75.13 & \bf 71.67 & \bf 71.06  & \bf 68.22 \\
	{\name}$^{\gamma}$  & 74.25 & 70.38 & 69.94 & 67.52 \\ \bottomrule
	\end{tabular}
	\label{tab:cross-task_one_stage}
\end{table}

\noindent{\bf One-stage vs. two-stage learning.}
We train {\name} in a two-stage manner in Alg.~\ref{alg:flow} as~\cite{RomeroBKCGB14,Ahn2019Variational}, where objectives in two stages could be trained in a joint way.
{\name}$^\gamma$ means training {\name} in a one-stage manner with additional balancing hyper-parameter $\gamma$.
From the model design perspective, the two-stage training in {\name} works well since the distilled discriminative embedding acts as a better initialization hence {\em improves the discerning ability} of the model. While training with a combined objective {\em regularizes} the classifier by matching the predictions between student and teacher, which relies on a suitable regularization strength. 
From the implementation perspective, an important issue for the joint training of the combined objective is to {\em set the right balance} among the embedding learning (relationship distillation), classification (cross-entropy), and knowledge transition (LKD) losses. In our empirical study, we tune $\gamma$ on validation set but it is a bit hard to find the optimal balance.
In the two-stage training strategy, we can first learn a good embedding till convergence, and then use such embedding to initialize the second stage, where the balance between classification and distillation is solved with an annealing strategy. 
From the results in Table~\ref{tab:cross-task_one_stage}, the two-stage training makes {\name} easier to achieve higher performance.

\subsubsection{Results and Analyses on Cross-Task KD}
We further consider a more difficult scenario where the student target fully non-overlapped classes w.r.t. the teacher (\textit{overlap ratio equals $0\%$}).

\begin{table}[t]
  \centering
  \caption{Mean accuracy of student on CUB. The class overlap ratio between student and teacher is 0\%. We set teacher as MobileNet-0.75 and vary the width multiplier of the student.}
    \begin{tabular}{c|cccc}
    \addlinespace
    \toprule
    Width Multiplier & 1     & 0.75  & 0.5   & 0.25 \\
    \midrule
    Student & 71.25 & 67.56 & 66.85 & 64.48 \\
    RKD~\cite{Park2019Relational}   & 70.94 & 67.95 & 67.32 & 65.03 \\
    AML~\cite{Budnik2021Asymmetric}   & 71.32 & 68.29 & 67.85 & 65.55 \\
    {\name}$^{\rm EMB}$ & 69.41 & 68.82 & 67.11 & 64.89 \\
    {\name}$^-$ & 72.58 & 68.35 & 68.72 & 65.46 \\
    \midrule
    {\name} & \bf 73.29 & \bf 69.32 & \bf 69.04 & \bf 66.04 \\
    \bottomrule
    \end{tabular}
  \label{tab:vary_teacher2}
\end{table}

\begin{table}[t]
  \centering
  \caption{Mean accuracy of student on CIFAR-100. The overlap ratio of student’s label space w.r.t. teacher’s is fixed to 0\%. We set teacher as a WRN-(40, 2), while varying the depth of the student's ResNet model. {\name} achieves the best performance among others.}
    \begin{tabular}{c|ccc}
    \addlinespace
    \toprule
    depth & 50    & 32    & 14 \\
    \midrule
    Student & 75.46 & 76.72 & 73.54 \\
    RKD~\cite{Park2019Relational}   & 76.95 & 76.91 & 73.88 \\
    AML~\cite{Budnik2021Asymmetric}   & 76.42 & 76.21 & 73.89 \\
    {\name}$^{\rm EMB}$ & 76.82 & 77.15 & 73.90 \\
    {\name}$^-$ & 77.05 & 77.41 & 74.11 \\
    \midrule
    {\name} & \bf 77.43 & \bf 77.92 & \bf 74.35 \\
    \bottomrule
    \end{tabular}
  \label{tab:vary_teacher1}
\end{table}

\noindent{\bf Results of {\name} on more teacher's architectures.}
In previous experiments, we fix the teacher's architecture and vary the complexity of student's model. 
We also investigate how {\name} influences the student's discriminative ability when we change the teacher's architecture. 
We consider GKD on two different sets of 100 classes on CUB, and we set both teacher and student as MobileNet but with different depths. We use a weaker teacher, which has multiplier width $0.75$, and we investigate whether such as weaker teacher can improve the student. The results are in Table~\ref{tab:vary_teacher2}. We find {\name} improves the classification ability of a student via a cross-task teacher even the teacher's complexity is weaker than the student. The phenomenon is consistent with~\cite{Yuan2020Revisiting}, and we attribute the improvement to the usage of relationship comparisons when training the target student.

Moreover, we set teacher and student to different neural network families on CIFAR-100, \ie, teacher is WRN-(40, 2) and the student is ResNet~\cite{HeZRS16}. We vary the depth (layer number) of the student's ResNet model in $\{50, 32, 14\}$. The results are listed in Table~\ref{tab:vary_teacher1}. 
% The overlapping ratio between the classes of teacher and student is 0\%.
Similarly, {\name} facilitates the cross-task distillation across different model families, which validates the general knowledge transfer ability of {\name}.

\begin{figure}[t]
	\centering
	\includegraphics[width=\textwidth]{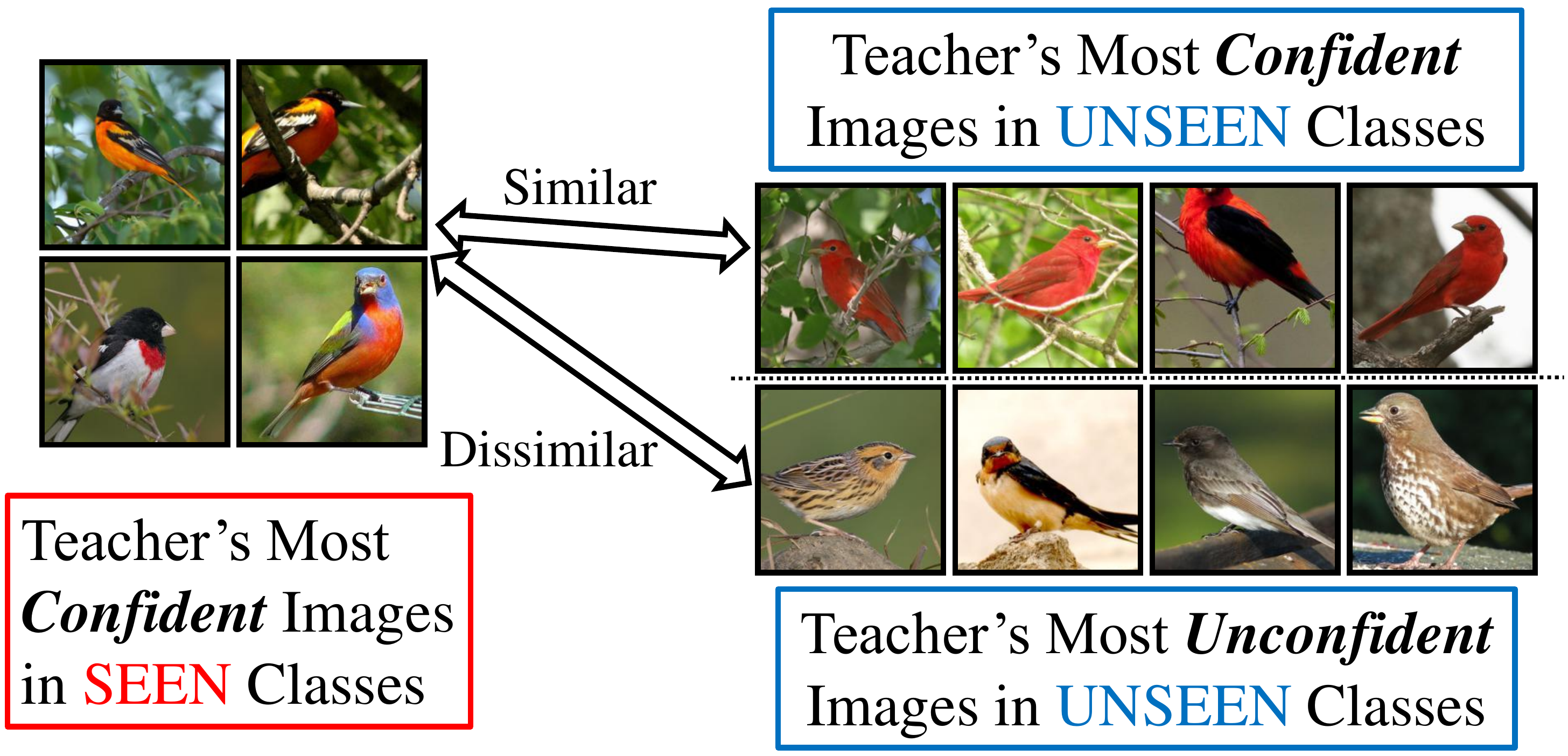}
	\caption{\small Visualization of teacher's most confident images (largest $\lambda_i$ values) in SEEN classes along with teacher's most confident/unconfident images in UNSEEN classes. Although classes in the student's training set are unseen to the teacher, our instance-specific weighting strategy assigns higher weights to those images  similar to teacher's confident images.}
	\label{fig:image}
\end{figure}

\noindent{\bf Will adaptive weights $\mathbf{\lambda_i}$ identify helpful instances?}
When the overlap ratio of student's label space w.r.t. to teacher's is $0\%$, all classes in student's training set are unseen to the teacher. Under this situation, which instances will be assigned highest confidence weights? The larger the $\lambda_i$, the more confident teacher is on $i$-th instance. In Figure~\ref{fig:image}, we can see that new images that are visually similar to the teacher's confident images in seen classes have higher weights $\lambda_i$. This means our proposed adaptive weights successfully find useful instances to transfer knowledge even though all classes are new.

\noindent{\bf {\name} with different sizes of target task data.} 
To test the extreme of the distillation ability of {\name}, we construct the target classification task with different sizes of training data on CUB. When the number of available training data is small, it is more difficult to train the student model, so that the help from the teacher becomes more important. We vary the number of instances per class (shot) in the student's task from $5$ to $30$, and the averaged classification accuracies are shown in Figure~\ref{fig:n_training_instances}. When the shot number increases, the student performs better since there are more training data.
{\name} keeps a performance margin with comparison methods in all cases. We find the gap between {\name} and the vanilla methods becomes larger when there are more shots, which indicates the distillation variants become powerful given more training data. More results on learning with limited target class data could be found in Section~\ref{sec:few-shot}.

\begin{table}[t]
  \centering
  \caption{\small Mean accuracy of student models on Dogs dataset. The teacher is trained on 200 classes from CUB. 
  We set teacher as MobileNet-1.0 and vary the width multiplier of the student.}
    \begin{tabular}{c|cccc}
    \addlinespace
    \toprule
    Width Multiplier & 1     & 0.75  & 0.5   & 0.25 \\
    \midrule
    Student & 72.35 & 70.69 & 70.11 & 68.57 \\
    SEED~\cite{Fang2021Seed}  & 72.78 & 70.90  & 70.55 & 68.32 \\
    {\name} & 72.27 & 70.81 & 70.35 & 68.64 \\
    \bottomrule
    \end{tabular}
  \label{tab:cubdog}
\end{table}

\noindent{\bf Knowledge transfer across distant tasks.}
At the end of this subsection, we analyze the phenomenon of {\name} when it transfers the knowledge across distant tasks which are not too related. 
Particularly, we train the teacher on all 200 classes on CUB for fine-grained birds, and set the target task as a 120-way fine-grained dog classification over the Stanford Dogs Dataset~\cite{dogdata}. We follow the standard splits of both datasets. 
We compare our method with a self-supervised learning method SEED~\cite{Fang2021Seed}, which distills the learned representation from previous generations. 
Mean accuracy results are listed in Table~\ref{tab:cubdog}.
In this case, {\name} is hard to improve based on the vanilla student, since the knowledge of the teacher to compare objects (about fine-grained birds) may not fit the target task (about fine-grained dogs). Since the comparison ability learned from data augmentations in SEED does not depend on classes, it transfers slightly better to a distant task than the comparison ability learned from class semantics in {\name}. 
But in this case, {\name} can still get nearly the same performance as the vanilla student. So although the teacher is not an expert on the target task, its experience will not negatively affect the training of the student.

\subsection{Standard Knowledge Distillation}\label{sec:skd}
The techniques for GKD in {\name} also facilitate standard KD where a student has the same classes with the teacher.

\noindent{\bf Datasets.} Following~\cite{Ahn2019Variational}, we test the KD ability of {\name} on CIFAR-100 and CUB. All classes in these datasets are used during training based on the standard training-test split. 

\noindent{\bf Implementation details.} 
Three different families of the neural networks are used to test the ability of {\name}, namely the ResNets~\cite{HeZRS16} , Wide ResNets~(WRN)~\cite{Zagoruyko2016Wide}, and MobileNets~\cite{Howard2017Mobile}. Towards getting different capacities of the model, we change the depth of the ResNet (through the number of layers), the (depth, width) pair of the Wide ResNet, and the width of the MobileNets (through the width multipliers).
We use similar ways in GKD to train both the teacher and the student model for standard KD. We set the temperature of the teacher's model to 4 and $\lambda=1$. For CIFAR-100, we pad 4 for each edge before the random crop. 

\noindent{\bf Evaluations.}
Both teacher and student are trained on the same set with three different seeds of initialization, and we report the mean accuracy of the student on the test set. 
Two protocols are used in standard KD, where the teacher and the student come from the same or different model families.
\begin{itemize}[leftmargin=*,noitemsep,topsep=0pt]
	\item \textit{Same-family knowledge distillation}. Both the teacher and the student come from the same model family. We use the same configuration as~\cite{Ahn2019Variational}. In CIFAR-100, both the teacher and the student are Wide ResNets. We set the (depth, width) pair of the teacher as $(40, 2)$, and change such configuration parameters of the student model among $(40, 2)$, $(16, 2)$, $(40, 1)$, and $(16, 1)$. On CUB, we consider the MobileNets, by setting the teacher's width multiplier to $1$, we vary the width multipliers of the student among $\{1, 0.75, 0.5, 0.25\}$.
	\item \textit{Different-family knowledge distillation}. The teacher and the student come from different architectural families, where the knowledge transfers from ResNets to MobileNets. Taking the computational burden into consideration, when in CIFAR-100, we choose the teacher as the ResNet-110, and we use ResNet-34 as the teacher in CUB. We only change the width multipliers of the student model in $\{0.75, 0.5, 0.25\}$ on CUB to keep the student model having a smaller capacity when compared with the teacher.
\end{itemize}

\begin{table}[tbp]
	\centering
	\caption{\small The average classification results of knowledge distillation methods on CIFAR-100 based on the Wide ResNet. We fix the teacher with (depth, width) $=(40,2)$, and set the student's capacity with different (depth, width) values. $^\dagger$ The reported results of LKD in~\cite{Li2020Local} is based on a stronger teacher with accuracy 75.61\%. $^\ddagger$ We apply SSKD for embedding distillation followed by our  classifier distillation step.}
	\begin{tabular}{c|cccc}
		\addlinespace
		\toprule
		(depth, width) & (40, 2) & (16, 2) & (40, 1) & (16, 1) \\
		\midrule
		Teacher & 74.44 &       &       &  \\
		Teacher+BAN~\cite{FurlanelloLTIA18} & 75.41 &       &       &  \\
		Student & 74.44 & 70.15 & 68.97 & 65.44 \\
		\midrule
		KD~\cite{hinton2015kd} & 75.47 & 71.87 & 70.46 & 66.54 \\
		FitNet~\cite{RomeroBKCGB14} & 74.29 & 70.89 & 68.66 & 65.38 \\
		AT~\cite{ZagoruykoK16a} & 74.76 & 71.06 & 69.85 & 65.31 \\
		NST~\cite{Huang2017Like} & 74.81 & 71.19 & 68.00    & 64.95 \\
		VID-I~\cite{Ahn2019Variational} & 75.25 & 73.31 & 71.51 & 66.32 \\
		KD+VID-I~\cite{Ahn2019Variational} & 76.11 & 73.69 & 72.16 & 67.19 \\
		SEED~\cite{Fang2021Seed} & 76.28 & 73.40 & 71.83 & 67.75 \\
		RKD~\cite{Park2019Relational} & 76.62 & 72.56 & 72.18 & 65.22 \\
		LKD$^\dagger$~\cite{Li2020Local} & -     & 75.44 & -     & 67.72 \\
	    SSKD$^\ddagger$~\cite{Xu20Knowledge} & 75.42  & 74.03 & \bf 72.71  & 67.30 \\
		\midrule
		{\name} & \bf 77.90 & \bf 75.71 & 72.54 & \bf 68.13 \\
		\bottomrule
	\end{tabular}
	\label{tab:same_model_cifar}
\end{table}

\begin{table}[tbp]
	\centering
	\caption{\small The average classification results of knowledge distillation methods on CUB based on MobileNets. We fix the teacher's width multiplier to $1.0$, and change the student's multipliers.}
	\begin{tabular}{c|cccc}
		\addlinespace
		\toprule
		Width Multiplier & 1     & 0.75  & 0.5   & 0.25 \\
		\midrule
		{Teacher} & 75.36 &       &       &  \\
		Teacher+BAN~\cite{FurlanelloLTIA18} & 76.87 &       &       &  \\
		{Student} & 75.36 & 74.87 & 72.41 & 69.72 \\
		\midrule
		{KD~\cite{hinton2015kd}} & 77.61 & 76.02 & 74.24 & 72.03 \\
		{FitNet~\cite{RomeroBKCGB14}} & 75.10  & 75.03 & 72.17 & 69.09 \\
		{AT~\cite{ZagoruykoK16a}} & 76.22 & 76.10  & 73.70  & 70.74 \\
		{NST~\cite{Huang2017Like}} & 76.91 & 77.05 & 74.03 & 71.54 \\
		{KD+VID-I~\cite{Ahn2019Variational}} & 77.03 & 76.91 & 75.62 & 72.23 \\
		{RKD~\cite{Park2019Relational}} & 77.72 & 76.80  & 74.99 & 72.55 \\
		\midrule
		{\name} & \bf 79.33 & \bf 78.52 & \bf 76.90 & \bf 74.04 \\
		\bottomrule
	\end{tabular}
	\label{tab:same_model_cub}
\end{table}

\subsubsection{Distillation From Same Architecture Family Models}
We first investigate the distillation ability when teacher and student come from the same model family. 
The results on CIFAR-100 and CUB are in Table~\ref{tab:same_model_cifar} and Table~\ref{tab:same_model_cub}, respectively. On CIFAR-100 we exactly follow the evaluation protocol in~\cite{Ahn2019Variational}, which implements teacher and student with the Wide ResNet. We re-implement RKD~\cite{Park2019Relational} and cite the results of other comparison methods from~\cite{Ahn2019Variational,Li2020Local}. For CUB, we use MobileNets as the basic model. Since the teacher possesses more capacity, its learning experience assists the training of the student once utilizing the knowledge distillation methods. {\name} achieves the best performance in almost all settings, which verifies transferring knowledge for both embedding and classifier is one of the key factors for KD.

\begin{table}[tbp]
	\centering
	\caption{\small The nearest class mean accuracy on CIFAR-100 to evaluate the embedding quality before and after the Comparison Matching~(CM) step, the first stage, in {\name}. $^\dagger$ We transform the distillation strategy of SEED~\cite{Fang2021Seed} to a supervised version using class semantics, and implement the embedding distillation over all instances in a mini-batches.}
	\begin{tabular}{c|cccc}
		\addlinespace
		\toprule
		(depth, width) & (40, 2) & (16, 2) & (40, 1) & (16, 1) \\
		\midrule
		{w/o CM} & 55.47 & 50.14 & 45.04 & 38.06 \\
		SEED$^\dagger$~\cite{Fang2021Seed} & 61.32 & 52.57 & 51.24 & 43.50 \\
		SSKD~\cite{Xu20Knowledge} & 61.25  & \bf 54.77 & 52.22  & 43.90 \\
		{w/ CM} & \bf 62.12 & 53.86 &\bf  52.71 & \bf 44.33 \\
		\bottomrule
	\end{tabular}\\
	\label{tab:NMI}
\end{table}

\begin{figure}[t]
	\centering
	\begin{minipage}[h]{0.48\textwidth}
		\centering
		\includegraphics[width=\textwidth]{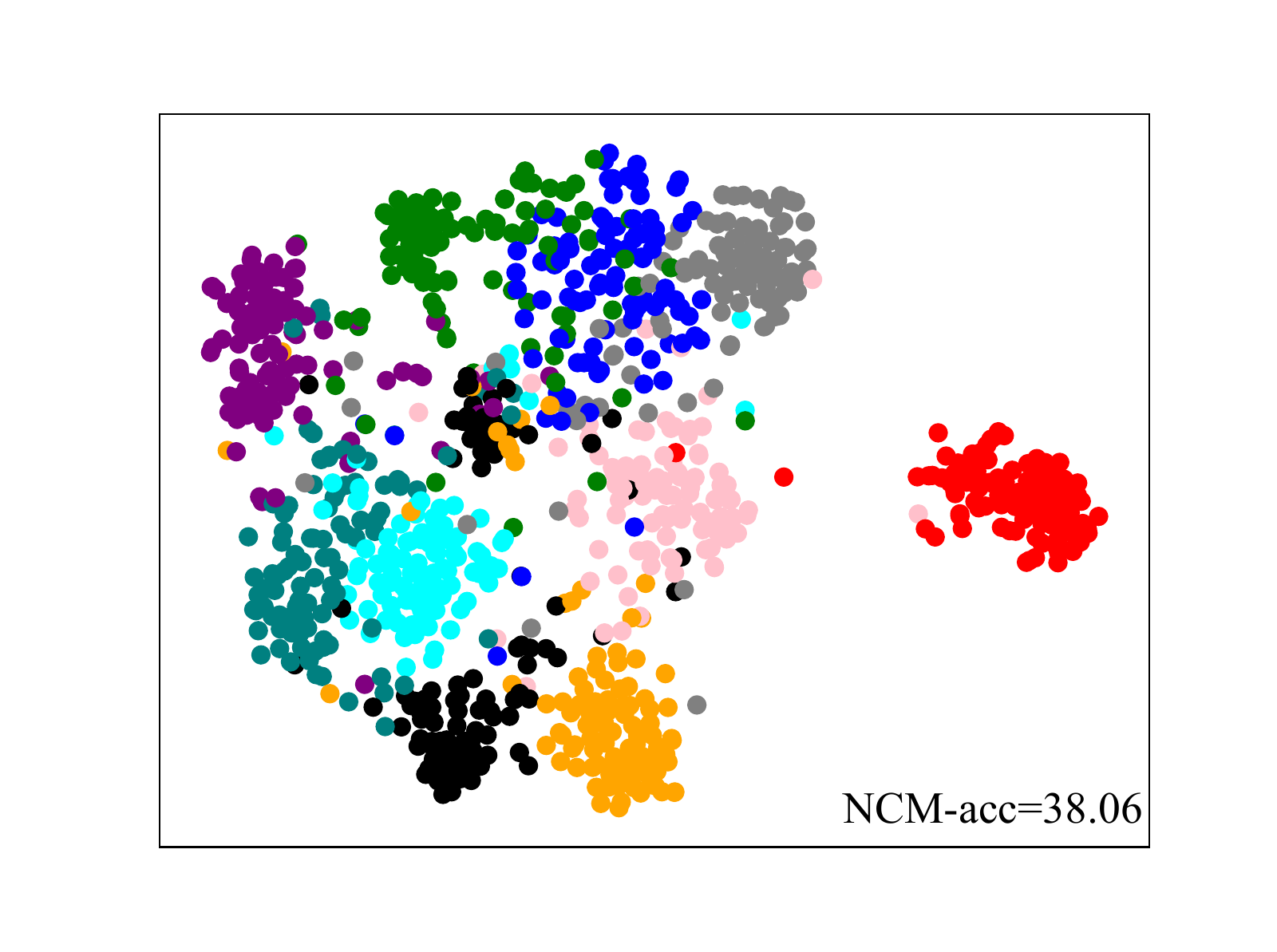}\\
	\end{minipage}
	\begin{minipage}[h]{0.48\textwidth}
		\centering
		\includegraphics[width=\textwidth]{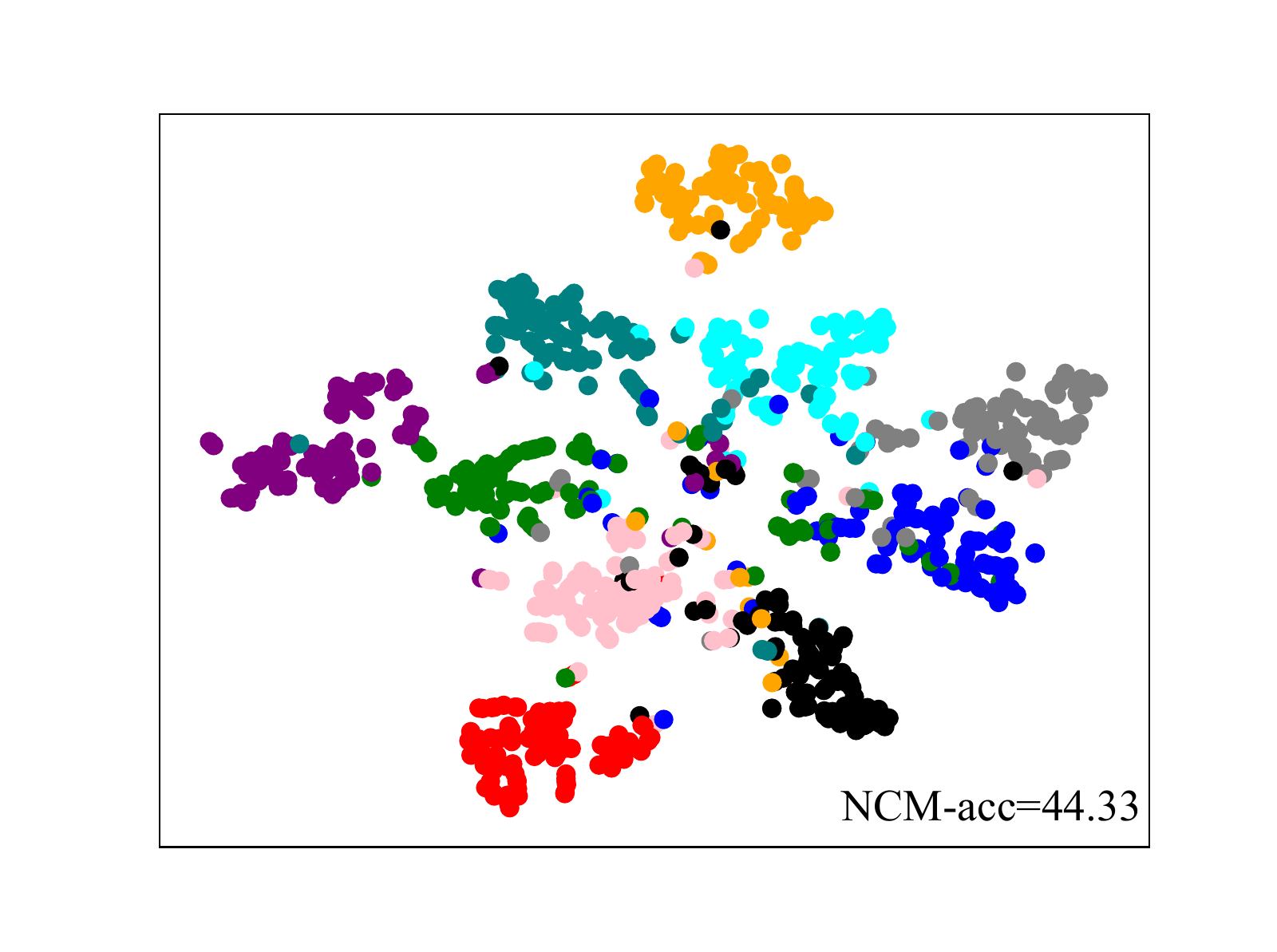}\\
	\end{minipage}
	\caption{\small The tSNE~\cite{Maaten2012stochastic} of the vanilla student training (left) and the improved embedding after the 1st stage of {\name} (right) over 10 classes sampled from CIFAR-100.}
	\label{fig:same_task_ablation}
\end{figure}

\noindent{\bf Will embedding distillation help?}
There are two stages in {\name}. In the first stage we distill the discriminative embedding from the teacher and improve the comparison ability of the student. 
We evaluate the quality of the embedding via its classification accuracy based on Nearest Class Mean classifier~(NCM). In detail, we extract features on instances from the student's training set, and then compute the center for each class. Finally, we predict a test instance based on the label of its nearest class in the embedding space.
In Table~\ref{tab:NMI}, we compute NCM accuracy for student model's embedding trained with and without aligning the teacher's tuples (denoted as comparison matching, CM) in CIFAR-100. 
We compare our comparison matching with the relationship distillation strategy SSKD in~\cite{Xu20Knowledge}, where self-distillation strategy is incorporated. We also transform the relationship distillation in SEED~\cite{Fang2021Seed} to a supervised version, which determines the similarity between instances through their class labels. 
Table~\ref{tab:same_model_cifar} contains the classification results when we distill the teacher with our LKD based on the embedding learned by SSKD, SEED, and {\name}.
Figure~\ref{fig:same_task_ablation} visualizes the embedding quality over 10 sampled classes using tSNE~\cite{Maaten2012stochastic}. 
Both quantitative and qualitative results validate that the quality of the student's embedding is improved after distilling the knowledge from the teacher. Thus the comparison matching step in {\name} is effective for knowledge distillation. The results also verify using the NCM predictions to direct the classifier training in the 2nd stage of {\name} is reasonable.

\begin{table}[t]
	\caption{\small The mean accuracy on CIFAR-100 to evaluate the effectiveness of Local Knowledge Distillation~(LKD) in {\name}.}
	\centering
	\begin{tabular}{c|cccc}
		\addlinespace
		\toprule
		(depth, width) & (40, 2) & (16, 2) & (40, 1) & (16, 1) \\
		\midrule
		{w/ KD} & 77.08 & 73.57 & 72.24 & 67.14 \\
		{w/ LKD} & \bf 77.90 & \bf 74.82 & \bf 72.54 & \bf 68.13 \\
		\bottomrule
	\end{tabular}
	\label{tab:local_KD}
\end{table}

\begin{table}[t]
	\centering
	\caption{\small The average classification accuracy of standard KD on CIFAR-100. The teacher is trained with ResNet-110, which gets 74.09\% test accuracy. The student is learned with MobileNets, whose width multiplier is changed.}
	\begin{tabular}{c|cccc}
		\addlinespace
		\toprule
		Width Multiplier & 1     & 0.75  & 0.5   & 0.25 \\
		\midrule
		Student & 68.57 & 67.92 & 65.66 & 60.87 \\
		\midrule
		KD~\cite{hinton2015kd}    & 70.34 & 68.21 & 66.06 & 61.38 \\
		FitNet~\cite{RomeroBKCGB14} & 67.99 & 67.85 & 65.12 & 61.01 \\
		AT~\cite{ZagoruykoK16a}    & 68.97 & 67.88 & 66.44 & 62.15 \\
		NST~\cite{Huang2017Like}   & 70.62 & 70.49 & 69.15 & 61.32 \\
		KD+VID-I~\cite{Ahn2019Variational} & 71.94 & 70.13 & 68.51 & 62.50 \\
		RKD~\cite{Park2019Relational}   & 70.41 & 68.93 & 66.24 & 61.44 \\
		\midrule
		{\name} & \bf 73.75 & \bf 72.65 & \bf 70.32 & \bf 62.90 \\
		\bottomrule
	\end{tabular}
	\label{tab:cifar_different_models}
\end{table}

\begin{table}[t]
	\centering
	\caption{\small The mean classification accuracy of standard KD on CUB. Teacher is trained with ResNet-34, which gets 75.31\% test accuracy. Student is learned with MobileNets, whose width multiplier is changed.}
	\begin{tabular}{c|ccc}
		\addlinespace
		\toprule
		width multiplier & 0.75  & 0.5   & 0.25 \\
		\midrule
		Student & 74.87 & 72.41 & 69.72 \\
		\midrule
		KD~\cite{hinton2015kd}    & 76.02 & 74.17 & 71.97 \\
		FitNet~\cite{RomeroBKCGB14} & 75.03 & 72.17 & 70.03 \\
		AT~\cite{ZagoruykoK16a}    & 76.11 & 72.94 & 70.99 \\
		NST~\cite{Huang2017Like}   & 75.89 & 73.82 & 71.92 \\
		KD+VID-I~\cite{Ahn2019Variational} & 76.41 & 74.04 & 72.20 \\
		RKD~\cite{Park2019Relational}   & 76.11 & 75.24 & 72.84 \\
		\midrule
		{\name} & \bf 78.40 & \bf 76.52 & \bf 73.44 \\
		\bottomrule
	\end{tabular}
	\label{tab:cub_different_models}
\end{table}    
\noindent{\bf Will local knowledge distillation help?}
Results in Table~\ref{tab:local_KD} verify the further improvement of Local Knowledge Distillation~(LKD) in Eq.~\ref{eq:obj2} compared with the vanilla Knowledge Distillation~(KD) when training based on the distilled embedding after the first stage of {\name}. A local consideration of probability matching helps.

\begin{table*}[htbp]
	\centering
	\caption{\small One-step incremental learning on CUB, where we follow the same setting as the GKD (overlap ratio equals $0\%$). There are four criteria: 1. the mean accuracy over instances from teacher's old classes ${\mathcal{C}^\prime}\rightarrow{\mathcal{C}}\cup{\mathcal{C}^\prime}$, 2. the mean accuracy over instances from target classes ${\mathcal{C}}\rightarrow{\mathcal{C}}\cup{\mathcal{C}^\prime}$, 3. the mean accuracy over all classes, 4. the harmonic mean value of the mean accuracy over instances from teacher's and student's classes. The two baselines of the teacher denote we concatenate the teacher's classifier with the LR and fine-tuned (FT) model trained over the student's instances, given teacher and student {\em have the same architecture}. EWC only works for a student network that shares the same architecture as the teacher (width=$1$).}
	\tabcolsep 2pt
	\begin{tabular}{@{}c|cccc|cccc|cccc|cccc@{}}
	\toprule
	Criterion      & \multicolumn{4}{c|}{Accuracy ${\mathcal{C}^\prime}\rightarrow{\mathcal{C}}\cup{\mathcal{C}^\prime}$} & \multicolumn{4}{c|}{Accuracy ${\mathcal{C}}\rightarrow{\mathcal{C}}\cup{\mathcal{C}^\prime}$}  & 			\multicolumn{4}{c|}{Accuracy}             & \multicolumn{4}{c}{Harmonic Mean}        \\ \midrule
	Width          & 1          & 0.75       & 0.5       & 0.25      & 1         & 0.75     & 0.5      & 0.25    & 1        	& 0.75     & 0.5      & 0.25     & 1        & 0.75     & 0.5      & 0.25    \\ \midrule
	Teacher        & \multicolumn{4}{c|}{LR: 63.73, FT: 49.85}       & \multicolumn{4}{c|}{LR: 31.97, FT: 45.70} & 			\multicolumn{4}{c|}{LR: 47.85, FT: 47.77} & \multicolumn{4}{c}{LR: 42.58, FT: 47.68} \\
	EWC~\cite{kirkpatrick2017overcoming} 
	& \bf 35.66      &     &     &     & 55.65    &      &      &    & 	45.66    &      &     &    & 43.47    &    &   &   \\
	Student + LwoF~\cite{Li2018Learning} & 33.79      & 33.55      & 29.39     & 21.47     & 56.84     & 56.62    & 55.76    & 53.46   & 			45.32    & 45.09    & 42.58    & 37.47    & 42.39    & 42.14    & 38.50    & 30.64   \\
	RKD~\cite{Park2019Relational} + LwoF~\cite{Li2018Learning}     & 34.30      & 34.02      & 30.21     & 21.98     & 57.27     & 57.63    & 55.90    & 53.81   & 			45.79    & 45.83    & 43.06    & 37.90    & 42.91    & 42.79    & 39.23    & 31.22   \\ \midrule
	{\name}        & 35.50      & \bf 35.13      & \bf 31.05     & \bf 24.51     & \bf 58.57     & \bf 58.70    & \bf 58.00    & \bf 53.98   & 			\bf 47.04    & \bf 46.92    & \bf 44.53    & \bf 39.25    & \bf 44.21    & \bf 43.96    & \bf 40.45    & \bf 33.72   \\ \bottomrule
	\end{tabular}
	\label{tab:incremental}
\end{table*}
\subsubsection{Distillation From Different Model Families}
To further evaluate the performance of {\name}, we use {\name} in standard KD but with a teacher with cross-family architecture. For CIFAR-100, we set the teacher as ResNet-110, and use the MobileNets with different channels as the student. For CUB, we set the teacher as ResNet-34, and use the MobileNets with different width multipliers (from $\{0.75, 0.5, 0.25\}$) as the student. Results on CIFAR-100 and CUB are reported in Table~\ref{tab:cifar_different_models} and Table~\ref{tab:cub_different_models}, respectively. 
{\name} keeps its superiority in all cases, which indicates its practical utility with different teacher's configurations.

\subsection{KD for One-Step Incremental Learning}
We claim that the distillation ability in {\name} acts as a better way to prevent catastrophic forgetting~\cite{Li2018Learning}, and facilitates to augment the discerning ability in the one-step incremental learning environment. In other words, the student not only distills the dark knowledge in $f_T$ to improve its classification ability in the target $C$ classes but also augments the classifier to discern those $C'=|\mathcal{C}'|$ classes from the teacher's task. 
Finally, $f$ becomes a {\em joint classifier} over $\mathcal{C}\cup\mathcal{C}'$ with $C+C'$ classes. The student's classifier $W$ is augmented with $\hat{W}\in\mathbb{R}^{d\times C'}$ for classes in $\mathcal{C}'$ simultaneously.
Note that different from the vanilla incremental learning, {\name} calibrates both old and target classes well and can be applied {\em across different architectures}.

\noindent{\bf Datasets.} Following~\cite{Ahn2019Variational}, we test {\name} on CUB, based on the scenario of the cross-task KD in section~\ref{sec:GKD} (there is no class shared between teacher and student). 

\noindent{\bf Evaluations.} We first consider the classification accuracy over all 200 classes, which encodes the 200-way classification for instances from both old and current 100 classes. The same number of instances from old and new classes are used to evaluate the model. To avoid a biased accuracy towards a certain split, we follow~\cite{Xian2017Zero,Ye2019Learning} and compute the {\em harmonic mean} value of the two types of mean accuracy, \ie, the mean accuracy for instances from classes in teacher's and student's tasks (denoted as ``${\mathcal{C}^\prime}\rightarrow{\mathcal{C}}\cup{\mathcal{C}^\prime}$'' and ``${\mathcal{C}}\rightarrow{\mathcal{C}}\cup{\mathcal{C}^\prime}$'', respectively). 
A model has a high harmonic mean if it performs well on both first and second splits of all classes. Detailed formulations are in the supplementary.

\noindent{\bf Implementation details.} 
We follow~\cite{Zhao2020Maintaining,Kang2020Decoupling} to discard the learned bias and $\ell_2$-normalize the learned weights of all 200 classes, which leads to a better calibrated joint classifier for compared and our methods.

\noindent{\bf Comparison methods.}
\begin{itemize}[leftmargin=*,noitemsep,topsep=0pt]
	\item \textit{Combined classifier with teacher}. 
	We concatenate the target class classifier with the teacher's classifier and get a joint one for all classes. We normalize classifiers for LR and FT.
	\item \textit{Variants of standard KD}. We match the student's predictions on the old classes with the teacher's classifier~\cite{Li2018Learning} (denoted as LwoF) to avoid catastrophic forgetting. We finally concatenate the learned student's classifier with the teacher's one and normalize them for better calibration.
	\item \textit{Incremental learning methods}. Since in our one-step incremental learning task, only a fixed model is provided, so we compare with incremental learning method EWC~\cite{kirkpatrick2017overcoming} using the fixed model only {\em without using a small number of examples from the previous task for experience replay}~\cite{Liu2020More,Tao2020Topology}. 
\end{itemize}

\noindent{\bf Results.}
The results are in Table~\ref{tab:incremental}, and the student is required to make a holistic classification on all classes, \ie, the union of teacher's and student's class sets. Four criteria are utilized to evaluate a GKD model. Since the accuracy is computed over all classes, a model may predict target class instances well and forget the knowledge the teacher introduced at the initial KD stage. The harmonic mean reveals the joint ability of the classifier over both previous and target classes, and a model achieves a high value only if it predicts both sets of classes well. 
GKD improves the classification ability of the model than using the teacher's embeddings directly.
% (we leave the 1NN results of the teacher in the supplementary).
Similar to LwoF~\cite{Li2018Learning}, we apply a KD term on all old classes to prevent forgetting during the training progress of the vanilla student model and RKD. With LwoF, the student and RKD can handle all classes. We find comparison methods could not balance the predictions of old and target classes.
Our {\name} calibrates the two sets of classes better and gets the best performance w.r.t. both the accuracy and harmonic mean criteria. The margin with other methods becomes larger when we distill the knowledge to a much smaller student. The results validate that the proposed {\name} expands the ability of the student model for more classes effectively. 

\begin{table}[tbp]
	\centering
	\caption{\small The mean accuracy of GKD on CUB dataset. Two criteria, \ie, accuracy and Harmonic Mean (HM) accuracy are shown. We fix the width multiplier of the student to 1.}
	\tabcolsep 3pt
	\begin{tabular}{c|cccc}
		\addlinespace
		\toprule
		Method & Vanilla & {\name}$^{\rm EMB}$ & {\name}$^-$   & {\name} \\
		\midrule
		Acc. & 45.32 & 45.87 & 46.39 & \bf 47.04 \\
		HM Acc. & 42.39 & 42.90 & 43.45 & \bf 44.21 \\
		\bottomrule
	\end{tabular}
	\label{tab:gkd_ablation}
\end{table}

\noindent{\bf Does {\name} calibrate better on GKD?} Similar to Table~\ref{tab:cross_task_ablation_cub}, we evaluate the GKD performance of {\name} variants in Table~\ref{tab:gkd_ablation}. We keep both teacher and student have the same architecture.
We normalize the classifier for all comparison methods. For comparison baselines, we concatenate the student's model with the teacher's classifier. For ``vanilla'', the student model is trained from scratch; for ``{\name}$^{\rm EMB}$'', the student model is trained over the embedding learned by {\name}; while for ``{\name}$^-$'', we tune the student model with the local knowledge distillation term without instance-specific weighting. We find {\name} gets the best results in both accuracy and harmonic mean accuracy.

\subsection{Few-Shot and Middle-Shot Learning}\label{sec:few-shot}
Training a deep neural network with limited data is a challenging task, where models are prone to over-fit. We apply our {\name} approach for few-shot and middle-shot learning, where the classification ability from a teacher trained on \textsc{seen} class can be used to help the student model training for \textsc{unseen} few-shot and middle-shot tasks.

\noindent{\bf Datasets.}
We use {\it Mini}ImageNet dataset~\cite{Vinyals2016Matching} with 100 classes in total and 600 images per class. All images are resized to $84\times 84$ before inputting into the models.
Following~\cite{Vinyals2016Matching,Sachin2017}, there are 64 classes (\textsc{seen} class) to train the teacher (a.k.a. the meta-train set), 16 classes for validation (a.k.a. the meta-val set), and we sample tasks from the remaining 20 classes (a.k.a. the meta-test set) to train the student. 

\noindent{\bf Implementation details.}
We set the student as a 4-layer ConvNet~\cite{Vinyals2016Matching,Snell2017Proto,Finn2017MAML}, and consider two types of the teacher model, \ie, the same 4-layer ConvNet (but trained on different classes in the meta-train set) and the ResNet~\cite{Oreshkin2018TADAM,Ye2018Learning}. The ConvNet contains 4 identical blocks, and each block is a sequential of convolution operator, batch normalization~\cite{IoffeS15}, ReLU, and Max pooling. We add another global max-pooling layer to reduce the computational burden after the 4 blocks, which gives rise to a 64-dimensional embedding before the top-layer classifier. ResNet removes the two down-sampling layers in the vanilla version~\cite{Oreshkin2018TADAM,Ye2018Learning}, and outputs 640-dimension embeddings.
We train a teacher on \textsc{seen} classes set with ResNet/ConvNet. Supervised by the cross-entropy loss, we use random crop and horizontal flip as the data augmentation, SGD w/ momentum 0.9 as the optimizer, and 128 as the batch size. The student is trained with the help of the teacher and limited \textsc{unseen} class examples.

\noindent{\bf Evaluations.} 
Define a $M$-shot $C$-way task as a $C$-class classification problem with $M$ instances in each class. Different from the few-shot learning where $C=5$ and $K\in\{1,5\}$, here we consider there are a bit more instances in each class, \ie, $K=\{10,30\}$. Although the value of $M$ increases in middle-shot learning, it is still small to train a complicated neural network from scratch.
We sample tasks from the 20-class split (meta-test set) to train the student and evaluate the classification accuracy over another 15 instances from each of the $C$ classes. Mean accuracy over 10,000 trials are reported. We omit the confidence interval since they all have similar values around 0.2\%.

\noindent{\bf Comparison methods.}
There are two branches of baselines:
\begin{itemize}[leftmargin=*,noitemsep,topsep=0pt]
	\item \textit{Meta-learning methods}. Meta-learning mimics the test case by sampling $C$-Way $M$-Shot tasks from the \textsc{seen} class set to learn task-level inductive bias like embedding~\cite{Vinyals2016Matching,Snell2017Proto}. However, the computational burden (\eg, the batch size) sampling episodes of tasks becomes large when the number of shots increases. Besides, meta-learning needs to specify the way to obtain a meta-model from the \textsc{seen} classes. We compare {\name} with the embedding-based meta-learning approaches ProtoNet~\cite{Snell2017Proto} and FEAT~\cite{Ye2018Learning}.
	\item \textit{Embedding-based baselines}. We can make predictions directly with the teacher's embedding, the penultimate layer of the teacher model, by leveraging the nearest neighbor classifier. Based on which, we also train linear classifiers like SVM on the current task's data or fine-tune the whole model. It is notable that we tune the hyper-parameters with sampled few/middle-shot tasks on the validation split. We compare {\name} with SimpleShot~\cite{Wang2019Simple} and Neg-Cosine~\cite{Liu2020Negative}.
\end{itemize}

\begin{table}[tbp]
	\centering
	\caption{\small The mean accuracy over 10,000 trials of 5-Way $M$-shot tasks from few-shot ($K\in\{1,5\}$) to middle-shot ($K\in\{10,30\}$). We set student as the ConvNet, and investigate both ResNet and ConvNet as the teacher. {\name}$^{\rm Res}$ and {\name}$^{\rm Conv}$ denote the results with a ResNet and a ConvNet teacher, respectively.}
	\begin{tabular}{c|cccc}
		\addlinespace
		\toprule
		$M$-Shot & 1 & 5 & 10  & 30 \\
		\midrule
		1NN    & 49.73 & 63.11 & 66.56 & 69.80 \\
		SVM   & 51.61 & 69.17 & 74.24 & 77.87 \\
		Fine-Tune & 45.89 & 68.61 & 74.95 & 78.62 \\
		\midrule
		SimpleShot~\cite{Wang2019Simple} & 49.95 & 69.62 & 74.48 & 78.43 \\
		MAML~\cite{Finn2017MAML} & 48.70 & 63.11 & - & - \\
		ProtoNet~\cite{Snell2017Proto} & 51.79 & 70.38 & 74.42 & 78.10 \\
		Neg-Cosine~\cite{Snell2017Proto} & 52.84 & 70.41 & - & - \\
		FEAT~\cite{Ye2018Learning} & 55.15 & 71.61 & 74.86 & 78.84 \\    
		
		\midrule
		{\name}$^{\rm Res}$ & \bf 55.13 & \bf 72.05 & \bf  76.93 & \bf 80.75\\
		{\name}$^{\rm Conv}$ & 53.81 & 71.86 & 75.56 & 78.98 \\
		\bottomrule
	\end{tabular}
	\label{tab:middle_shot}
\end{table}

\noindent{\bf Results.}
The results of 5-way $M$-shot classification are reported in Table~\ref{tab:middle_shot}.
Both ProtoNet and FEAT are meta-learned over the pre-trained embeddings (the ConvNet teacher) from the \textsc{seen} class set (meta-train set).
When $K=\{1,5\}$ as in the standard few-shot learning setting, the meta-learning approaches perform well than the embedding-based baselines, but fine-tuning becomes a very strong baseline when the number of shots becomes large, which gets better results than ProtoNet or FEAT. 
{\name} gets better results when distilling the knowledge from a stronger teacher (\ie, the ResNet), which demonstrates the influence of teacher's capacity. 
{\name} achieves competitive results when $M$ is small, and gets better results than other methods with large $M$, which validates the importance of distilling the knowledge of a cross-task teacher for training a classifier.

\section{Conclusion}
Although knowledge distillation makes it easier to transfer learning experiences between related heterogeneous models, reusing models learned from a {\em general} label space is still difficult.
We propose generalized knowledge distillation where the student is not restricted to having identical classes with the teacher.
Our {\name} improves the learning efficiency of the target model with the help	of two stages, \ie, comparison matching and adaptive local knowledge distillation. 
{\name} aligns the comparison ability w.r.t. embeddings, removing the label space constraint while simultaneously capturing high order relationships among instances. Then, emphasizing the teacher's confident supervision makes {\name} automatically match the predictions between two models locally.
Experiments validate that {\name} improves classification performance in a variety of tasks, including general and standard knowledge distillation.

\ifCLASSOPTIONcompsoc
  \section*{Acknowledgments}
\else
  \section*{Acknowledgment}
\fi

This work is partially supported by NSFC (62006112, 61921006, 62176117), NSF of Jiangsu Province (BK20200313), and Nanjing University-Huawei Joint Research Program.

\ifCLASSOPTIONcaptionsoff
  \newpage
\fi

{\small
\bibliographystyle{IEEEtran}
\bibliography{egbib}}

\newpage

\appendices
\section{Interpretation of Comparison Matching}
The main idea of comparison matching in section 4.2.1 in the main paper is to align the similarity predictions over tuples between teacher and student. Recall that given a tuple $(\x_i, \x^\mathcal{P}_i, \x^\mathcal{N}_{i1}, \ldots, \x^\mathcal{N}_{iK})$ with one target neighbor $\x^\mathcal{P}_i$ and $K$ impostors $\{\x^\mathcal{N}_{i1}, \ldots, \x^\mathcal{N}_{iK}\}$, we optimize the embedding through
\begin{equation}
	\min_{\phi} \sum_{i} \mathbf{KL} \left(p_{i}(\phi_T) \;\big\|\; p_{i}(\phi)\right)\;.\;\label{eq:obj1}
\end{equation}
Here $p_{i}(\phi_T)$ and $p_{i}(\phi)$ are the validness probability based on teacher's and student's embedding $\phi_T$ and $\phi$, respectively.

We provide the concrete derivation for Eq. 6 in the main paper to explain the effect of comparison matching and show how the teacher's estimation of the tuples influences the embedding optimization. We first illustrate the idea in a triplet form --- an anchor $\x_i$ with one positive neighbor $\x_i^\mathcal{P}$ and {\em one negative impostor} $\x_i^\mathcal{N}$, which is easy to explain. Then we extend the analysis to the general case. 

When there is only one impostor in the tuple, we can simplify the term $p_i(\phi)$ in Eq. 3 in the main paper as 
\begin{equation}
\sigma\left(\mathbf{D}_{\phi}\left(\x_i, \x^\mathcal{N}_{i}\right)-\mathbf{D}_{\phi}\left(\x_i, \x^\mathcal{P}_i\right)\right) = \sigma\left(\mathbf{Diff}_{\x_i}\right)\;.
\end{equation}
where $\sigma(x) = 1 / (1+\exp(-x))$ is the logistic function squashing the input into $[0,1]$. We define $\mathbf{Diff}_{\x_i} = \mathbf{D}_{\phi}\left(\x_i, \x^\mathcal{N}_{i}\right)-\mathbf{D}_{\phi}\left(\x_i, \x^\mathcal{P}_i\right)$, $\rho_{i}=1-p_{i}(\phi_T)$, and logistic loss $\iota(x) = \ln(1+\exp(-x))$. In the vanilla case, we can obtain the embedding via minimizing the triplets with the logistic loss, which pushes impostor $\x_i^\mathcal{N}$ away and pulls the target neighbor $\x_i^\mathcal{P}$ close.

We can reformulate Eq. 4 in the main paper as
\begin{align}
	&\mathbf{KL}\left(p_i(\phi_T) \;\|\; p_i(\phi)\right)\label{eq:expansion}\\
	=\;\;&\;p_i(\phi_T)\ln{\frac{p_i(\phi_T)}{p_i(\phi)}} + \left(1-p_i(\phi_T)\right)\ln{\frac{1-p_i(\phi_T)}{1-p_i(\phi)}}\notag\\
	=\;\;&\;\underbrace{p_i(\phi_T)\ln{p_i(\phi_T)}}_{{\rm constant}} - p_i(\phi_T)\ln{p_i(\phi)}\notag\\
	&\qquad\;+\quad \underbrace{\left(1-p_i(\phi_T)\right)\ln{\left(1-p_i(\phi_T)\right)}}_{{\rm constant}}\notag\\
		&\;\quad\qquad\qquad-\quad (1-p_i(\phi_T))\ln{(1-p_i(\phi))}\notag
	\end{align}
Then we have 
	\begin{align*}
	&\mathbf{KL}\left(p_i(\phi_T) \;\|\; p_i(\phi)\right)\\
	\cong\;\;&\;-p_i(\phi_T)\ln{p_i(\phi)}\;-\;\ln{(1-p_i(\phi))}\\
	&\;\qquad+\;\; p_i\left(\phi_T\right)\ln{\left(1-p_i\left(\phi\right)\right)}\\
	=\;\;&\;-p_i\left(\phi_T\right)\left(-\mathbf{D}_{\phi}\left(\x_i,\x_i^\mathcal{P}\right) - \ln{\Delta}\right) +\ln{\Delta}\\
	&\;\qquad+\;\mathbf{D}_{\phi}\left(\x_i,\x_i^\mathcal{N}\right)+p_i\left(\phi_T\right)\left(-\mathbf{D}_{\phi}\left(\x_i,\x_i^\mathcal{N}\right)-\ln{\Delta}\right)
		\end{align*}
    The notation $\cong$ neglects the constant term in the equation. We set $\Delta \triangleq \exp{\left(-\mathbf{D}_{\phi}\left(\x_i,\x_i^\mathcal{P}\right)\right)} + \exp{\left(-\mathbf{D}_{\phi}\left(\x_i,\x_i^\mathcal{N}\right)\right)}$. Thus, 
	\begin{align*}
	&\mathbf{KL}\left(p_i(\phi_T) \;\|\; p_i(\phi)\right)\\
	=\;\;&\;p_i(\phi_T)\mathbf{D}_{\phi}\left(\x_i,\x_i^\mathcal{P}\right) + \left(1-p_i\left(\phi_T\right)\right)\mathbf{D}_{\phi}\left(\x_i,\x_i^\mathcal{N}\right)+\ln{\Delta}\\
	=\;\;&\;p_i\left(\phi_T\right)\mathbf{D}_{\phi}\left(\x_i,\x_i^\mathcal{P}\right) + \left(1-p_i\left(\phi_T\right)\right)\mathbf{D}_{\phi}\left(\x_i,\x_i^\mathcal{N}\right)\\
	&\quad\;+\;\ln{\left(\exp{\left(-\mathbf{D}_{\phi}\left(\x_i,\x_i^\mathcal{P}\right)\right)} + \exp{\left(-\mathbf{D}_{\phi}\left(\x_i,\x_i^\mathcal{N}\right)\right)}\right)}\\
	=\;\;&\;\left(p_i(\phi_T)-1\right)\mathbf{D}_{\phi}\left(\x_i,\x_i^\mathcal{P}\right) + \left(1-p_i\left(\phi_T\right)\right)\mathbf{D}_{\phi}\left(\x_i,\x_i^\mathcal{N}\right)\\
	&\quad\;+\;\ln{\big(1+\exp{\big(-\big(\mathbf{Diff}_{\x_i}\big)\big)}
		\big)}\\
	=\;\;&\rho_{i}\left(\mathbf{Diff}_{\x_i}\right) +\iota\left(\mathbf{Diff}_{\x_i}\right)\;.
\end{align*}
At last, minimizing the KL-divergence equals minimizing two losses. The first part is a rectification term and the second part is the vanilla logistic loss. Therefore, the comparison matching weakens the effect of the student's embedding updates from the label supervision by incorporating the teacher's knowledge. More discussions are in Section 4.4.1 in the main paper.

In the following, we extend the previous analysis to the general case when $K>1$. Based on Eq.~\ref{eq:obj1}, we have
\begin{align}
	&\mathbf{KL} \left(p_{i}(\phi_T) \;\big\|\; p_{i}(\phi)\right)\notag\\
\cong\;\;&	\underbrace{\mathbf{KL} \left(p_{i}(\phi_T) - \mathbf{e}_0 \;\big\|\; p_{i}(\phi)\right)}_{{\rm rectification\;term}} + \underbrace{\mathbf{KL} \left(\mathbf{e}_0 \;\big\|\; p_{i}(\phi)\right)}_{{\rm contrastive\;loss}}\;.\label{eq:decomposition}
\end{align}
We set $\mathbf{e}_0 = [1, 0, \ldots, 0]$ as an all-zero value vector except for the first element, whose size is the same as $p_{i}(\phi)$. Particularly, we have
\begin{align}
	& \mathbf{KL} \left(\mathbf{e}_0 \;\big\|\; p_{i}(\phi)\right)\notag\\ \cong\;\;& -\ln \s_{\tau}\left(\left[-\mathbf{D}_{\phi}\left(\x_i, \x^\mathcal{P}_i\right), \ldots, -\mathbf{D}_{\phi}\left(\x_i, \x^\mathcal{N}_{iK}\right)\right]\right)\notag\\
	=\;\;& -\ln \frac{\exp\left(-\mathbf{D}_{\phi}\left(\x_i, \x^\mathcal{P}_i\right)\right)}{\exp\left(-\mathbf{D}_{\phi}\left(\x_i, \x^\mathcal{P}_i\right)\right) + \sum_{k=1}^K{\exp\left(-\mathbf{D}_{\phi}\left(\x_i, \x^\mathcal{N}_{ik}\right)\right)}}\notag\\
	=\;\;& \ln \left(1 + \sum_{k=1}^K{\exp\left(\mathbf{D}_{\phi}\left(\x_i, \x^\mathcal{P}_i\right)-\mathbf{D}_{\phi}\left(\x_i, \x^\mathcal{N}_{ik}\right)\right)}\right)\label{eq:multi-logistic}\;.
\end{align}
By minimizing the KL-divergence between the student's similarity prediction $p_{i}(\phi)$ with $\mathbf{e}_0$, we get a multi-impostor extension of the logistic loss. With a bit abuse of notation, we also denote the loss in Eq.~\ref{eq:multi-logistic} as $\iota\left(\mathbf{Diff}_{\x_i}\right)$. Eq.~\ref{eq:multi-logistic} promotes the probability between the anchor and the similar term to be the largest one --- the similarity between the anchor and the target neighbor becomes larger than those between the anchor and impostors. 
The comparison matching adds another rectification term, \ie, the first term in Eq.~\ref{eq:decomposition}. If the teacher's prediction is close to the similarity supervision indicated by the label, then $p_{i}(\phi_T)$ should be close to $\mathbf{e}_0$ --- the first term approaches to zero and Eq.~\ref{eq:decomposition} degenerates to the vanilla loss. 

On the contrary, if the teacher's comparison over a tuple is not consistent with their relationship indicated by the class labels, \eg, those impostors should not be too distant from the anchor and the anchor should have relative close distance between target neighbor and impostors. In this case, $p_{i}(\phi_T) - \mathbf{e}_0$ becomes a vector with all negative elements except for the first one. We can analyze the effect of the rectification term by decoupling the influence of the first and other elements in $p_{i}(\phi_T) - \mathbf{e}_0$. We define
\begin{equation*}
	p^\mathcal{P}_{i}(\phi) = \frac{\exp\left(-\mathbf{D}_{\phi}\left(\x_i, \x^\mathcal{P}_i\right)\right)}{\exp\left(-\mathbf{D}_{\phi}\left(\x_i, \x^\mathcal{P}_i\right)\right) + \sum_{k=1}^K{\exp\left(-\mathbf{D}_{\phi}\left(\x_i, \x^\mathcal{N}_{ik}\right)\right)}}\;,
\end{equation*}
then we can derive $(p^\mathcal{P}_{i}(\phi) - 1)\cdot\iota\left(\mathbf{Diff}_{\x_i}\right)$ from the first element of the rectification term, which weakens the force of the contrastive loss. In addition, the other elements in the rectification term make the impostors in the tuples not far away. 
Thus, the rectification term mitigates the supervision from the labels and injects the teacher's similarity estimation of the tuple.

\section{Harmonic Mean for GKD}\label{sec:harmonic}
There are two sets of classes, \ie, the old classes $\mathcal{C}'$ used to train the teacher, and the current classes $\mathcal{C}$ in the student's task, in the cross-task distillation scenario. Given a model discerning the joint classes in $\mathcal{C}'\cup\mathcal{C}$, we collect the same number of instances from $\mathcal{C}$ and $\mathcal{C}'$ to construct the test set. For an instance $\x\in\mathcal{C}'\cup\mathcal{C}$, we may measure the performance of the model by averaging the classification accuracy over $\mathcal{C}'\cup\mathcal{C}$ for all instances in the test set. We abbreviate this process as $\mathcal{C}'\cup\mathcal{C} \rightarrow \mathcal{C}'\cup\mathcal{C}$.

Directly computing the average classification accuracy will be biased towards the current classes, since the student model has no access to the old class instances. 
Similar to~\cite{Xian2017Zero,Ye2019Learning}, we consider the harmonic mean accuracy as a more balanced measure, which is a joint measure over two kinds of mean accuracy.
First, we measure the joint classification accuracy Acc$_{\mathcal{C}'}$ for instances from the class set $\mathcal{C}'$, \ie, $\mathcal{C}' \rightarrow \mathcal{C}'\cup\mathcal{C}$. Then we compute Acc$_\mathcal{C}$ as the mean accuracy based on $\mathcal{C} \rightarrow \mathcal{C}'\cup\mathcal{C}$.
The harmonic mean accuracy is 
$$\frac{2{\rm Acc}_{\mathcal{C}'}{\rm Acc}_\mathcal{C}}{{\rm Acc}_{\mathcal{C}'} + {\rm Acc}_\mathcal{C}}\;.$$
A model has a high harmonic mean value only it has a relative higher Acc$_{\mathcal{C}'}$ and Acc$_\mathcal{C}$. In other words, the harmonic mean measures the balanced classification ability over both old and current classes.

\begin{table}[t]
\centering
\caption{\small Mean accuracy on GKD tasks upon CUB. MobileNets with different width multipliers in $\{1, 0.75, 0.5, 0.25\}$ are used as students. The teacher is MobileNet with width multiplier 1. Different overlap ratios $\{0\%, 25\%, 50\%, 75\%, 100\%\}$ between teacher's and student's class set  are considered.}
\begin{tabular}{@{}ccccc@{}}
\toprule
\multicolumn{5}{c}{Overlap Ratio = $0\%$}                          \\ \midrule
\multicolumn{1}{c|}{Width Multiplier} & 1     & 0.75  & 0.5   & 0.25  \\ \midrule
\multicolumn{1}{c|}{Student}          & 71.25 & 67.56 & 66.85 & 64.48 \\
\multicolumn{1}{c|}{RKD}              & 72.24 & 68.42 & 66.85 & 65.74 \\
\multicolumn{1}{c|}{AML}              & 72.86 & 68.79 & 68.59 & 66.83 \\
\multicolumn{1}{c|}{{\name}$^\mathrm{EMB}$}  & 71.88 & 68.02 & 67.40 & 65.12 \\
\multicolumn{1}{c|}{{\name}$^{-}$}    & 74.07 & 70.93 & 70.62 & 67.58 \\
\multicolumn{1}{c|}{{\name}}          & 75.13 & 71.67 & 71.06 & 68.22 \\ \midrule \midrule
\multicolumn{5}{c}{Overlap Ratio = $25\%$}                         \\ \midrule
\multicolumn{1}{c|}{Width Multiplier} & 1     & 0.75  & 0.5   & 0.25  \\ \midrule
\multicolumn{1}{c|}{Student}          & 71.30 & 71.08 & 68.56 & 65.71 \\
\multicolumn{1}{c|}{RKD}              & 72.07 & 71.70 & 68.56 & 66.43 \\
\multicolumn{1}{c|}{AML}              & 72.35 & 72.05 & 70.37 & 67.20 \\
\multicolumn{1}{c|}{{\name}$^\mathrm{EMB}$}  & 72.04 & 71.36 & 68.90 & 66.29 \\
\multicolumn{1}{c|}{{\name}$^{-}$}    & 74.14 & 73.09 & 72.33 & 68.96 \\
\multicolumn{1}{c|}{{\name}}          & 75.09 & 73.92 & 72.99 & 70.04 \\ \midrule \midrule
\multicolumn{5}{c}{Overlap Ratio = $50\%$}                         \\ \midrule
\multicolumn{1}{c|}{Width Multiplier} & 1     & 0.75  & 0.5   & 0.25  \\ \midrule
\multicolumn{1}{c|}{Student}          & 68.20 & 66.11 & 65.23 & 62.26 \\
\multicolumn{1}{c|}{RKD}              & 68.72 & 66.82 & 65.58 & 62.79 \\
\multicolumn{1}{c|}{AML}              & 67.94 & 67.34 & 66.29 & 63.64 \\
\multicolumn{1}{c|}{{\name}$^\mathrm{EMB}$}  & 68.79 & 66.56 & 65.68 & 62.94 \\
\multicolumn{1}{c|}{{\name}$^{-}$}    & 69.14 & 67.43 & 67.18 & 64.62 \\
\multicolumn{1}{c|}{{\name}}          & 70.25 & 68.39 & 68.50 & 65.33 \\ \midrule \midrule
\multicolumn{5}{c}{Overlap Ratio = $75\%$}                         \\ \midrule
\multicolumn{1}{c|}{Width Multiplier} & 1     & 0.75  & 0.5   & 0.25  \\ \midrule
\multicolumn{1}{c|}{Student}          & 65.53 & 66.73 & 64.10 & 60.81 \\
\multicolumn{1}{c|}{RKD}              & 65.89 & 67.28 & 64.66 & 61.35 \\
\multicolumn{1}{c|}{AML}              & 66.32 & 66.92 & 65.03 & 62.09 \\
\multicolumn{1}{c|}{{\name}$^\mathrm{EMB}$}  & 66.47 & 67.20 & 64.56 & 61.37 \\
\multicolumn{1}{c|}{{\name}$^{-}$}    & 67.01 & 67.93 & 66.10 & 62.35 \\
\multicolumn{1}{c|}{{\name}}          & 67.28 & 68.35 & 66.72 & 63.03 \\ \midrule \midrule
\multicolumn{5}{c}{Overlap Ratio = $100\%$}                        \\ \midrule
\multicolumn{1}{c|}{Width Multiplier} & 1     & 0.75  & 0.5   & 0.25  \\ \midrule
\multicolumn{1}{c|}{Student}          & 67.76 & 67.98 & 64.91 & 62.17 \\
\multicolumn{1}{c|}{RKD}              & 67.23 & 68.25 & 65.73 & 62.04 \\
\multicolumn{1}{c|}{AML}              & 67.06 & 68.35 & 66.27 & 62.69 \\
\multicolumn{1}{c|}{{\name}$^\mathrm{EMB}$}  & 68.13 & 68.34 & 65.72 & 63.05 \\
\multicolumn{1}{c|}{{\name}$^{-}$}    & 68.96 & 69.07 & 68.53 & 63.10 \\
\multicolumn{1}{c|}{{\name}}          & 68.77 & 69.10 & 68.44 & 63.33 \\ \bottomrule
\end{tabular}
\label{tab:gkd_full_cub}
\end{table}

\begin{table}[]
\centering
\caption{\small Mean accuracy on GKD tasks upon CIFAR-100. Wide ResNets with different depth and width parameters are used as students, and teacher is Wide ResNet with depth 40 and width 2. Different overlap ratios in $\{0\%, 20\%, 40\%, 60\%, 80\%, 100\%\}$ are considered. {\name} outperforms other comparison methods and baselines.}
\begin{tabular}{@{}ccccc@{}}
\toprule
\multicolumn{5}{c}{Overlap Ratio = $0\%$}                                 \\ \midrule
\multicolumn{1}{c|}{(depth, width)}  & (40, 2) & (16, 2) & (40, 1) & (16, 1) \\ \midrule
\multicolumn{1}{c|}{Student}         & 81.02   & 78.94   & 78.98   & 73.70   \\
\multicolumn{1}{c|}{RKD}             & 81.46   & 79.23   & 78.80   & 73.45   \\
\multicolumn{1}{c|}{AML}             & 79.99   & 79.11   & 78.99   & 73.68   \\
\multicolumn{1}{c|}{{\name}$^\mathrm{EMB}$} & 81.35   & 79.43   & 79.34   & 74.02   \\
\multicolumn{1}{c|}{{\name}$^{-}$}   & 81.79   & 79.74   & 79.60   & 74.02   \\
\multicolumn{1}{c|}{{\name}}         & 82.60   & 80.70   & 80.18   & 74.42   \\ \midrule \midrule
\multicolumn{5}{c}{Overlap Ratio = $20\%$}                                \\ \midrule
\multicolumn{1}{c|}{(depth, width)}  & (40, 2) & (16, 2) & (40, 1) & (16, 1) \\ \midrule
\multicolumn{1}{c|}{Student}         & 80.34   & 76.12   & 75.43   & 70.84   \\
\multicolumn{1}{c|}{RKD}             & 80.56   & 76.77   & 75.82   & 71.05   \\
\multicolumn{1}{c|}{AML}             & 80.05   & 76.95   & 75.37   & 70.79   \\
\multicolumn{1}{c|}{{\name}$^\mathrm{EMB}$} & 79.89   & 76.45   & 75.99   & 71.35   \\
\multicolumn{1}{c|}{{\name}$^{-}$}   & 80.90   & 77.32   & 76.95   & 71.53   \\
\multicolumn{1}{c|}{{\name}}         & 81.40   & 77.82   & 77.24   & 72.28   \\ \midrule \midrule
\multicolumn{5}{c}{Overlap Ratio = $40\%$}                                \\ \midrule
\multicolumn{1}{c|}{(depth, width)}  & (40, 2) & (16, 2) & (40, 1) & (16, 1) \\ \midrule
\multicolumn{1}{c|}{Student}         & 78.86   & 75.67   & 74.98   & 69.36   \\
\multicolumn{1}{c|}{RKD}             & 79.33   & 75.58   & 74.69   & 69.07   \\
\multicolumn{1}{c|}{AML}             & 79.98   & 75.84   & 74.28   & 68.87   \\
\multicolumn{1}{c|}{{\name}$^\mathrm{EMB}$} & 79.27   & 76.03   & 75.38   & 69.78   \\
\multicolumn{1}{c|}{{\name}$^{-}$}   & 79.84   & 76.16   & 75.30   & 69.58   \\
\multicolumn{1}{c|}{{\name}}         & 80.40   & 76.26   & 75.62   & 70.08   \\ \midrule \midrule
\multicolumn{5}{c}{Overlap Ratio = $60\%$}                                \\ \midrule
\multicolumn{1}{c|}{(depth, width)}  & (40, 2) & (16, 2) & (40, 1) & (16, 1) \\ \midrule
\multicolumn{1}{c|}{Student}         & 78.90   & 76.37   & 75.14   & 68.48   \\
\multicolumn{1}{c|}{RKD}             & 78.69   & 76.20   & 75.50   & 68.23   \\
\multicolumn{1}{c|}{AML}             & 79.32   & 76.45   & 75.23   & 68.64   \\
\multicolumn{1}{c|}{{\name}$^\mathrm{EMB}$} & 79.33   & 76.90   & 75.67   & 69.78   \\
\multicolumn{1}{c|}{{\name}$^{-}$}   & 80.02   & 76.66   & 75.79   & 68.72   \\
\multicolumn{1}{c|}{{\name}}         & 80.66   & 76.66   & 76.52   & 69.50   \\ \midrule \midrule
\multicolumn{5}{c}{Overlap Ratio = $80\%$}                                \\ \midrule
\multicolumn{1}{c|}{(depth, width)}  & (40, 2) & (16, 2) & (40, 1) & (16, 1) \\ \midrule
\multicolumn{1}{c|}{Student}         & 80.50   & 77.43   & 76.96   & 72.16   \\
\multicolumn{1}{c|}{RKD}             & 81.21   & 77.65   & 77.34   & 72.05   \\
\multicolumn{1}{c|}{AML}             & 81.06   & 77.20   & 77.06   & 72.35   \\
\multicolumn{1}{c|}{{\name}$^\mathrm{EMB}$} & 80.70   & 77.79   & 77.54   & 72.46   \\
\multicolumn{1}{c|}{{\name}$^{-}$}   & 81.92   & 78.24   & 78.33   & 73.54   \\
\multicolumn{1}{c|}{{\name}}         & 82.56   & 78.76   & 79.28   & 73.92   \\ \midrule \midrule
\multicolumn{5}{c}{Overlap Ratio = $100\%$}                               \\ \midrule
\multicolumn{1}{c|}{(depth, width)}  & (40, 2) & (16, 2) & (40, 1) & (16, 1) \\ \midrule
\multicolumn{1}{c|}{Student}         & 80.66   & 77.94   & 76.35   & 71.56   \\
\multicolumn{1}{c|}{RKD}             & 80.52   & 78.03   & 76.82   & 72.04   \\
\multicolumn{1}{c|}{AML}             & 80.73   & 78.24   & 77.15   & 71.79   \\
\multicolumn{1}{c|}{{\name}$^\mathrm{EMB}$} & 81.02   & 78.32   & 76.67   & 72.08   \\
\multicolumn{1}{c|}{{\name}$^{-}$}   & 81.74   & 78.74   & 77.77   & 73.06   \\
\multicolumn{1}{c|}{{\name}}         & 81.58   & 78.60   & 78.04   & 73.52   \\ \bottomrule
\end{tabular}
\label{tab:gkd_full_cifar}
\end{table}

\begin{figure}[t]
	\centering
	\includegraphics[width=\textwidth]{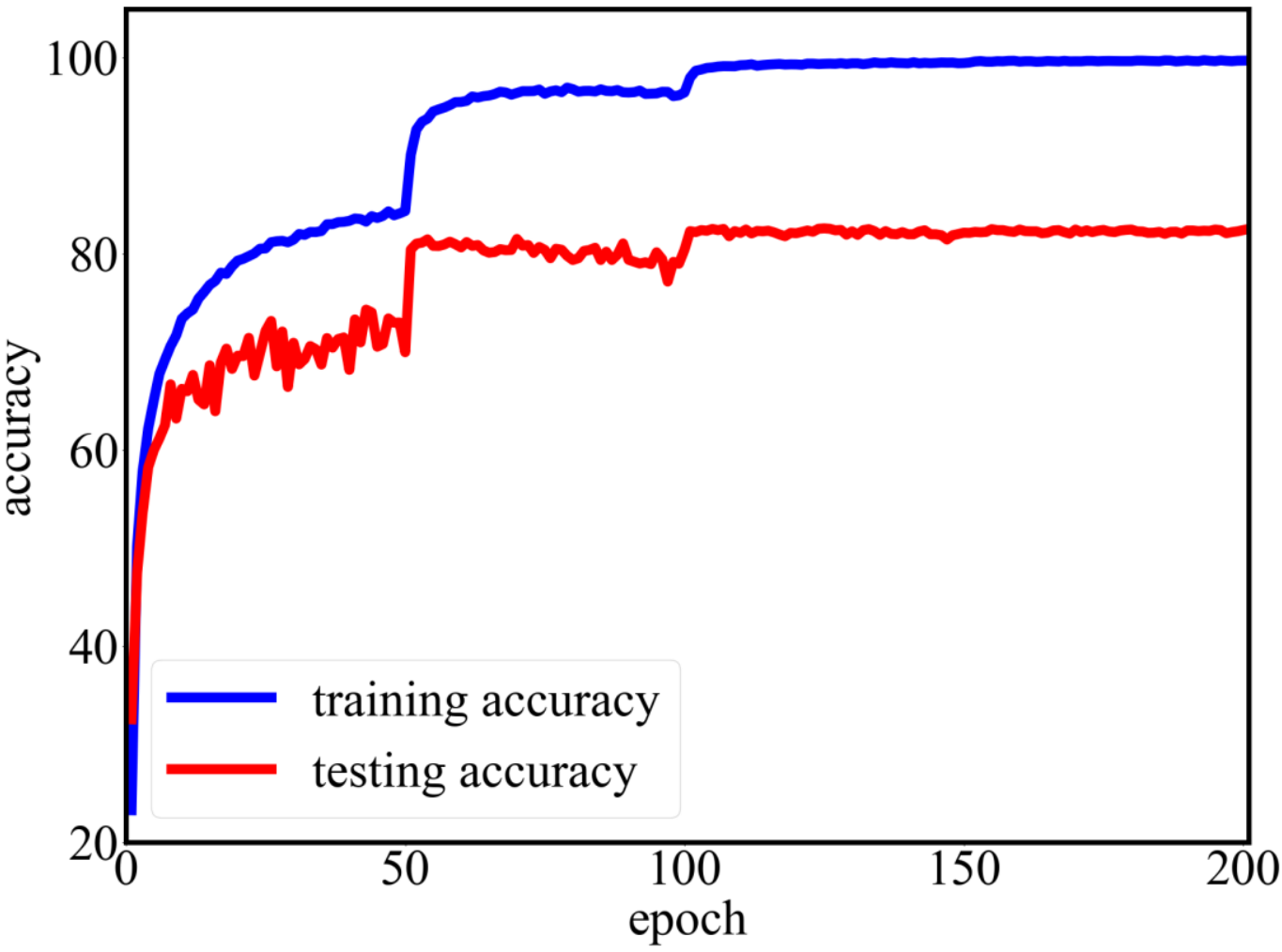}
	\caption{Convergence curve of {\name} on CIFAR.}
	\label{fig:convergence_cifar}
\end{figure}

\begin{figure}[t]
	\centering
	\includegraphics[width=\textwidth]{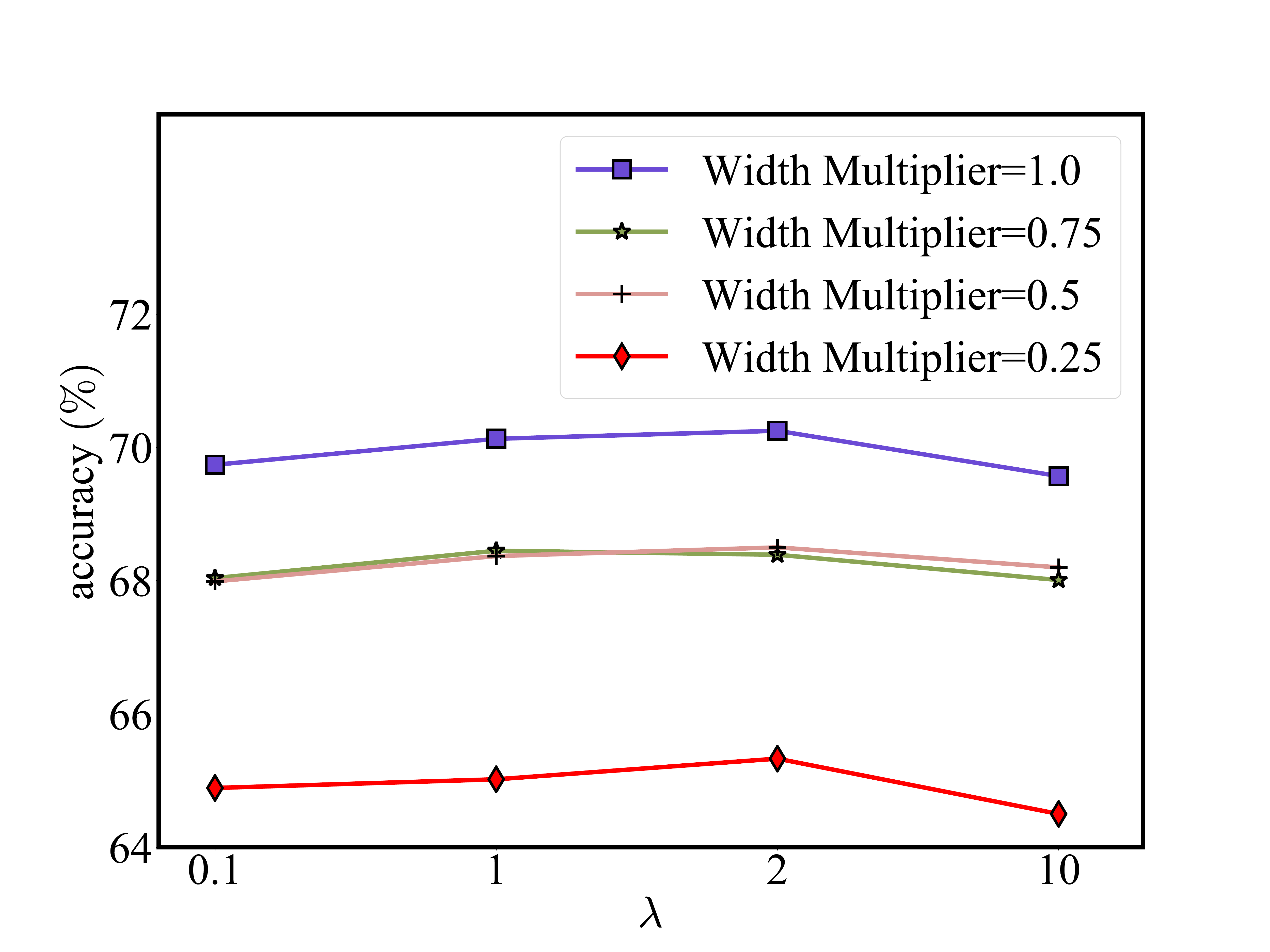}
	\caption{{\name}'s accuracy on CUB dataset width different $\lambda$ values. Class overlap ratio is set to $50\%$. Four student networks are considered, {\em i.e.}, MobileNet-\{1.0, 0.75, 0.5, 0.25\}.}
	\label{fig:lambda}
\end{figure}

\section{More Experiment Results}\label{sec:exp}
\noindent{\bf Full results for GKD.}
We list the full results of GKD evaluations on CUB and CIFAR-100 in Table~\ref{tab:gkd_full_cub} and Table~\ref{tab:gkd_full_cifar} respectively. These results are in correspondence with those in Figure 5(a) - Figure 5(h). From these two tables, we can see that {\name} outperforms other comparison methods and baseline methods when class overlap ratio ranges in $[0\%, 100\%]$. {\name} with adaptive weights improves better than the degenerated version {\name}$^-$ when the overlap ratio is low.

\begin{table}[t]
  \centering
  \caption{The average classification accuracy of standard KD on CIFAR-100. The teacher is trained with Wide ResNet with depth 28 and width 4, which gets 78.91\% test accuracy. The student is learned with Wide ResNet with different configurations of (depth, width).}
    \begin{tabular}{c|ccc}
    \addlinespace
    \toprule
    (depth, width) & (16, 4) & (28, 2) & (16, 2) \\
    \midrule
    Student & 77.28 & 75.12 & 72.68 \\
    KD~\cite{hinton2015kd}    & 78.31 & 76.57 & 73.53 \\
    FitNet~\cite{RomeroBKCGB14} & 78.15 & 76.06 & 73.7 \\
    AT~\cite{ZagoruykoK16a}    & 77.93 & 76.20  & 73.44 \\
    Jacobian~\cite{Srinivas2018Knowledge} & 77.82 & 76.3  & 73.29 \\
    Overhaul~\cite{Heo2019Comprehensive} & 79.11 & 78.02 & 75.92 \\
    \midrule
    {\name} & \bf 79.95 & \bf 79.04 & \bf 76.33 \\
    \bottomrule
    \end{tabular}
  \label{tab:standard_kd_add}
\end{table}

\noindent{\bf Comparisons with more methods on standard KD.} 
We compare with more methods on standard KD. We follow the setups in~\cite{Heo2019Comprehensive} on CIFAR-100. We set the teacher as a Wide ResNet with depth 28 and width 4. The test accuracy of the teacher is 78.91\%. Different architectures of the Wide ResNets student are investigated, namely, (16,4), (28,2), and (16-2).
Table~\ref{tab:standard_kd_add} shows the classification accuracy. {\name} improves the student and gets the best results in all cases.

\noindent{\bf Number of negative instances $K$.}
In {\name}, we formulate the similarity relationship between an instance $\mathbf{x}_i$ and other instances into a tuple, {\em i.e.}, $(\mathbf{x}_i, \mathbf{x}_i^{\mathcal{P}}, \mathbf{x}_{i1}^{\mathcal{N}}, \ldots, \mathbf{x}_{iK}^{\mathcal{K}})$. Here $K$ is the number of negative instances in a tuple. In experiments, we set $K$ to the minimum number of available negative instances of each anchor point in a mini-batch. We discuss the influence of $K$ on model performance. 

We set $K=\min(K^\prime,\; \hat{K})$ where $K^\prime$ is the number of semi-hard negatives for a pair of anchor and neighbor instances. $\hat{K}$ is a manually tuned hyper-parameter. 
If $\hat{K}$ is set to $1$, tuples sampled in each mini-batch degenerate to triplets, which means we keep only one impostor ($K=1$) for each comparison tuple. 
On the other hand, if $\hat{K}$ is set to a very large value, $K$ always equals $K^\prime$. In this case, there are a lot of negative instances in a tuple. 
Table~\ref{tab:ablation_k} shows embedding quality on CUB dataset when overlap ratio is $50\%$ and $0\%$. We set both teacher and student as MobileNet-1.0. We report the accuracy of NCM classifier based on student's embeddings after the first stage distillation. A larger $K^\prime$ value means we use more negative instances in a tuple. The results indicate that using more impostors in a tuple facilitates the distillation of the embedding. 

\begin{table}[t]
\centering
\caption{Quality of student's embedding trained with different $\hat{K}$ on CUB dataset. A larger $\hat{K}$ means there are more impostors in a tuple. Both teacher and student are MobileNet-1.0. Embeddings trained with tuples outperforms those trained with triplets.}
\begin{tabular}{c|ccc}
\toprule
$\hat{K}$              & 1 (triplet)     & 5     & $\infty$\; (tuple)    \\ \midrule
overlap ratio = $0\%$  & 30.54 & 31.42 & \bf 32.69 \\
overlap ratio = $50\%$ & 29.61 & 29.23 & \bf 30.27 \\ \bottomrule
\end{tabular}
\label{tab:ablation_k}
\end{table}

\noindent{\bf Convergence curve.}
We plot convergence curve of {\name}'s training process on CIFAR-100 in Figure~\ref{fig:convergence_cifar}. We set the overlap ratio to $60\%$. The student is WRN-(40,2). We find that {\name} successfully converges to a stable point. Convergence curves under other experiment configurations are similar to the plotted one.

\noindent{\bf Influence of the hyper-parameter $\lambda$.}
We study the influence of hyper-parameter $\lambda$. Figure~\ref{fig:lambda} shows {\name}'s testing accuracy on CUB with different $\lambda$ values, and the class overlap ratio is set to $50\%$. Various architectures of the student network are considered. The results indicate that {\name} achieves the highest performance when $\lambda=2$, and the influence of $\lambda$ on the model's performance is limited.

\begin{figure}[t]
	\begin{subfigure}[b]{0.46\linewidth}
	\centering
	\includegraphics[width=\textwidth]{figures/dis-cifar.pdf}
	\caption{Overlap ratio = $60\%$.}
	\label{fig:weight_dis_gkd}
	\end{subfigure}
	\begin{subfigure}[b]{0.52\linewidth}
	\centering
	\includegraphics[width=\textwidth]{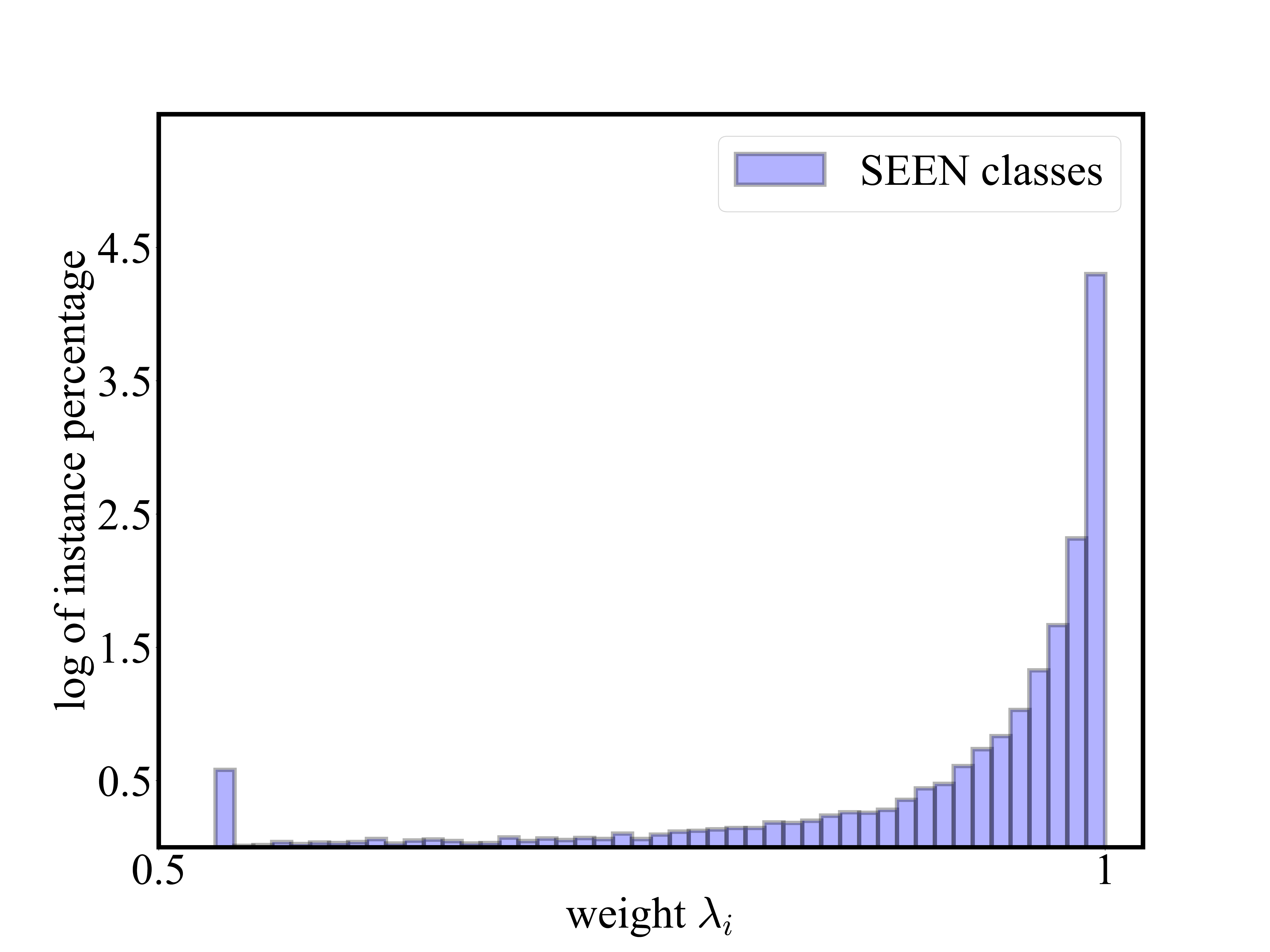}
	\caption{Overlap ratio = $100\%$.}
	\label{fig:weight_dis_skd}
	\end{subfigure}
	\caption{The histogram of $\lambda_i$ over all target task instances with a learned {\name} model. We set $\lambda=1$ and get $\lambda_i\in(0, 1]$. The classes the teacher trained on are denoted as ``seen'', otherwise denoted as ``unseen".
	(a) When the class overlap ratio is $60\%$, the weights of instances from seen classes are observably higher than those of unseen classes, and it is easy to recognize unseen classes. (b) When the ratio is $100\%$, almost all instances have weights close to $1$.}
\end{figure}

\noindent{\bf The effect of the adaptive weight $\lambda_i$.}
The instance-specific weights $\lambda_i$ in Eq. 9 in the main paper is a key component for GKD. Since there simultaneously exist cross-task instances (denoted as unseen) and same-task instances (denoted as seen) in the student’s training set, and the teacher may predict with different confidences to them. For example, the teacher will have higher confidence over those seen instances where the teacher is trained from, and lower confidence otherwise. 

We investigate the distribution of $\lambda_i$ in two GKD scenarios on CIFAR-100 where the class overlap is 60\% and 100\%. 
The distribution (histogram) of $\lambda_i$ based on a learned {\name} model in GKD is shown in Figure~\ref{fig:weight_dis_gkd}.
We set $\lambda$ to 1 so that we have $\lambda_i\in(0,1]$. 
We find weights for instances belonging to the seen classes are observably larger than those for instances belonging to unseen classes. This means our re-weight strategy can successfully recognize seen/unseen classes and put larger weights on those familiar instances automatically.
We compute the AUC when we use $\lambda_i$ to differentiate seen and unseen classes in the GKD scenario, whose value is $0.959$. Thus, $\lambda_i$ extracts useful information from the teacher’s supervision adaptively.

We also plot the results on standard KD in Figure~\ref{fig:weight_dis_skd}. In standard KD, all instances come from seen classes, and the learned {\name} model uses larger weights for most instances. There are still a small number of instances that have weights $\lambda_i\approx0.5$, which may be noisy ones that the teacher cannot provide confident predictions. The teacher's supervision will be weakened with a relatively smaller $\lambda_i$.

\end{document}